\pdfoutput=1

\documentclass[acmsmall,screen]{acmart}
\usepackage{amsmath,amsfonts}

\usepackage{algorithm}
\usepackage{algorithmic}
\usepackage{booktabs}
\usepackage{subfigure}
\usepackage{multirow}
\usepackage[switch]{lineno}
\usepackage{titlesec}
\usepackage{makecell}
\usepackage{hyperref}

\AtBeginDocument{%
  \providecommand\BibTeX{{%
    \normalfont B\kern-0.5em{\scshape i\kern-0.25em b}\kern-0.8em\TeX}}}

\setcopyright{acmcopyright}
\copyrightyear{2024}
\acmYear{2024}
\acmDOI{XXXXXXX.XXXXXXX}

\acmJournal{JACM}

\begin{document}

\title{A Comprehensive Survey on Deep Graph Representation Learning}


\author{Wei Ju}
\email{juwei@pku.edu.cn}
\author{Zheng Fang}
\email{fang_z@pku.edu.cn}
\author{Yiyang Gu}
\email{yiyanggu@pku.edu.cn}
\author{Zequn Liu}
\email{zequnliu@pku.edu.cn}
\author{Qingqing Long}
\email{qingqinglong@pku.edu.cn}
\affiliation{%
  \institution{Peking University}
  \city{Beijing}
  \country{China}
  \postcode{100871}
}

\author{Ziyue Qiao}
\email{ziyuejoe@gmail.com}
\affiliation{%
  \institution{The Hong Kong University of Science and Technology}
  \city{Guangzhou}
  \country{China}
  \postcode{511453}
}

\author{Yifang Qin}
\email{qinyifang@pku.edu.cn}
\author{Jianhao Shen}
\email{jhshen@pku.edu.cn}
\affiliation{%
  \institution{Peking University}
  \city{Beijing}
  \country{China}
  \postcode{100871}
}

\author{Fang Sun} 
\email{fts@cs.ucla.edu}
\author{Zhiping Xiao}
\email{patricia.xiao@cs.ucla.edu}
\affiliation{%
  \institution{University of California, Los Angeles}
  \country{USA}
  \postcode{90095}
}

\author{Junwei Yang}
\email{yjwtheonly@pku.edu.cn}
\author{Jingyang Yuan}
\email{yuanjy@pku.edu.cn}
\author{Yusheng Zhao}
\email{yusheng.zhao@stu.pku.edu.cn}
\affiliation{%
  \institution{Peking University}
  \city{Beijing}
  \country{China}
  \postcode{100871}
}

\author{Yifan Wang}
\email{yifanwang@uibe.edu.cn}
\affiliation{%
  \institution{University of International Business and Economics}
  \city{Beijing}
  \country{China}
  \postcode{100029}
}

\author{Xiao Luo}
\email{xiaoluo@cs.ucla.edu}
\affiliation{%
  \institution{University of California, Los Angeles}
  \country{USA}
  \postcode{90095}
}
\authornote{Corresponding authors.}

\author{Ming Zhang}
\email{mzhang_cs@pku.edu.cn}
\affiliation{%
  \institution{Peking University}
  \city{Beijing}
  \country{China}
  \postcode{100871}
}
\authornotemark[1]

\renewcommand{\shortauthors}{W. Ju, et al.}


\begin{abstract}
Graph representation learning aims to effectively encode high-dimensional sparse graph-structured data into low-dimensional dense vectors, which is a fundamental task that has been widely studied in a range of fields, including machine learning and data mining. Classic graph embedding methods follow the basic idea that the embedding vectors of interconnected nodes in the graph can still maintain a relatively close distance, thereby preserving the structural information between the nodes in the graph. However, this is sub-optimal due to: (i) traditional methods have limited model capacity which limits the learning performance; (ii) existing techniques typically rely on unsupervised learning strategies and fail to couple with the latest learning paradigms; (iii) representation learning and downstream tasks are dependent on each other which should be jointly enhanced. With the remarkable success of deep learning, deep graph representation learning has shown great potential and advantages over shallow (traditional) methods, there exist a large number of deep graph representation learning techniques have been proposed in the past decade, especially graph neural networks. In this survey, we conduct a comprehensive survey on current deep graph representation learning algorithms by proposing a new taxonomy of existing state-of-the-art literature. Specifically, we systematically summarize the essential components of graph representation learning and categorize existing approaches by the ways of graph neural network architectures and the most recent advanced learning paradigms. Moreover, this survey also provides the practical and promising applications of deep graph representation learning. Last but not least, we state new perspectives and suggest challenging directions which deserve further investigations in the future.

\end{abstract}



\maketitle

\section{Introduction}
Graphs have recently emerged as a powerful tool for representing a variety of structured and complex data, including social networks, traffic networks, information systems, knowledge graphs, protein-protein interaction networks, and physical interaction networks. As a kind of general form of data organization, graph structures are capable of naturally expressing the intrinsic relationship of these data, and thus can characterize plenty of non-Euclidean structures that are crucial in a variety of disciplines and domains due to their flexible adaptability. For example, to encode a social network as a graph, nodes on the graph are used to represent individual users, and edges are used to represent the relationship between two individuals, such as friends. In the field of biology, nodes can be used to represent proteins, and edges can be used to represent biological interactions between various proteins, such as the dynamic interactions between proteins. Thus, by analyzing and mining the graph-structured data, we can understand the deep meaning hidden behind the data, and further discover valuable knowledge, so as to benefit society and human beings.

In the last decade years, a wide range of machine learning algorithms have been developed for graph-structured data learning. Among them, traditional graph kernel methods~\citep{gartner2003graph,kashima2003marginalized,shervashidze2011weisfeiler,shervashidze2009efficient} usually break down graphs into different atomic substructures and then use kernel functions to measure the similarity between all pairs of them. Although graph kernels could provide a perspective on modeling graph topology, these approaches often generate substructures or feature representations based on given hand-crafted criteria. These rules are rather heuristic, prone to suffer from high computational complexity, and therefore have weak scalability and subpar performance.

In the past few years, graph embedding algorithms~\citep{ahmed2013distributed,perozzi2014deepwalk,tang2015line,grover2016node2vec,tang2015pte,wang2016structural} have ever-increasing emerged, which attempt to encode the structural information of the graph (usually a high-dimensional sparse matrix) and map it into a low-dimensional dense vector embedding to preserve the topology information and attribute information in the embedding space as much as possible, so that the learned graph embeddings can be naturally integrated into traditional machine learning algorithms. Compared to previous works which use feature engineering in the pre-processing phase to extract graph structural features, current graph embedding algorithms are conducted in a data-driven way leveraging machine learning algorithms (such as neural networks) to encode the structural information of the graph. Specifically, existing graph embedding methods can be categorized into the following main groups: (i) matrix factorization based methods~\citep{ahmed2013distributed,ou2016asymmetric,cao2015grarep} that factorize the matrix to learn node embedding which preserves the graph property; (ii) deep learning based methods~\citep{perozzi2014deepwalk,tang2015line,grover2016node2vec,wang2016structural} that apply deep learning techniques specifically designed for graph-structured data; (iii) edge reconstruction based methods~\citep{tang2015pte,man2016predict,liu2016aligning} that either maximizes edge reconstruction probability or minimizes edge reconstruction loss. Generally, these methods typically depend on shallow architectures, and fail to exploit the potential and capacity of deep neural networks, resulting in sub-optimal representation quality and learning performance.

\begin{figure*}
    \centering
    \includegraphics[width=1\textwidth]{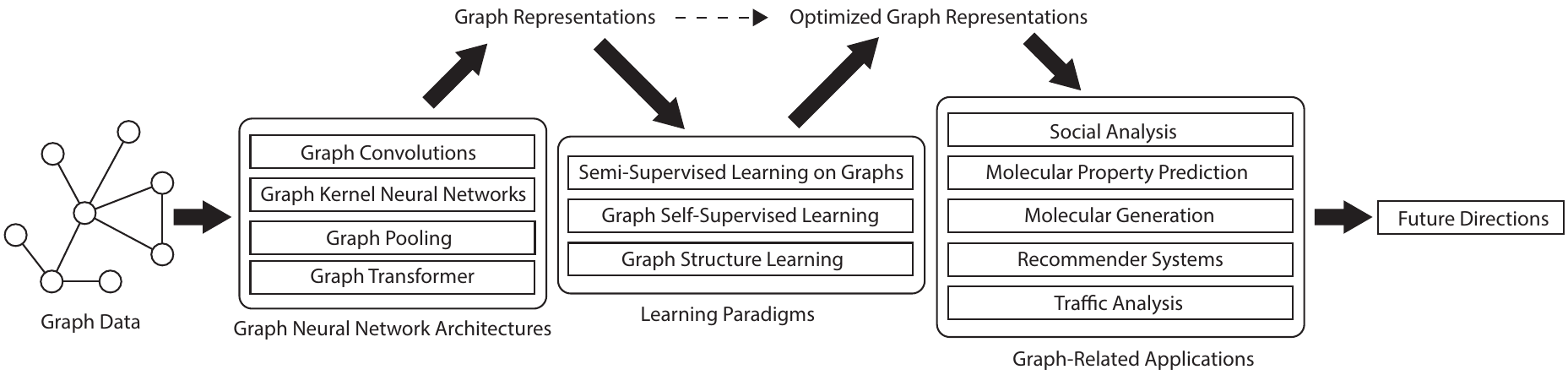}
    \caption{The architecture of this paper.} 
    \label{fig:framework}
\end{figure*}

Inspired by the recent remarkable success of deep neural networks, a range of deep learning algorithms has been developed for graph-structured data learning.
The core of these methods is to generate effective node and graph representations using graph neural networks (GNNs), followed by a goal-oriented learning paradigm. In this way, the derived representations can be adaptively coupled with a variety of downstream tasks and applications. Following this line of thought, in this paper, we propose a new taxonomy to classify the existing graph representation learning algorithms, i.e., graph neural network architectures, learning paradigms, and various promising applications, as shown in Fig.~\ref{fig:framework}. Specifically, for the architectures of GNNs, we investigate the studies on graph convolutions, graph kernel neural networks, graph pooling, and graph transformer. For the learning paradigms, we explore three advanced types namely supervised/semi-supervised learning on graphs, graph self-supervised learning, and graph structure learning. To demonstrate the effectiveness of the learned graph representations, we provide several promising applications to build tight connections between representation learning and downstream tasks, such as social analysis, molecular property prediction and generation, recommender systems, and traffic analysis. Last but not least, we present some perspectives for thought and suggest challenging directions that deserve further study in the future.

\smallskip
\noindent\textbf{Differences between this survey and existing ones}.
Up to now, there exist some other overview papers focusing on different perspectives of graph representation learning\citep{wu2020comprehensive,zhou2020graph,zhang2020deep,chami2022machine,bacciu2020gentle,xia2021graph,zhou2022graph,khoshraftar2022survey,chen2022survey,chen2020graph} that are closely related to ours. However, there are very few comprehensive reviews have summarized deep graph representation learning simultaneously from the perspective of diverse GNN architectures and corresponding up-to-date learning paradigms. Therefore, we here clearly state their distinctions from our survey as follows. There have been several surveys on classic graph embedding\citep{goyal2018graph,cai2018comprehensive}, these works categorize graph embedding
methods based on different training objectives. Wang et al.~\citep{wang2022survey} goes further and provides a comprehensive review of existing heterogeneous graph embedding approaches. With the rapid development of deep learning, there are a handful of surveys along this line. For example, Wu et al.~\citep{wu2020comprehensive} and Zhang et al.~\citep{zhang2020deep} mainly focus on several classical and representative GNN architectures without exploring deep graph representation learning from a view of the most recent advanced learning paradigms such as graph self-supervised learning and graph structure learning. Xia et al.~\citep{xia2021graph} and Chami et al.~\citep{chami2022machine} jointly summarize the studies of graph embeddings and GNNs. Zhou et al.~\citep{zhou2020graph} explores different types of computational modules for GNNs. One recent survey under review \citep{khoshraftar2022survey} categorizes the existing works in graph representation learning from both static and dynamic graphs. However, these taxonomies emphasize the basic GNN methods but pay insufficient attention to the learning paradigms, and provide few discussions of the most promising applications, such as recommender systems as well as molecular property prediction and generation. To the best of our knowledge, the most relevant survey published formally is \citep{zhou2022graph}, which presents a review of GNN architectures and roughly discusses the corresponding applications. Nevertheless, this survey merely covers methods up to the year of 2020, missing the latest developments in the past three years. 

Therefore, it is highly desired to summarize the representative GNN methods, the most recent advanced learning paradigms, and promising applications into one unified and comprehensive framework. Moreover, we strongly believe this survey with a new taxonomy of literature and more than $600$ studies will strengthen future research on deep graph representation learning.

\smallskip
\noindent\textbf{Contribution of this survey.} 
The goal of this survey is to systematically review the literature on the advances of deep graph representation learning and discuss further directions. It aims to help the researchers and practitioners who are interested in this area, and support them in understanding the panorama and the latest developments of deep graph representation learning. The key contributions of this survey are summarized as follows:

\begin{itemize}
    \item \textbf{Systematic Taxonomy.} We propose a systematic taxonomy to organize the existing deep graph representation learning approaches based on the ways of GNN architectures and the most recent advanced learning paradigms via providing some representative branches of methods. Moreover, several promising applications are presented to illustrate the superiority and potential of graph representation learning.
    \item \textbf{Comprehensive Review.} For each branch of this survey, we review the essential components and provide detailed descriptions of representative algorithms, and systematically summarize the characteristics to make the overview comparison.
    \item \textbf{Future Directions.} 
    Based on the properties of existing deep graph representation learning algorithms, we discuss the limitations and challenges of current methods and propose the potential as well as promising research directions deserving of future investigations. 
\end{itemize}

\section{Background}

In this section, we first briefly introduce some definitions in deep graph representation learning that need to be clarified, and then we explain the reasons why we need graph representation learning.

\subsection{Problem Definition}

\smallskip
\noindent\textbf{Definition: Graph.} Given a graph $G=(V, E, \mathbf{X})$, where $V = \{v_{1}, \cdots, v_{|V|}\}$ is the set of nodes, $E = \{e_{1}, \cdots, e_{|V|}\}$ is the set of edges, and the edge $e = (v_i, v_j)\in E$ represent the connection relationship between nodes $v_i$ and $v_j$ in the graph. $\mathbf{X} \in \mathbb{R}^{|V| \times M}$ is the node feature matrix with $M$ being the dimension of each node feature. The adjacency matrix of a graph can be defined as $\mathbf{A}\in \mathbb{R}^{|V| \times |V|}$, where $\mathbf{A}_{ij}=1$ if $(v_i, v_j)\in E$, otherwise $\mathbf{A}_{ij}=0$.

The adjacency matrix can be regarded as the structural representation of the graph-structured data, in which each row of the adjacency matrix $\mathbf{A}$ represents the connection relationship between the corresponding node of the row and all other nodes, which can be regarded as a discrete representation of the node. However, in real-life circumstances, the adjacency matrix $\mathbf{A}$ corresponding to $G$ is a highly sparse matrix, and if $\mathbf{A}$ is used directly as node representations, it will be seriously affected by impractical storage demands and computational overhead. The storage space of the adjacency matrix $\mathbf{A}$ is $|V| \times |V|$, which is usually unacceptable when the total number of nodes grows to the order of millions. At the same time, the value of most dimensions in the node representation is $0$. The sparsity will make subsequent machine learning tasks very difficult.

\emph{Graph representation learning} is a bridge between the original input data and the task objectives in the graph. The fundamental idea of the graph representation learning algorithm is first to learn the embedded representations of nodes or the entire graph from the input graph structure data and then apply these embedded representations to downstream related tasks, such as node classification, graph classification, link prediction, community detection, and visualization, etc. Specifically, it aims to learn low-dimensional, dense distributed embedding representations for nodes in the graph. Formally, the goal of graph representation learning is to learn its embedding vector representation $R_v \in \mathbb{R}^d$ for each node $v \in V$, where the dimension $d$ of the vector is much smaller than the total number of nodes $|V|$ in the graph.

\subsection{Traditional Graph Embedding}
Traditional graph embedding learning methods, as part of dimensionality reduction techniques, aimed to embed graph data into a lower-dimensional vector space with the idea that connected nodes in the graph should still be closer to each other in this lower-dimensional space, thereby preserving the structural information between nodes in the graph. Influenced by classical dimensionality reduction techniques, early graph embedding methods are primarily inspired by classic matrix factorization techniques~\citep{belkin2001laplacian} and multi-dimensional scaling~\citep{kruskal1964multidimensional}. The following three sections describe these methods in more detail, distinguishing among matrix factorization-based methods, random walks-based methods and other non-GNN deep methods. In Table~\ref{tab:traditional}, we summarize different categories of traditional graph embedding methods.

\begin{table}
\caption{Summary of traditional graph embedding methods.}
\label{tab:traditional}
\centering
\resizebox{0.8\linewidth}{!} {
\begin{tabular}{ccccc}
\toprule
Type & Method & Similarity measure & Loss function ($L$)\\
\midrule
\multirow{5}{*}{Matrix Factorization} 
& {\small LLE} \citep{roweis2000nonlinear} & general & ${|z_i - \sum_{j\in N_i}{W_{ij}z_j}|}^2$\\

& {\small LE} \citep{anderson1985eigenvalues} & general & $Z^TLZ,\text{s.t.}Z^TDZ=I$\\

& {\small GF} \citep{ahmed2013distributed} & $A_{i,j}$ & ${|W_{i,j} - \langle z_i, z_j \rangle|}^2$\\

& {\small GraRep} \citep{cao2015grarep} & $A_{i,j}, A^2_{i,j}, ..., A^k_{i,j}$ & ${|W_{i,j} - \langle z_i, z_j \rangle|}^2$\\

& {\small HOPE} \citep{ou2016asymmetric} & general & ${|W_{i,j} - \langle z_i, z_j \rangle|}^2$\\
\midrule

\multirow{4}{*}{Random Walk}
& {\small DeepWalk} \citep{perozzi2014deepwalk} & $p(v_i | v_i)$ & $-A_{ij}\log\langle z_i, z_j \rangle$\\
 
& {\small Node2vec} \citep{grover2016node2vec} & $p(v_i | v_i)$  (biased)& $-A_{ij}\log\langle z_i, z_j \rangle$ \\

& {\small HARP} \citep{chen2018harp} & $p(v_i | v_i)$  (biased)& $-A_{ij}\log\langle z_i, z_j \rangle$\\

& {\small LINE} \citep{tang2015line} & \text{Two-order Similarities} & \text{Corresponding Loss}\\
\midrule

\multirow{2}{*}{Non-GNN Deep}
& {\small SDNE} \citep{wang2016structural} & \text{Two-order Proximities} &\text{Corresponding Loss}\\

& {\small DNGR} \citep{cao2016deep} & \text{Two-order Proximities} & \text{Corresponding Loss}\\
\bottomrule
\end{tabular}
}
\end{table}

\subsubsection{Matrix factorization-based methods}
Matrix factorization-based methods are the early endeavors in graph embedding learning. These approaches can be outlined in a two-step process. In the initial step, a proximity-based matrix is constructed for the graph, where each element of the matrix represents the proximity measure between two nodes in the graph. Subsequently, a dimensionality reduction technique is employed on this matrix in the second step to generate the node embeddings.

\smallskip
\emph{Locally Linear Embedding}~\emph{(LLE)}~\citep{roweis2000nonlinear}. LLE assumes that node representations are sampled from the same manifold space, and any node in the graph and its neighboring nodes are located in a local region of that manifold space. Therefore, node representations can be obtained by linearly combining them with their neighboring nodes. LLE first constructs a local reconstruction weight matrix, $W_{ij}$, for nodes in the graph to linearly combine neighboring nodes. By computing the distance between the linear combination and the central node, the problem is reduced to solving for matrix eigenvalues to learn low-dimensional vector representations for nodes. The objective function is computed as follows:
\begin{equation}
\phi(Z) = \frac{1}{2}\sum_{i}{|z_i - \sum_{j\in N_i}{W_{ij}z_j}|}^2,
\end{equation}
where $z_i$ represents the low-dimensional representation of the $i$-th node, and $N_i$ is the set of neighboring nodes for the central node $i$.

\smallskip
\emph{Laplacian Eigenmaps}~\emph{(LE)}~\citep{anderson1985eigenvalues}. LE believes that nodes directly connected in graph data should be kept as close as possible in the embedding space. Specifically, it achieves this by defining the distance between connected nodes in the embedding space using the square of the Euclidean distance. It transforms the final optimization objective into the computation of the Laplacian matrix's eigenvectors. The objective function is computed as follows:
\begin{equation}
\phi(Z) = \frac{1}{2}\sum_{i,j}{|z_i-z_j|^2}W_{ij}=Z^TLZ,\qquad\text{s.t.\ \ }Z^TDZ=I,
\end{equation}
where $W_{ij}$ represents the connection weight between nodes $i$ and $j$ in the graph. After linear transformation, the optimization of $\phi(Z)$ can be reformulated as $Z^TLZ$, where $L = D - W$ is the constructed graph Laplacian matrix, and $D$ is a symmetric matrix.

\smallskip
\emph{Graph Factorization}~\emph{(GF)}~\citep{ahmed2013distributed}. The matrix eigenvector-based methods mentioned before consider the similarity between nodes throughout the entire graph, which can result in excellent node feature representations. However, with the ever-growing scale of real-world graph data, computing matrix eigenvectors for large graphs can be computationally expensive and memory-intensive. GF introduces a graph embedding method with a time complexity of $O(|E|)$ by factorizing the adjacency matrix of the graph. The objective function is as follows:
\begin{equation}
\phi(Z, \lambda) = \frac{1}{2}\sum_{i,j\in E}{|W_{i,j} - \langle z_i, z_j \rangle|}^2 + \frac{\lambda}{2}\sum_{i}{|z_i|^2},
\end{equation}
where $\lambda$ is a regularization coefficient, and $\langle z_i, z_j \rangle$ represents the corresponding inner-product operation. Moreover, these inner-product methods also contain GraRep~\citep{cao2015grarep} and HOPE~\citep{ou2016asymmetric}, which consider higher-order and general node similarity respectively.

\subsubsection{Random walk-based methods}
Random walk-based methods have also attracted a lot of attention in graph embedding learning. The basic idea of these methods is to create random walks among nodes in the graph to capture its structural characteristics. Thus, nodes tend to have similar embedding if they co-occur on short random walks. Compared to fixed proximity measures in traditional matrix factorization-based methods, these approaches use co-occurrence in a random walk as a measure of node similarity, which is more flexible and has demonstrated promising performance across various applications.

\smallskip
\emph{DeepWalk}~\citep{perozzi2014deepwalk}. DeepWalk analogizes nodes in a graph to words in text. It uses random walks on the graph to generate numerous node sequences $S=\{v_1,\dots,v_{|s|}\}$, treating these sequences as sentences, and then inputting them into the Word2vec~\citep{mikolov2013efficient}, which aims to maximize the probability of node context given the target node $v_i$. It can be written as: 
\begin{equation}
\frac{1}{|S|}\sum_{i=1}^{|S|}\sum_{-t\leq j\leq t,j\neq 0}\log p(v_{i+j}|v_i),
\end{equation}
where $t$ is the context window size. Compared to matrix factorization-based methods, DeepWalk exhibits extremely low time complexity and is suitable for large-scale graph representation learning. However, DeepWalk only considers local information between nodes in the graph, making it challenging to find the optimal random walk sampling sequences. 

\smallskip
\emph{Node2vec}~\citep{grover2016node2vec}. Based on DeepWalk, Node2vec utilizes parameters $p$ and $q$ to guide the random walk. Parameter $p$ allows the algorithm to revisit previously traversed nodes $t$, with smaller values of $p$ increasing the likelihood of returning to $t$. Parameter $q$ facilitates both inward and outward exploration; when $q>1$, the algorithm tends to visit nodes closer to $t$; while for $q<1$, it leans toward nodes farther away from $t$. Thus, it approximates a balance between two sampling strategies, Depth-First Search (DFS) and Breadth-First Search (BFS), which can better capture topological information in the graph. Moreover, there are also some approaches that leverage graph preprocessing, i.e., HARP~\citep{chen2018harp} or different random walk strategies~\citep{perozzi2017don} to extend random walk approaches.

\smallskip
\emph{Large-Scale Information Network Embeddings}~(\emph{LINE})~\citep{tang2015line}. LINE is not based on random walks but is often compared with DeepWalk and Node2vec. The method models first-order and second-order similarity between nodes in a graph. First-order similarity characterizes the similarity between two directly connected nodes in the graph and is represented through the joint probability distribution of the two nodes. 
\begin{equation}
p(v_i, v_j)=\frac{1}{1+\exp(-{z_i}^\top z_j)},
\end{equation}
where $z_i$ represents the low-dimensional representation of node $v_i$. Second-order similarity, on the other hand, describes the number of common neighboring nodes between nodes in the graph, measuring the similarity of the local neighborhood structures (context) between nodes, which can be expressed as:
\begin{equation}
p(v_j|v_i)=\frac{\exp({z'_j}^\top z_i)}{\sum_{k=1}^{|V|}\exp({z'_k}^\top\cdot z_i)},
\end{equation}
where $|V|$ represents the number of nodes in the graph, and $z'_j$ represents the representation of node $v_j$ when it serves as a specific context.

\subsubsection{Non-GNN Deep methods}
In contrast to the above shallow graph embedding methods, non-GNN deep methods directly integrate the graph's structural information into the encoder algorithm through deep neural networks. The fundamental concept underlying these methods involves employing the autoencoder approach~\citep{hinton2006reducing} to compress information related to a node's local neighborhood.

\smallskip
\emph{Structural Deep Network Embeddings}~\emph{(SDNE)}~\citep{wang2016structural}. SDNE jointly preserves the first-order and second-order proximities between the nodes in the graph. The second-order proximity of the nodes can be defined as:
\begin{equation}
L_1= \sum_{v_i\in V}|(x_i-x'_i)\odot b_i|,
\end{equation}
where $x_i$ is the row corresponding to node $v_i$ in the adjacency mactix A, $\odot$ means the Hadamard product. $b_i=\{b_{ij}\}_{j=1}^{|V|}$, if $A_{ij}=0, b_{ij}=1$; otherwise, $b_{ij}=\beta>1$. And the first-order similarity can be defined as:
\begin{equation}
    L_2=\sum_{(v_i,v_j)\in E} A_{ij}|z_i-z_j|,
\end{equation}
where $z_i$ is the learned representation of node $v_i$. 

\smallskip
\emph{Deep Neural Graph Representations}~\emph{(DNGR)}~\citep{cao2016deep}. Similar to SDNE, DNGR utilizes pointwise mutual information between two nodes co-occurring in random walks instead of the adjacency matrix values.

\subsection{Why study deep graph representation learning}

With the rapid development of deep learning techniques, deep neural networks such as convolutional neural networks and recurrent neural networks have made breakthroughs in the fields of computer vision, natural language processing, and speech recognition. They can well abstract the semantic information of images, natural languages, and speeches. However, current deep learning techniques fail to handle more complex and irregular graph-structured data. To effectively analyze and model this kind of non-Euclidean structure data, many graph representation learning algorithms have emerged in recent years, including graph embedding and graph neural networks. At present, compared with Euclidean-style data such as images, natural language, and speech, graph-structured data is high-dimensional, complex, and irregular. Therefore, the graph representation learning algorithm is a rather powerful tool for studying graph-structured data. To encode complex graph-structured data, deep graph representation learning needs to meet several characteristics: (1) \emph{topological properties}: Graph representations need to capture the complex topological information of the graph, such as the relationship between nodes and nodes, and other substructure information, such as subgraphs, motif, etc; (2) \emph{feature attributes}: It is necessary for graph representations to describe high-dimensional attribute features in the graph, including the attributes of nodes and edges themselves; (3) \emph{scalability}: Because different real graph data have different characteristics, graph representation learning algorithms should be able to efficiently learn its embedding representation on different graph structure data, making it universal and transferable.

\section{Graph Convolutions}
Graph convolutions have become the basic building blocks in many deep graph representation learning algorithms and graph neural networks developed recently. In this section, we provide a comprehensive review of graph convolutions, which generally fall into two categories: spectral graph convolutions and spatial graph convolutions. Based on the solid mathematical foundations of Graph Signal Processing (GSP)~\citep{shuman2013gsp, sandryhaila2013gsp, hammond2011wavelets}, spectral graph convolutions seek to capture the patterns of the graph in the frequency domain. On the other hand, spatial graph convolutions inherit the idea of message passing from Recurrent Graph Neural Networks (RecGNNs), and they compute node features by aggregating the features of their neighbors. Thus, the computation graph of a node is derived from the local graph structure around it, and the graph topology is naturally incorporated into the way node features are computed. In this section, we first introduce spectral graph convolutions and then spatial graph convolutions, followed by a brief summary. In Table~\ref{tab:graph_conv}, we summarize a number of graph convolutions proposed in recent years.

\begin{table*}[t]
\centering
\caption{Summary of graph convolution methods.}
\label{tab:graph_conv}
\resizebox{0.9\linewidth}{!} {
\begin{tabular}{l ccc}
    \toprule
    Method  & Category & Aggregation & Time Complexity \\
    \midrule
    Spectral CNN~\citep{bruna2013spectral}  & Spectral Graph Convolution & - & $O(n^3)$ \\\midrule
    Henaff et al.~\citep{henaff2015spectral}  & Spectral Graph Convolution & - & $O(n^3)$ \\\midrule
    ChebNet~\citep{defferrard2016convolutional}  & Spectral Graph Convolution & - & $O(m)$ \\\midrule
    GCN~\citep{kipf2016semi} & Spectral / Spatial & Weighted Average & $O(m)$ \\\midrule
    CayleyNet~\citep{levie2018cayleynets}  & Spectral Graph Convolution & - & $O(m)$ \\\midrule
    GraphSAGE~\citep{hamilton2017inductive}  & Spatial Graph Convolution & General & $O(m)$\\\midrule
    GAT~\citep{velivckovic2017graph}  & Spatial Graph Convolution & Attentive & $O(m)$\\\midrule
    DGCNN~\citep{wang2019dynamic} & Spatial Graph Convolution & General & $O(m)$ \\\midrule
    LanzcosNet~\citep{liao2019lanczosnet}  & Spectral Graph Convolution & - & $O(n^2)$\\\midrule
    SGC~\citep{wu2019simplifying}  & Spatial Graph Convolution & Weighted Average & $O(m)$ \\\midrule
    GWNN~\citep{xu2019graph}  & Spectral Graph Convolution & - & $O(m)$ \\\midrule
    GIN~\citep{xu2018powerful}  & Spatial Graph Convolution & Sum & $O(m)$\\\midrule
    GraphAIR~\citep{hu2021graphair}  & Spatial Graph Convolution & Sum & $O(m)$ \\\midrule
    PNA~\citep{corso2020principal}  & Spatial Graph Convolution & Multiple & $O(m)$ \\\midrule
    S$^2$GC~\citep{zhu2021simple}  & Spectral Graph Convolution & - & $O(m)$ \\\midrule
    GNNML3~\citep{balcilar2021breaking}  & Spatial / Spectral & - & $O(m)$ \\\midrule
    MSGNN~\citep{he2022msgnn}  & Spectral Graph Convolution & - & $O(m)$ \\\midrule
    EGC~\citep{tailor2021we}  & Spatial Graph Convolution & General & $O(m)$ \\\midrule
    APPNP \citep{APPNP} & Spatial Graph Convolution & (Approximate) Personalized Pagerank & $O(m)$\\\midrule
    GCNII \citep{GCNII} &Spatial Graph Convolution & - & $O(m)$ \\\midrule
    GATv2 \citep{GATv2} & Spatial Graph Convolution & Attentive & $O(m)$\\
    \bottomrule
\end{tabular}
}
\end{table*}

\subsection{Spectral Graph Convolutions}
With the success of Convolutional Neural Networks (CNNs) in computer vision~\citep{krizhevsky2017imagenet}, efforts have been made to transfer the idea of convolution to the graph domain. However, this is not an easy task because of the non-Euclidean nature of graphical data. Graph signal processing (GSP)~\citep{shuman2013gsp, sandryhaila2013gsp, hammond2011wavelets} defines the Fourier Transform on graphs and thus provides a solid theoretical foundation of spectral graph convolutions.

In graph signal processing, a graph signal refers to a set of scalars associated with every node in the graph, \emph{i.e.} $f(v),~\forall v\in V$, and it can be written in the $n$-dimensional vector form $\mathbf{x} \in \mathbb R^{n}$, where $n$ is the number of nodes in the graph. Another core concept of graph signal processing is the symmetric normalized graph Laplacian matrix (or simply, the graph Laplacian), defined as $\mathbf L = \mathbf I - \mathbf D^{-1/2}\mathbf A \mathbf D^{-1/2}$, where $\mathbf I$ is the identity matrix, $\mathbf D$ is the degree matrix (\emph{i.e.} a diagonal matrix $\mathbf D_{ii} = \sum_j \mathbf A_{ij}$), and $\mathbf A$ is the adjacency matrix. In the typical setting of graph signal processing, the graph $G$ is undirected. Therefore, $\mathbf L$ is real symmetric and positive semi-definite. This guarantees the eigen decomposition of the graph Laplacian: $\mathbf L = \mathbf U\mathbf \Lambda \mathbf U^T$, where $\mathbf U = [\mathbf u_0, \mathbf u_1, ..., \mathbf u_{n-1}]$ is the eigenvectors of the graph Laplacian and the diagonal elements of $\mathbf \Lambda=\text{diag}(\lambda_0, \lambda_1, ..., \lambda_{n-1})$ are the eigenvalues. With this, the Graph Fourier Transform (GFT) of a graph signal $\mathbf x$ is defined as $\tilde{\mathbf x} = \mathbf U^T \mathbf x$, where $\tilde{\mathbf x}$ is the graph frequencies of $\mathbf x$. Correspondingly, the Inverse Graph Fourier Transform can be written as $\mathbf x = \mathbf U \tilde{\mathbf x}$.

With GFT and the Convolution Theorem, the graph convolution of a graph signal $\mathbf x$ and a filter $\mathbf g$ can be defined as $\mathbf g *_{G} \mathbf x = \mathbf U (\mathbf U^{T}\mathbf g \odot \mathbf U^{T} \mathbf x)$. To simplify this, let $\mathbf g_{\theta} = \text{diag}(\mathbf U^T g)$, the graph convolution can be written as:
\begin{equation}
    \mathbf g *_{G} \mathbf x = \mathbf U \mathbf g_{\theta} \mathbf U^T \mathbf x,
\end{equation}
which is the general form of most spectral graph convolutions. The key of spectral graph convolutions is to parameterize and learn the filter $\mathbf g_\theta$.

\emph{Spectral Convolutional Neural Network~(Spectral CNN)}~\citep{bruna2013spectral} sets graph filter as a learnable diagonal matrix $\mathbf W$. The convolution operation can be written as $\mathbf y = \mathbf U \mathbf W \mathbf U^T \mathbf x$. In practice, multi-channel signals and activation functions are common, and the graph convolution can be written as
\begin{equation}
\label{eq:spectral_cnn}
    \mathbf Y_{:, j} = \sigma\left(\mathbf U \sum_{i=1}^{c_{in}}\mathbf W_{i,j} \mathbf U^T \mathbf X_{:, i}\right), ~j=1,2,...,c_{out},
\end{equation}
where $c_{in}$ is the number of input channel, $c_{out}$ is the number of output channel, $\mathbf X$ is a $n\times c_{in}$ matrix representing the input signal, $\mathbf Y$ is a $n\times c_{out}$ matrix denoting the output signal, $\mathbf W_{i,j}$ is a parameterized diagonal matrix, and $\sigma(\cdot)$ is the activation function. For mathematical convenience we sometimes use single-channel versions of graph convolutions omitting activation functions, and the multi-channel versions are similar to Eq.~\ref{eq:spectral_cnn}. 

Spectral CNN has several limitations. Firstly, the filters are basis-dependent, which means that they cannot be generalized across graphs. Secondly, the algorithm requires eigen decomposition, which is computationally expensive. Thirdly, it has no guarantee of spatial localization of filters. To make filters spatially localized, \citet{henaff2015spectral} propose to use a smooth spectral transfer function $\Theta(\mathbf \Lambda)$ to parameterize the filter, and the convolution operation can be written as:
\begin{equation}
    \mathbf y = \mathbf U F(\mathbf \Lambda)\mathbf U^T \mathbf x.
\end{equation}

\emph{Chebyshev Spectral Convolutional Neural Network~(ChebNet)}~\citep{defferrard2016convolutional} extends this idea by using truncated Chebyshev polynomials to approximate the spectral transfer function. The Chebyshev polynomial is defined as $T_0(x)=1,~ T_1(x)=x,~ T_k(x)=2xT_{k-1}(x)-T_{k-2}(x)$, and the spectral transfer function $F(\mathbf \Lambda)$ is approximated to the order of $K-1$ as 
\begin{equation}
    F(\mathbf \Lambda) = \sum_{k=0}^{K-1}\theta_k T_k(\tilde{\mathbf \Lambda}),
\end{equation}
where the model parameters $\theta_k,~k\in\{0, 1, ..., K-1\}$ are the Chebyshev coefficients, and $\tilde{\mathbf \Lambda}=2\mathbf \Lambda / \lambda_{max}-\mathbf I$ is a diagonal matrix of scaled eigenvalues. Thus, the graph convolution can be written as:
\begin{equation}
\label{eq:chebnet}
    \mathbf g*_{G} \mathbf x = \mathbf U F(\mathbf \Lambda) \mathbf U^T \mathbf x = \mathbf U \sum_{k=0}^{K-1}\theta_k T_k(\tilde{\mathbf \Lambda}) \mathbf U^T \mathbf x =  \sum_{k=0}^{K-1} \theta_k T_k(\tilde{\mathbf L}) \mathbf x,
\end{equation}
where $\tilde{\mathbf L} = 2\mathbf L / \lambda_{max}-\mathbf I$.

\emph{Graph Convolutional Network~(GCN)}~\citep{kipf2016semi} is proposed as the localized first-order approximation of ChebNet. Assuming $K=2$ and $\lambda_{max} = 2$, Eq.~\ref{eq:chebnet} can be simplified as:
\begin{equation}
    \mathbf g*_{G} \mathbf x = \theta_0 \mathbf x + \theta_1 (\mathbf L - \mathbf I)\mathbf x = \theta_0 \mathbf x - \theta_1 \mathbf D^{-1/2}\mathbf A \mathbf D^{-1/2}\mathbf x.
\end{equation}

To further constraint the number of parameters, we assume $\theta = \theta_0 = -\theta_1$, which gives a simpler form of graph convolution:
\begin{equation}
    \mathbf g*_{\mathcal G} \mathbf x = \theta (\mathbf I + \mathbf D^{-1/2}\mathbf A \mathbf D^{-1/2})\mathbf x.
\end{equation}

As $\mathbf I + \mathbf D^{-1/2}\mathbf A \mathbf D^{-1/2}$ now has the eigenvalues in the range of $[0, 2]$ and repeatedly multiplying this matrix can lead to numerical instabilities, GCN empirically proposes a renormalization trick to solve this problem by using $\mathbf{\tilde{D}}^{-1/2} \mathbf{\tilde{A}} \mathbf{\tilde{D}}^{-1/2}$ instead, where $\mathbf{\tilde{A}} = \mathbf A + \mathbf I$ and $\mathbf{\tilde{D}}_{ii} = \sum_i \mathbf{\tilde{A}}_{ij}$.

Allowing multi-channel signals and adding activation functions, the more common formula in literature is:
\begin{equation}
\label{eq:gcn_mat}
    \mathbf Y = \sigma((\mathbf{\tilde{D}}^{-1/2} \mathbf{\tilde{A}} \mathbf{\tilde{D}}^{-1/2})\mathbf X\mathbf \Theta),
\end{equation}
where $\mathbf X$, $\mathbf Y$ have the same shape as in Eq.~\ref{eq:spectral_cnn} and $\mathbf \Theta$ is a $c_{in} \times c_{out}$ matrix as model's parameters.

Apart from the aforementioned methods, other spectral graph convolutions have been proposed. \citet{levie2018cayleynets} propose CayleyNets that utilize Cayley Polynomials to equip the filters with the ability to detect narrow frequency bands. \citet{liao2019lanczosnet} propose LanczosNets that employ the Lanczos algorithm to construct a low-rank approximation of graph Laplacian to improve the computation efficiency of graph convolutions. The proposed model is able to efficiently utilize the multi-scale information in the graph data. Instead of using Graph Fourier Transform, \citet{xu2019graph} propose a Graph Wavelet Neural Network (GWNN) that uses graph wavelet transform to avoid matrix eigendecomposition. Moreover, graph wavelets are sparse and localized, which provides good interpretations for the convolution operation. \citet{zhu2021simple} derive a Simple Spectral Graph Convolution (S$^2$GC) from a modified Markov Diffusion Kernel, which achieves a trade-off between low-pass and high-pass filter bands.

\subsection{Spatial Graph Convolutions}
Inspired by the convolution on Euclidean data (\emph{e.g.} images and texts), which applies data transformation on a small region, spatial graph convolutions compute the central node's feature via transforming and aggregating its neighbors' features. In this way, the graph structure is naturally embedded in the computation graph of node features. Moreover, the idea of sending one node's feature to another node is similar to the message passing used in 
recurrent graph neural networks. In the following, we will introduce several seminal spatial graph convolutions as well as some recently proposed promising methods.

Spatial graph convolutions generally follow a three-step paradigm: message generation, feature aggregation and feature update. This can be mathematically written as:
\begin{equation}
    \mathbf y_i = \text{UPDATE}\left(\mathbf x_i, \text{AGGREGATE}\left(\{\text{MESSAGE}\left(\mathbf x_i, \mathbf x_j, \mathbf e_{ij}\right), ~j\in \mathcal N(i)\}\right)\right),
\end{equation}
where $\mathbf x_i$ and $\mathbf y_i$ is the input and output feature vector of node $i$, $\mathbf e_{ij}$ is the feature vector of the edge (or more generally, the relationship) between node $i$ and its neighbor node $j$, and $\mathcal N(i)$ denote the neighbor of node $i$, which could be more generally defined.

In the previous subsection, we show the spectral interpretation of GCN~\citep{kipf2016semi}. The model also has its spatial interpretation, which can be mathematically written as:
\begin{equation}
    \mathbf y_i = \mathbf \Theta^T \sum_{j\in \mathcal N(i)\cup {i}}\frac 1 {\sqrt{\hat d_i \hat d_j}}\mathbf x_j,
\end{equation}
where $\hat d_i$ and $\hat d_j$ is the $i$-th and $j$-th row sums of $\hat{\mathbf A}$ in Eq.~\ref{eq:gcn_mat}. For each node, the model takes a weighted sum of its neighbors' features as well as its own features and applies a linear transformation to obtain the result. In practice, multiple GCN layers are often stacked together with non-linear functions after convolution to encode complex and hierarchical features. Nonetheless, \citet{wu2019simplifying} show that the model still achieves competitive results without non-linearity.

Although GCN as well as other spectral graph convolutions achieve competitive results on a number of benchmarks, these methods assume the presence of all nodes in the graph and fall in the category of transductive learning. \citet{hamilton2017inductive} propose GraphSAGE that performs graph convolutions in inductive settings, when there are new nodes during inference (\emph{e.g.} newcomers in the social network). For each node, the model samples its $K$-hop neighbors and uses $K$ graph convolutions to aggregate their features hierarchically. Furthermore, the use of sampling also reduces the computation when a node has too many neighbors.

The attention mechanism has been successfully used in natural language processing~\citep{vaswani2017attention}, computer vision~\citep{ liu2021swin} and multi-modal tasks~\citep{td-stp, hamt, transrefer3d, multimodal-coattention}. Graph Attention Networks (GAT)~\citep{velivckovic2017graph} introduces the idea of attention to graphs. The attention mechanism uses an adaptive, feature-dependent weight (\emph{i.e.} attention coefficient) to aggregate a set of features, which can be mathematically written as:
\begin{equation}
    \alpha_{i,j} = \frac {\exp\left({\text{LeakyReLU}\left(\mathbf a^T[\mathbf \Theta \mathbf x_i || \mathbf \Theta \mathbf x_j]\right)}\right)} {\sum_{k \in \mathcal N(i)\cup \{i\}} \exp\left({\text{LeakyReLU}\left(\mathbf a^T[\mathbf \Theta \mathbf x_i || \mathbf \Theta \mathbf x_j]\right)}\right)},
\end{equation}
where $\alpha_{i,j}$ is the attention coefficient, $\mathbf a$ and $\mathbf \Theta$ are model parameters, and $[\cdot||\cdot]$ means concatenation. After the $\alpha$s are obtained, the new features are computed as a weighted sum of input node features, which is:
\begin{equation}
    \mathbf y_i = \alpha_{i,i}\mathbf \Theta \mathbf x_i + \sum_{j\in \mathcal N(i)} \alpha_{i,j} \mathbf \Theta \mathbf x_j.
\end{equation}

\citet{xu2018powerful} explore the representational limitations of graph neural networks. What they discover is that message passing networks like GCN~\citep{kipf2016semi} and GraphSAGE~\citep{hamilton2017inductive} are incapable of distinguishing certain graph structures. To improve the representational power of graph neural networks, they propose the Graph Isomorphism Network (GIN) that gives an adjustable weight to the central node feature, which can be mathematically written as:
\begin{equation}
    \mathbf y_i = \text{MLP} \left((1+\epsilon)\mathbf x_i + \sum_{j\in \mathcal N(i)}\mathbf x_j\right),
\end{equation}
where $\epsilon$ is a learnable parameter.

More recently, efforts have been made to improve the representational power of graph neural networks. For example, \citet{hu2021graphair} propose GraphAIR that explicitly models the neighborhood interaction to better capture complex non-linear features. Specifically, they use the Hadamard product between pairs of nodes in the neighborhood to model the quadratic terms of neighborhood interaction. \citet{balcilar2021breaking} propose GNNML3 that breaks the limits of the first-order Weisfeiler-Lehman test (1-WL) and reaches the third-order WL test (3-WL) experimentally. They also show that the Hadamard product is required for the model to have more representational power than the first-order Weisfeiler-Lehman test. Other elements in spatial graph convolutions are widely studied. For example, \citet{corso2020principal} explore the aggregation operation in GNN and proposes Principal Neighbourhood Aggregation (PNA) that uses multiple aggregators with degree-scalers. \citet{tailor2021we} explore the anisotropism and isotropism in the message passing process of graph neural networks, and proposes Efficient Graph Convolution (EGC) that achieves promising results with reduced memory consumption due to isotropism. In order to increase the size of the neighborhood of a node, \citet{APPNP} propose personalized propagation of neural predictions (PPNP) and its approximation using power iteration (APPNP). To increase the depth of graph neural networks, \citet{GCNII} propose GCNII that uses initial residual and identity mapping to mitigate the over-smoothing problem. \citet{GATv2} propose GATv2 that uses dynamic attention and improves the expressive power of GAT \citep{velivckovic2017graph}.

\subsection{Summary}

This section introduces graph convolutions. We provide the summary as follows:

\begin{itemize}
    \item \textbf{Techniques.} Graph convolutions mainly fall into two types, \emph{i.e.} spectral graph convolutions and spatial graph convolutions. Spectral graph convolutions have solid mathematical foundations of Graph Signal Processing and therefore their operations have theoretical interpretations. Spatial graph convolutions are inspired by Recurrent Graph Neural Networks and their computation is simple and straightforward, as their computation graph is derived from the local graph structure. Generally, spatial graph convolutions are more common in applications.
     \item \textbf{Challenges and Limitations.} Despite the great success of graph convolutions, their performance is unsatisfactory in more complicated applications. On the one hand, the performance of graph convolutions relies heavily on the construction of the graph. Different constructions of the graph might result in different performances of graph convolutions. On the other hand, graph convolutions are prone to over-smoothing when constructing very deep neural networks.
    \item \textbf{Future Works.} In the future, we expect that more powerful graph convolutions will be developed to mitigate the problem of over-smoothing and we also hope that techniques and methodologies in Graph Structure Learning (GSL) can help learn more meaningful graph structure to benefit the performance of graph convolutions.
\end{itemize}

\section{Graph Kernel Neural Networks}
Graph kernels (GKs) are historically the most widely used technique
on graph analyzing and representation tasks \citep{gartner2003graph,zhou2020graph,krishnagopal2023graph,song2020deep}. However, traditional graph kernels rely on hand-crafted patterns or domain knowledge on specific tasks\citep{kriege2020survey,shervashidze2009efficient}. 
Over the years, an amount of research has been conducted on graph kernel neural networks (GKNNs), which has yielded promising results. Researchers have explored various aspects of GKNNs, including their theoretical foundations, algorithmic design, and practical applications. These efforts have led to the development of a wide range of GKNN-based models and methods that can be used for graph analysis and representation tasks, such as node classification ~\citep{yang2023comprehensive,fang2022polarized,ju2024survey,long2024unveiling}, link prediction ~\citep{chen2020convolutional,long2019hierarchical,wu2023turbomgnn,yan2024inductive}, and graph clustering ~\citep{krishnagopal2023graph,long2021theoretically}.

The success of GKNNs can be attributed to their ability to leverage the strengths of both graph kernels and neural networks \citep{wu2023turbomgnn,long2021theoretically,ju2022kgnn}. By using kernel functions to measure similarity between graphs, GKNNs can capture the structural properties of graphs, while the use of neural networks enables them to learn more complex and abstract representations of graphs \citep{chen2022structure,zang2023hierarchical}. This combination of techniques allows GKNNs to achieve state-of-the-art performance on a wide range of graph-related tasks \citep{krishnagopal2023graph,ju2022ghnn,wang2022imbalanced}. 

In this section, we begin with introducing the most representative traditional graph kernels. Then we summarize the basic framework for combining GNNs and graph kernels. Finally, we categorize the popular graph kernel Neural networks into several categories and compare their differences.

\subsection{Graph Kernels}
Graph kernels generally evaluate pairwise similarity between nodes or graphs by decomposing them into basic structural units. Random walks \citep{kang2012fast}, subtrees \citep{she2011weisfeiler}, shortest paths \citep{borgwardt2005shortest} and graphlets \citep{shervashidze2009efficient} are representative categories.

Given two graphs $G_1 = (V_1, E_1, X_1)$ and $G_2 = (V_2, E_2, X_2)$, a graph kernel function $K(G_1, G_2)$ measures the similarity between $G_1$ and $G_2$ through the following formula: 
\begin{equation}
    K(G_1, G_2) = \sum_{u_1\in V_1}\sum_{u_2\in V_2} \kappa_{base}\left(l_{G_1}(u_1), l_{G_2}(u_2)\right),
    \label{eqn:gk}
\end{equation}
where $l_{G}(u)$ denotes a set of local substructures centered at node $u$ in graph $G$, and $\kappa_{base}$ is a base kernel measuring the similarity between the two sets of substructures. For simplicity, we may rewrite Eq. \ref{eqn:gk} as:
\begin{equation}
    K(G_1, G_2) = \sum_{u_1\in V_1}\sum_{u_2\in V_2} \kappa_{base}(u_1, u_2),
    \label{eqn:gk_sim}
\end{equation}
the uppercase letter $K(G_1, G_2)$ is denoted as graph kernels, $\kappa(u_1, u_2)$ is denoted as node kernels, and lowercase $k(x, y)$ is denoted as general kernel functions. 

The kernel mapping of a kernel $\psi$ maps a data point into its corresponding Reproducing Kernel Hilbert Space (RKHS) $\mathcal{H}$. Specifically, given a kernel $k_{*}(\cdot, \cdot)$, its kernel mapping $\psi_{*}$ can be formalized as,
\begin{equation}
    \forall x_1, x_2, k_*(x_1, x_2) = \langle \psi_*(x_1), \psi_*(x_2)\rangle_{\mathcal{H}_{*}},
    \label{eqn:RKHS}
\end{equation}
where $\mathcal{H}_*$ is the RKHS of $k_*(\cdot, \cdot)$.

We introduce several representative and popular graph kernels below. 

\smallskip
\emph{Walk and Path Kernels.} A $l$-walk kernel $K_{walk}^{(l)}$ compares all length $l$ walks starting from each node in two graphs $G_1, G_2$,
\begin{equation}
    \begin{aligned}
        \kappa_{walk}^{(l)}(u_1, u_2) &= \sum_{w_1\in \mathcal{W}^l (G_1, u_1)}\sum_{w_2\in \mathcal{W}^l(G_2, u_2)}\delta(X_1(w_1), X_2(w_2)),\\
        K_{walk}^{(l)}(G_1, G_2) &= \sum_{u_1\in V_1}\sum_{u_2\in V_2} \kappa_{walk}^{(l)}(u_1, u_2).
    \end{aligned}
    \label{eqn:walk_gk}
\end{equation}

Substituting $\mathcal{W}$ with $\mathcal{P}$ is able to get the $l$-path kernel. 

\smallskip
\emph{Subtree Kernels.} The WL subtree kernel is the most popular one in subtree kernels. It is a finite-depth kernel variant of the 1-WL test. The WL subtree kernel with depth $l$, $K_{WL}^{(l)}$ compares all subtrees with depth $\le l$ rooted at each node. 
\begin{equation}
    \begin{aligned}
        \kappa_{subtree}^{(i)}(u_1, u_2) &= \sum_{t_1 \in \mathcal{T}^i(G_1, u_2)}\sum_{t_2\in \mathcal{T}^i(G_2, u_2)} \delta(t_1, t_2),\\
        K_{subtree}^{(i)}(G_1, G_2) &= \sum_{u_1\in V_1}\sum_{u_2\in V_2} \kappa_{subtree}^{(i)}(u_1, u_2),\\
        K_{WL}^{(l)}(G_1, G_2) &= \sum_{i=0}^{l} K_{subtree}^{(i)}(G_1, G_2),\\
    \end{aligned}
    \label{eqn:wl-kernel}
\end{equation}
where $t\in \mathcal{T}^{(i)}(G, u)$ denotes a subtree of depth $i$ rooted at $u$ in $G$. 

\subsection{General Framework of GKNNs}
In this section, we summarize the general framework of GKNNs. For the first step, a kernel that measures similarities of heterogeneous features from heterogeneous nodes and edges $(u_1, e_{\cdot,u_2})$ and $(u_2, e_{\cdot,u_2})$ should be defined. Take the inner product of neighbor tensors as an example, its neighbor kernel is defined as follows,
$$\kappa ((u_1,e_{\cdot,u_1}),(u_2,e_{\cdot,u_2}))=\langle f(u_1),f(u_2)\rangle\cdot \langle f(e_{\cdot,u_1}),f(e_{\cdot,u_2})\rangle.$$

Based on the neighbor kernel, a kernel with two $l$-hop neighborhoods for central node $u_1$ and $u_2$ can be defined as 
$K^{(l)}(u_1,u_2)=$
\begin{equation}\label{eq:kernel_ori}
  \left\{
  \begin{aligned}
 & \langle f(u_1),f(u_2)\rangle &\ l = 0 \\
 & \langle f(u_1),f(u_2)\rangle \cdot \sum_{v_1\in N(u_1)}\sum_{v_2\in N(u_2)} 
K^{(l-1)} (v_1, v_2)\cdot \langle f(e_{\cdot,v_1}),f(e_{\cdot,v_2})\rangle &\ l > 0
  \end{aligned}
  \right.,
\end{equation}

By regarding the lower-hop kernel $\kappa^{(l-1)}(u_1,u_2)$, as the inner product of the $(l-1)$-th hidden representations of $u_1$ and $u_2$. Furthermore, by recursively
applying the neighborhood kernel, the $l$-hop graph kernel can be derived as
\begin{equation}
K^{l}(G_1,G_2) = \sum_{\boldsymbol{w}_1\in \mathcal{W}^l(G_1)}\sum_{\boldsymbol{w}_2\in \mathcal{W}^l(G_2)}\left (\prod_{i=0}^{l-1}\langle f(\boldsymbol{w}_1^{(i)}),f(\boldsymbol{w}_2^{(i)})\rangle \times\prod_{i=0}^{l-2}\langle f(e_{\boldsymbol{w}_1^{(i)},\boldsymbol{w}_1^{(i+1)}}),f(e_{\boldsymbol{w}_2^{(i)},\boldsymbol{w}_2^{(i+1)}})\rangle \right ),
\label{eq:gk}
\end{equation}
where $\mathcal{W}^l(G)$ denotes the set of all walk sequences with length $l$ in graph $G$, and $\boldsymbol{w}_1^{(i)}$ denotes the $i$-th node in sequence $\boldsymbol{w}_1$.

As shown in Eq. \ref{eqn:RKHS}, kernel methods implicitly perform projections from original data spaces to their RKHS $\mathcal{H}$. Hence, as GNNs also project nodes or graphs into vector spaces, connections have been established between GKs and GNNs through the kernel mappings. And several works conducted research on the connections \citep{lei2017deriving,williams2001using}, and found some foundation conclusions. Take the basic rule introduced in  \citep{lei2017deriving} as an example, the proposed graph kernel in Eq. \ref{eqn:gk} can be derived as the general formulas,
\begin{equation}
  \begin{aligned}
 h^{(0)}(v) =& \boldsymbol{W}^{(0)}_{t_V(v)}f(v),\\
 h^{(l)}(v) =& \boldsymbol{W}^{(l)}_{t_V(v)}f(v) \odot \sum_{u\in N(v)}(\boldsymbol{U}_{t_V(v)}^{(l)}h^{(l-1)}(u)  \odot\boldsymbol{U}_{t_E(e_{u,v})}^{(l)}f(e_{u,v})), \qquad 1 < l \leq L,
  \end{aligned}
\end{equation}
where $\odot$ is the element-wise product and $h^{(l)}(v)$ is the cell state vector of node v.
The parameter matrices $\boldsymbol{W}^{(l)}_{t_V(v)}$, $\boldsymbol{U}_{t_V(v)}^{(l)}$ and $\boldsymbol{U}_{t_E(e_{u,v})}^{(l)}$ are learnable parameters related to types of nodes and edges. 

Then mean embeddings of all nodes are usually used to represent the graph-level embedding, let $|G_i|$ denote the number of nodes in the $i$-th graph, then the graph-level embeddings are generated as,
\begin{equation}
    \Phi(G_i) = \sum_{u\in G_i}\frac{1}{|G_i|}h^{(L)}(v).
    \label{eq:graph_embedding}
\end{equation}

For semi-supervised multiclass classification, the cross entropy is used as the objective function over all training examples \citep{cao2016deep,cao2015grarep},
\begin{equation}
    \mathcal{L}=-\sum_{l\in \mathcal{Y}_L}\sum_{f=1}^{F}Y_{lf}\ln{Z_{lf}},
\label{eq:obj}
\end{equation}
where $\mathcal{Y}_L$ is the set of node indices that have labels in node classification tasks or the set of graph indices in graph classification tasks.
$Z_{lf}$ denotes the prediction of labels, which are outputs of a linear layer with an activation function, inputting $h^{(l)}(v)$ in the node classification task, and $\Phi(G_i)$ in the graph classification task.

\subsection{Popular Variants of GKNNs}
We summarize the popular variants of GKNNs and compare their differences in Table \ref{tab:gknn}. Specifically, we conclude their basic graph kernels, whether designed for heterogeneous graphs, experimental datasets, etc. As the originally designed graph-kernel based GNNs have high complexity, they usually by acceleration strategies, such as sampling strategies, simplification and approximation, etc. In this section, We select four typical and popular  GKNNs to introduce their well-designed graph kernels and corresponding GNN frameworks.

\begin{table*}[t]
\centering
\caption{Summary of popular GKNNs. }
\label{tab:gknn}
\resizebox{0.85\linewidth}{!} {
\begin{tabular}{l cccc}
    \toprule
    Method  & Type & Related GK & Adaptive & Datasets \\
    \midrule
     $k$-GNN ~\citep{morris2019weisfeiler}  & Isomorphic & WL Subtree &  \checkmark  & \makecell{Biochemical Network}  \\\midrule
    RetGK~\citep{zhang2018retgk}  & Isomorphic & Random Walk & \checkmark &  \makecell{Biochemical Network, \\Social Network} \\\midrule
    GNTK~\citep{du2019graph}  & Isomorphic & Neural Tangent Kernel & & \makecell{Biochemical Network, \\Social Network} \\\midrule
    DDGK~\citep{al2019ddgk}  & Isomorphic & Random Walk &  &  Biochemical Network\\\midrule
    GCKN~\citep{chen2020convolutional}  & Isomorphic & Random Walk &  \checkmark  &  \makecell{Biochemical Network, \\Social Network} \\\midrule
    GSKN~\citep{long2021theoretically}  & Isomorphic &    \makecell{Random Walk, \\ Anonymous Walk}& \checkmark  & \makecell{Biochemical Network, \\Social Network} \\\midrule
    KerGNNs~\citep{feng2022kergnns} & Isomorphic &  Random Walk & \checkmark   & \makecell{Biochemical Network, \\Academic Network} \\\midrule
    TurboMGNN ~\citep{wu2023turbomgnn}  &  Isomorphic & Random Walk & \checkmark & \makecell{Social Network} \\
    \midrule
   GCN-LASE~\citep{gcnlase}  & Heterogeneous & Random Walk &  &  \makecell{Social Network\\Academic Network} \\\midrule
   HGK-GNN~\citep{long2021hgk}  & Heterogeneous & Random Walk & \checkmark & \makecell{Social Network\\Academic Network} \\
   
    \bottomrule
\end{tabular}}
\end{table*}

\smallskip
\emph{$k$-dimensional Graph Neural Networks}~(\emph{$k$-GNN)}~\citep{morris2019weisfeiler} is the pioneer of GKNNs, it incorporates the WL-subtree graph kernel and graph neural networks.
For better scalability, the paper considers a set-based
version of the $k$-WL. Let $h^{(l)}_k(s)$ and $h^{(l)}_{k,L}(s)$ denote the \textit{global} and \textit{local}  hidden representation for node $s$ in the $l$-th layer respectively. $k$-GNN defined the end-to-end hierarchical trainable framework as, 
\begin{equation}
\begin{aligned}
 h^{(0)}_k(s) =& \sigma\left (\left[ h^{iso}(s),\sum_{u\in s}  h^{(T_k-1)}(u)\right]\cdot\boldsymbol{W}^{(0)}_{k-1}\right ),\\
 h^{(l)}_k(s) =& \sigma\left ( h^{(l-1)}_k\cdot \boldsymbol{W}_1^{(l)}+\sum_{u\in N_L(s)\cup N_G(s)} h^{(l-1)}_k(u)\cdot\boldsymbol{W}_2^{(l)}\right ),\qquad 1 < l \leq L,\\
  h^{(l)}_{k,L}(s) =& \sigma\left ( h^{(l-1)}_{k,L}(s)\cdot \boldsymbol{W}_1^{(l)}+\sum_{u\in N_L(s)} h^{(l-1)}_{k,L}(u)\cdot \boldsymbol{W}_2^{(l)}\right ),\qquad 1 < l \leq L.
\end{aligned}
\end{equation}
where $h^{iso}(s)$ is the one-hot encoding of the isomorphism type of $G[s]$, $N_L(s)$ is the local neighborhood, $N_G(s)$ is the global neighborhood, 
$h^{(l)}_{k,L}(s)$ is designed for better scalability and running speed, and $h^{(l)}_k(s)$ has better performance due to its larger neighbor sets.

\smallskip
\emph{Graph Convolutional Kernel Networks}~\emph{(GCKN)}~\citep{chen2020convolutional}. GCKN is a representative random walk and path-based GKNN. 
The Gaussian kernel $k$ can be written as,
\begin{equation}
k(z_1, z_2)=e^{-\frac{\alpha_1}{2}||z_1-z_2||^2}=e^{\alpha(z_1^Tz_2-k-1)}=\sigma(z_1^Tz_2),
\end{equation}
then the GNN architecture can be written as
\begin{equation}
\begin{aligned}
h^{(l)} (u)=&\sum_{z\in}K(z_1,z_2)\cdot\sigma\left (Z^T h^{(l-1)}(p)\right )\\
=&\sigma\left (ZZ^T\right )^{-\frac{1}{2}}\cdot\sum_{p\in \mathcal{P}_k(G,u)}\sigma\left (Z^T h^{(l-1)}(p)\right ),
\end{aligned}
\end{equation}
where $Z$ is the matrix of prototype path attributes.

Furthermore, the paper analyzes the relationship between GCKN and the WL-subtree based $k$-GNN. Theorem 1 in paper~\citep{chen2020convolutional} shows that WL-subtree based GKNNs can be seen as a special case in GCKN.

\smallskip
\emph{Graph Neural Tangent Kernel}~\emph{(GNTK)}~\citep{du2019graph}. 
Different from the above two works, GNTK proposed a new class of graph kernels. GNTK is a general recipe which translates a GNN architecture to its corresponding GNTK. 

The neighborhood aggregation operation in GNTK is defined as,
\begin{equation}
\begin{aligned}
\left[\sum_{(0)}^{(l)}(G,G^{'})\right ]_{uu^{'}} =& c_uc_{u^{'}}\sum_{u\in N(v)\cup \{u\}}\sum_{v^{'}\in N(u^{'})\cup \{u^{'}\}}\left [ \sum_{R}^{(l-1)}(G,G^{'}) \right ]_{vv^{'}},\\
\left[\Theta_{(0)}^{(l)}(G,G^{'})\right ]_{uu^{'}} =& c_uc_{u^{'}}\sum_{u\in N(v)\cup \{u\}}\sum_{v^{'}\in N(u^{'})\cup \{u^{'}\}}\left [ \Theta_{R}^{(l-1)}(G,G^{'}) \right ]_{vv^{'}},
\end{aligned}
\end{equation}
where $\sum_{R}^{(0)}(G,G^{'})$ and $\Theta_{R}^{(0)}(G,G^{'})$ are defined as $\sum^{(0)}(G,G^{'})$. Then GNTK performed $R$ transformations to capture high-order structural information.
\begin{equation}
\left[A_{(r)}^{(l)}(G,G^{'})\right ]_{uu^{'}} =\left ( 
\begin{aligned}
  &\left [\sum_{(r-1)}^{(l)}(G,G) \right]_{uu^{'}}&,\ & \left[\sum_{(r-1)}^{(l)}(G,G^{'}) \right]_{uu^{'}} \\
  &\left[\sum_{(r-1)}^{(l)}(G^{'},G) \right]_{uu^{'}}&,\ &\left[\sum_{(r-1)}^{(l)}(G^{'},G^{'}) \right]_{uu^{'}}
  \end{aligned}
  \right ).
\end{equation}
\begin{equation}
\begin{aligned}
\left[\sum_{(r)}^{(l)}(G,G^{'})\right ]_{uu^{'}} =& \mathbb{E}_{(a,b)\sim \mathcal{N}(0,[A_{(r)}^{(l)}(G,G^{'})]_{uu^{'}})}[\sigma(a)\cdot\sigma(b)],\\
\left[\sum_{(r)}^{(l)}(G,G^{'})\right ]_{uu^{'}} =& \mathbb{E}_{(a,b)\sim \mathcal{N}(0,[A_{(r)}^{(l)}(G,G^{'})]_{uu^{'}})}[\sigma(a)\cdot\sigma(b)],
\end{aligned}
\end{equation}
then the $r$ order can be calculated as,
\begin{equation}
\left[\Theta_{(r)}^{(l)}(G,G^{'})\right ]_{uu^{'}} = \left[\Theta_{(r-1)}^{(l)}(G,G^{'})\right ]_{uu^{'}}\left[\sum{(r)}^{(l)}(G,G^{'})\right ]_{uu^{'}}+\left[\sum{(r)}^{(l)}(G,G^{'})\right ]_{uu^{'}}.
\end{equation}

Finally, GNTK calculates the final output as,
\begin{equation}
\Theta (G,G^{'})=\sum_{u\in V,u^{'}\in V^{'}}\left [ \sum_{l=0}^L\Theta_{(R)}^{(l)}(G,G^{'})\right ]_{u,u^{'}}.
\end{equation}

\smallskip
\emph{Heterogeneous Graph Kernel based Graph Neural Network}~\emph{(HGK-GNN)}~\citep{long2021hgk}. HGK-GNN first proposed GKNN for heterogeneous graphs. It adopted $\langle f(u_1),f(u_2)\rangle_M$ as graph kernel based on the Mahalanobis Distance to build connections among heterogeneous nodes and edges,
$$\langle f(u_1),f(u_2)\rangle_{M_1}=f(u_1)^T\boldsymbol{M}_1f(u_2),$$
$$\langle f(e_{\cdot,u_1}),f(e_{\cdot,u_2})\rangle_{M_2}=f(e_{\cdot,u_1})^T\boldsymbol{M}_2f(e_{\cdot,u_2}).$$

Following the route introduced in \citet{lei2017deriving} , the corresponding neural network architecture of heterogeneous graph kernel can be derived as 
\begin{equation}
\begin{aligned}
 h^{(0)}(v) =& \boldsymbol{W}^{(0)}_{t_V(v)}f(v),\\
 h^{(l)}(v) =& \boldsymbol{W}^{(l)}_{t_V(v)}f(v) \odot \sum_{u\in N(v)}(\boldsymbol{U}_{t_V(v)}^{(l)}h^{(l-1)}(u)  \odot\boldsymbol{U}_{t_E(e_{u,v})}^{(l)}f(e_{u,v})), \qquad 1 < l \leq L,
\end{aligned}
\end{equation}
where $h^{(l)}(v)$ is the cell state vector of node v, and $\boldsymbol{W}^{(l)}_{t_V(v)}$, $\boldsymbol{U}_{t_V(v)}^{(l)}$ $\boldsymbol{U}_{t_E(e_{u,v})}^{(l)}$ are learnable parameters.

\subsection{Summary}

This section introduces graph kernel neural networks. We provide the summary as follows:

\begin{itemize}
    \item \textbf{Techniques.} 
    Graph kernel neural networks (GKNNs) are a recent popular research area that combines the advantages of graph kernels and GNNs to learn more effective graph representations. Researchers have studied GKNNs in various aspects, such as theoretical foundations, algorithmic design, and practical applications. As a result, a wide range of GKNN-based models and methods have been developed for graph analysis and representation tasks, including node classification, link prediction, and graph clustering. 

    \item \textbf{Challenges and Limitations.} Although GKNNs have shown great potential in graph-related tasks, they also have several limitations that need to be addressed.  Scalability is a significant challenge, particularly when dealing with large-scale graphs and networks. As the size of the graph increases, the computational cost of GKNNs grows exponentially, which can limit their ability to handle large and complex real-world applications.
    \item \textbf{Future Works.} For future works, we expect the GKNNs can integrate more domain-specific knowledge into the designed kernels. Domain-specific knowledge has been shown to significantly improve the performance of many applications, such as drug discovery~\citep{zeng2023substructure}, knowledge graph-based information retrieval systems, and molecular analysis \citep{zang2023hierarchical,wang2019multi,feinberg2018potentialnet}.  Incorporating domain-specific knowledge into GKNNs can enhance their ability to handle complex and diverse data structures, leading to more accurate and interpretable models.
\end{itemize}


\section{Graph Pooling}  

When it comes to graph-level tasks, such as graph classification and graph regression, graph pooling is an essential component for generating the whole graph representation from the learned node embeddings. To ensure isomorphic graphs have the same representation, the graph pooling operations should be invariant to the permutations of nodes. In this section, we give a systematic review of existing graph pooling algorithms and generally classify them into two categories: global pooling algorithms and hierarchical pooling algorithms. The global pooling algorithms aggregate the node embeddings to the final graph representation directly, while the hierarchical pooling algorithms reduce the graph size and generate the immediate representations gradually to capture the hierarchical structure and characteristics of the input graph. A summary is provided in Table \ref{tab:graphpooling}.

\subsection{Global Pooling}
Global pooling operations generate a holistic graph representation from the learned node embeddings in one step, which are referred to as readout functions in some literature \citep{xu2018powerful,corso2020principal} as well. Several simple permutation-invariant operations, such as mean, sum, and max, are widely employed on the node embeddings to output the graph-level representation. To enhance the adaptiveness of global pooling operators, GGS-NN \citep{li2016gated} introduces a soft attention mechanism to evaluate the importance of nodes for a particular task and then takes a weighted sum of the node embeddings. 
SortPool \citep{zhang2018end} exploits Weisfeiler-Lehman methods \citep{weisfeiler1968reduction} to sort nodes based on their structural positions in the graph topology, and produces the graph representation from the sorted node embeddings by traditional convolutional neural networks. Recently, SSRead~\citep{lee2021learnable} proposes a structural-semantic pooling method, which first aligns nodes and learnable structural prototypes semantically, and then aggregates node representations in groups based on matching structural prototypes.

\begin{table*}[t]
\centering
\caption{Summary of graph pooling methods. }
\label{tab:graphpooling}
\resizebox{0.8\textwidth}{!}{
\begin{tabular}{l cccc}
    \toprule
    Method  & Type & TopK-based & Cluster-based & Attention Mechanism \\
    \midrule
    Mean/Sum/Max  & \multirow{2}{*}{Global} & \multirow{2}{*}{} & \multirow{2}{*}{} & \multirow{2}{*}{}\\
    Pooling   &  &  &  &\\\midrule
     GGS-NN \citep{li2016gated}  & Global &  &  &  \checkmark \\\midrule
     SortPool \citep{zhang2018end}  & Global &  &  &  \\\midrule
     SSRead \citep{lee2021learnable}  & Global &  &  &  \\\midrule
     gPool~\citep{gao2019graph}  & Hierarchical & \checkmark &  &  \\\midrule
     SAGPool~\citep{lee2019self}  & Hierarchical & \checkmark &  & \checkmark \\\midrule
     HGP-SL~\citep{zhang2019hierarchical}  & Hierarchical & \checkmark &  & \checkmark \\\midrule
     TAPool~\citep{gao2021topology}  & Hierarchical & \checkmark &  &  \\\midrule
     DiffPool~\citep{ying2018hierarchical}  & Hierarchical &  & \checkmark &  \\\midrule
     MinCutPool~\citep{bianchi2020spectral}  & Hierarchical &  & \checkmark &  \\\midrule
     SEP~\citep{wu2022structural}  & Hierarchical &  & \checkmark &  \\\midrule
     ASAP~\citep{ranjan2020asap} & Hierarchical & \checkmark & \checkmark & \checkmark \\\midrule
     MuchPool~\citep{ijcai2021p199}  & Hierarchical & \checkmark & \checkmark &  \\
    \bottomrule
\end{tabular}
}
\end{table*}

\subsection{Hierarchical Pooling}
Different from global pooling methods, hierarchical pooling methods coarsen the graph gradually, to preserve the structural information of the graph better. To adaptively coarse the graph and learn optimal hierarchical representations according to a particular task, many learnable hierarchical pooling operators have been proposed in the past few years, which can be integrated with multifarious graph convolution layers. There are two common approaches to coarsening the graph, one is selecting important nodes and dropping the others by TopK selection, and the other one is merging nodes and generating the coarsened graph by clustering methods. We call the former \textit{TopK-based pooling}, and the latter \textit{cluster-based pooling} in this survey. In addition, some works combine these two types of pooling methods, which will be reviewed in the \textit{hybrid pooling} section.

\subsubsection{TopK-based Pooling} 
Typically, TopK-based pooling methods first learn a scoring function to evaluate the importance of nodes of the original graph. Based on importance score $\mathbf{Z} \in \mathbb{R}^{|V| \times 1}$ generated, they select the top $K$ nodes out of all nodes,
\begin{equation}
    idx=\operatorname{TOP}_k\left(\mathbf{Z}\right),
\end{equation}
where $idx$ denotes the index of the top $K$ nodes. Based on these selected nodes, most methods directly employ the induced subgraph as the pooled graph, 
\begin{equation}
    \mathbf{A}_{\text{pool}}=\mathbf{A}_{idx, idx},
\end{equation}
where $A_{idx, idx}$ denotes the adjacency matrix indexed by the selected rows and columns. Moreover, to make the scoring function learnable, they further use score $Z$ as a gate for the selected node features, 
\begin{equation}
    \mathbf{X}_{\text{pool}}=\mathbf{X}_{idx,:} \odot \mathbf{Z}_{idx},
\end{equation}
where $X_{idx,:}$ denotes the feature matrix indexed by the selected nodes, and $\odot$ denotes the broadcasted element-wise product. With the help of the gate mechanism, the scoring function can be trained by back-propagation, to adaptively evaluate the importance of nodes according to a certain task. Several representative TopK-based pooling methods are reviewed in detail in the following. 

\smallskip
\textit{gPool}~\citep{gao2019graph}. gPool is one of the first works to select the most important node subset from the original graph to construct the coarsened graph by Top K operation. The key idea of gPool is to evaluate the importance of all nodes
by learning a projection vector $\mathbf{p}$, which projects node features to a scalar score, 
\begin{equation}
\mathbf{Z}_j=\mathbf{X}_{j,:} \mathbf{p} /\|\mathbf{p}\|,
\end{equation}
and then select nodes with K-highest scores to form the pooled graph.

\smallskip
\textit{Self-Attention Graph Pooling~(SAGPool)}~\citep{lee2019self}. Unlike gPool \citep{gao2019graph}, which only uses node features to generate projection scores, SAGPool captures both graph topology and node features to obtain self-attention scores by graph convolution. The various formulas of graph convolution can be employed to compute the self-attention score $\mathbf{Z}$, 
\begin{equation}
    \mathbf{Z}=\sigma(\operatorname{GNN}(\mathbf{X}, \mathbf{A})),
\end{equation}
where $\sigma$ denotes the activation function, and $\operatorname{GNN}$ denotes various graph convolutional layers or stacks of them, whose output dimension is one.

\smallskip
\textit{Hierarchical Graph Pooling with Structure Learning~(HGP-SL)}~\citep{zhang2019hierarchical} . HGP-SL evaluates the importance score of a node according to the information it contains given its neighbors. It supposes that a node which can be easily represented by its neighborhood carries relatively little information.
Specifically, the importance score can be defined by the Manhattan distance between the original node representation and the reconstructed one aggregated from its neighbors' representation,
\begin{equation}
    \mathbf{Z}=\left\|\left(\mathbf{I}-\mathbf{D}^{-1} \mathbf{A}\right) \mathbf{X}\right\|_1,
\end{equation}
where $\mathbf{I}$ denotes the identity matrix, $\mathbf{D}$ denotes the diagonal degree matrix of $\mathbf{A}$, and $\|\cdot\|_1$ means $\ell_1$ norm. Furthermore, to reduce the loss of topology information, HGP-SL leverages structure learning to learn a refined graph topology for the reserved nodes. Specifically, it utilizes the attention mechanism to compute the similarity of two nodes as the weight of the reconstructed edge, 
\begin{equation}
    \widetilde{\boldsymbol{A}}^{\text{pool}}_{i j}=\operatorname{sparsemax}\left(\sigma\left(\mathbf{w}\left[\boldsymbol{X}^{\text{pool}}_{i,:} \| \boldsymbol{X}^{\text{pool}}_{j,:}\right]^{\top}\right)+\lambda \cdot \boldsymbol{A}^{\text{pool}}_{i j}\right),
\end{equation}
where $\widetilde{\boldsymbol{A}}^{\text{pool}}$ denotes the refined adjacency matrix of the pooled graph, $\operatorname{sparsemax}(\cdot)$ truncates the values below a threshold to zeros, $\mathbf{w}$ denotes a learnable weight vector, and $\lambda$ is a weight parameter between the original edges and the reconstructed edges. These reconstructed edges may capture the underlying relationship between nodes disconnected due to node dropping.

\smallskip
\textit{Topology-Aware Graph Pooling~(TAPool)}~\citep{gao2021topology}. TAPool takes both the local and global significance of a node into account. On the one hand, it utilizes the average similarity between a node and its neighbors to evaluate its local importance, 
\begin{equation}
    \hat{\mathbf{R}} = \left(\mathbf{X} \mathbf{X}^T\right) \odot\left(\mathbf{D}^{-1} \mathbf{A}\right), 
    \mathbf{Z}^l = \operatorname{softmax}\left(\frac{1}{n}\hat{\mathbf{R}}\mathbf{1}_n\right),
\end{equation}
where $\hat{\mathbf{R}}$ denotes the localized similarity matrix, and $\mathbf{Z}^l$ denotes the local importance score. On the other hand, it measures the global importance of a node according to the significance of its one-hop neighborhood in the whole graph, 
\begin{equation}
    \hat{\mathbf{X}} = \mathbf{D}^{-1} \mathbf{A}\mathbf{X}, \mathbf{Z}^g = \operatorname{softmax} \left(\hat{\mathbf{X}} \mathbf{p} \right),
\end{equation}
where $\mathbf{p}$ is a learnable and globally shared projector vector, similar to the aforementioned gPool~\citep{gao2019graph}. However, $\hat{\mathbf{X}}$ here further aggregates the features from the neighborhood, which enables the global importance score $\mathbf{Z}^g$ to capture more topology information such as salient subgraphs. Moreover, TAPool encourages connectivity in the coarsened graph with the help of a degree-based connectivity term, then obtaining the final importance score $\mathbf{Z} = \mathbf{Z}^l + \mathbf{Z}^g + \lambda \mathbf{D} / |V|$, where $\lambda$ is a trade-off hyperparameter.

\subsubsection{Cluster-based Pooling} 
Pooling the graph by clustering and merging nodes is the main concept behind cluster-based pooling methods. Typically, they allocate nodes to a collection of clusters by learning a cluster assignment matrix $\mathbf{S} \in \mathbb{R}^{|V| \times K}$, where $K$ is the number of the clusters. After that, they merge the nodes within each cluster to generate a new node in the pooled graph. 
The feature (embedding) matrix of the new nodes can be obtained by aggregating the features (embeddings) of nodes within the clusters, according to the cluster assignment matrix,
\begin{equation}
\mathbf{X}^{\text{pool}}=\mathbf{S}^T \mathbf{X}.
\end{equation}

While the adjacency matrix of the pooled graph can be generated by calculating the connectivity strength between each pair of clusters, 
\begin{equation}
\mathbf{A}^{\text{pool}}=\mathbf{S}^T \mathbf{A} \mathbf{S}.
\label{eq:adj_cluster}
\end{equation}

Then, we review several representative cluster-based pooling methods in detail.

\smallskip
\textit{DiffPool}~\citep{ying2018hierarchical}. DiffPool is one of the first and classic works to hierarchically pool the graph by graph clustering. Specifically, it uses an embedding GNN to generate embeddings of nodes, and a pooling GNN to generate the cluster assignment matrix, 
\begin{equation}
    \hat{\mathbf{X}}=\operatorname{GNN}_{\text{embed}}\left(\mathbf{X}, \mathbf{A}\right), \mathbf{S}=\operatorname{softmax}\left(\operatorname{GNN}_{\text{pool}}\left(\mathbf{X}, \mathbf{A}\right)\right), \mathbf{X}^{\text{pool}}=\mathbf{S}^T \hat{\mathbf{X}}.
\end{equation}

Besides, DiffPool leverages an auxiliary link prediction objective $L_{\mathrm{LP}}=\left\|\mathbf{A}, \mathbf{S} \mathbf{S}^T\right\|_F$ to encourage the adjacent nodes to be in the same cluster and avoid fake local minima, where $\|\cdot \|_F$ is the Frobenius norm. And it utilizes an entropy regularization term $L_{\mathrm{E}}=\frac{1}{|V|} \sum_{i=1}^{|V|} H\left(\mathbf{S}_i\right)$ to impel clear cluster assignments, where $H(\cdot)$ represents the entropy.

\smallskip
\textit{Graph Pooling with Spectral Clustering~(MinCutPool)}~\citep{bianchi2020spectral}. MinCutPool takes advantage of the properties of spectral clustering (SC) to provide a better inductive bias and avoid degenerate cluster assignments. It learns to cluster like SC by optimizing the MinCut loss, 
\begin{equation}
    L_c = -\frac{\operatorname{Tr}\left(\mathbf{S}^T {\mathbf{A}} \mathbf{S}\right)}{\operatorname{Tr}\left(\mathbf{S}^T {\mathbf{D}} \mathbf{S}\right)}.
\end{equation}

In addition, it utilizes an orthogonality loss $L_o = \left\|\frac{\mathbf{S}^T \mathbf{S}}{\left\|\mathbf{S}^T \mathbf{S}\right\|_F}-\frac{\mathbf{I}_K}{\sqrt{K}}\right\|_F$ to encourage orthogonal and uniform cluster assignments, and prevent the bad minima of $L_c$, where $K$ is the number of the clusters. When performing a specific task, it can optimize the weighted sum of the unsupervised loss $L_u = L_c + L_o$ and a task-specific loss to find the optimal balance between the theoretical prior and the task objective.

\smallskip
\textit{Structural Entropy Guided Graph Pooling~(SEP)}~\citep{wu2022structural}. In order to lessen the local structural harm and suboptimal performance caused by separate pooling layers and predesigned pooling ratios, SEP leverages the concept of structural entropy to generate the global and hierarchical cluster assignments at once. Specifically, SEP treats the nodes of a given graph as the leaf nodes of a coding tree and exploits the hierarchical layers of the coding tree to capture the hierarchical structure of the graph. The optimal code tree $T$ can be obtained by minimizing the structural entropy \citep{li2016structural},
\begin{equation}
    \mathcal{H}^T(G)=-\sum_{v_i \in T} \frac{g(P_{v_i})}{\operatorname{vol}(V)} \log \frac{\operatorname{vol}\left(P_{v_i}\right)}{\operatorname{vol}\left(P_{v_i^{+}}\right)},
\end{equation}
where 
$v_i^{+}$ represents the father node of node $v_i$, $P_{v_i}$ denotes the partition of leaf nodes which are descendants of $v_i$ in the coding tree $T$, 
$g(P_{v_i})$ denotes the number of edges that have a terminal in the $P_{v_i}$, and $\operatorname{vol}(\cdot)$ denotes the total degrees of leaf nodes in the given partition. Then, the cluster assignment matrix for each pooling layer can be derived from the edges of each layer in the coding tree. With the help of the one-step joint assignments generation based on structural entropy, it can not only make the best use of the hierarchical relationships of pooling layers, but also reduce the structural noise in the original graph.

\subsubsection{Hybrid pooling.} 

Hybrid pooling methods combine TopK-based pooling methods and cluster-based pooling methods, to exert the advantages of the two methods and overcome their respective limitations. Here, we review two representative hybrid pooling methods, Adaptive Structure Aware Pooling ~\citep{ranjan2020asap} and Multi-channel Pooling ~\citep{ijcai2021p199}.

\smallskip
\textit{Adaptive Structure Aware Pooling~(ASAP)}~\citep{ranjan2020asap}. Considering that TopK-based pooling methods are not good at capturing the connectivity of the coarsened graph, while cluster-based pooling methods fail to be employed for large graphs because of the dense assignment matrix, ASAP organically combines the two types of pooling methods to overcome the above limitations. Specifically, it regards the $h$-hop ego-network $c_h(v_i)$ of each node $v_i$ as a cluster. Such local clustering enables the cluster assignment matrix to be sparse. Then, a new self-attention mechanism Master2Token is used to learn the cluster assignment matrix $\mathbf{S}$ and the cluster representations,
\begin{equation}
    \mathbf{m}_i=\max _{v_j \in c_h\left(v_i\right)}\left(\mathbf{X}_j^{\prime}\right),
    \mathbf{S}_{j, i}=\operatorname{softmax}\left(\mathbf{w}^T \sigma\left(\mathbf{W} \mathbf{m}_i \| \mathbf{X}_j^{\prime}\right)\right),
    \mathbf{X}_i^c=\sum_{j=1}^{\left|c_h\left(v_i\right)\right|} \mathbf{S}_{j, i} \mathbf{X}_j,
\end{equation}
where $\mathbf{X}^{\prime}$ is the node embedding matrix after passing GCN, $\mathbf{w}$ and $\mathbf{W}$ denote the trainable vector and matrix respectively, and $\mathbf{X}_i^c$ denotes the representation of the cluster $c_h(v_i)$. Next, it utilizes the graph convolution and TopK selection to choose the top $K$ clusters, whose centers are treated as the nodes of the pooled graph. The adjacency matrix of the pooled graph can be calculated like common cluster-based pooling methods (\ref{eq:adj_cluster}),
preserving the connectivity of the original graph well.

\smallskip
\textit{Multi-channel Pooling~(MuchPool)}~\citep{ijcai2021p199}. The key idea of MuchPool is to capture both the local and global structure of a given graph by combining the TopK-based pooling methods and the cluster-based pooling methods. MuchPool has two pooling channels based on TopK selection to yield two fine-grained pooled graphs, whose selection criteria are node degrees and projected scores of node features respectively, so that both the local topology and the node features are considered. Besides, it leverages a channel based on graph clustering to obtain a coarse-grained pooled graph, which captures the global and hierarchical structure of the input graph. To better integrate the information of different channels, a cross-channel convolution is proposed, which fuses the node embeddings of the fine-grained pooled graph $\mathbf{X}^{\text{fine}}$ and the coarse-grained pooled graph $\mathbf{X}^{\text{coarse}}$ with the help of the cluster assignments $\mathbf{S}$ of the cluster-based pooling channel,
\begin{equation}
    \widetilde{\mathbf{X}}^{\text{fine}}=\sigma\left(\left[\mathbf{X}^{\text{fine}}+ \mathbf{S} \mathbf{X}^{\text{coarse}}\right] \cdot \mathbf{W} \right),
\end{equation}
where $\mathbf{W}$ denotes the learnable weights. Finally, it merges the node embeddings and the adjacency matrices of the two fine-grained pooled graphs to obtain the eventually pooled graph.

\subsection{Summary}

This section introduces graph pooling. We provide the summary as follows:

\begin{itemize}
    \item \textbf{Techniques.} 
    Graph pooling methods play a vital role in generating an entire graph representation by aggregating node embeddings. There are mainly two categories of graph pooling methods: global pooling methods and hierarchical pooling methods. While global pooling methods directly aggregate node embeddings in one step, hierarchical pooling methods gradually coarsen a graph to capture hierarchical structure characteristics of the graph based on TopK selection, clustering methods, or hybrid methods.
    \item \textbf{Challenges and Limitations.} 
    Despite the great success of graph pooling methods for learning the whole graph representation, there remain several challenges and limitations unsolved:
    1) For hierarchical pooling, most cluster-based methods involve the dense assignment matrix, which limits their application to large graphs, while TopK-based methods are not good at capturing structure information of the graph and may lose information due to node dropping. 2) Most graph pooling methods are designed for simple attributed graphs, while pooling algorithms tailored to other types of graphs, like dynamic graphs and heterogeneous graphs, are largely under-explored.
    \item \textbf{Future Works.} In the future, we expect that more hybrid or other pooling methods can be studied to capture the graph structure information sufficiently as well as be efficient for large graphs. In realistic scenarios, there are various types of graphs involving dynamic, heterogeneous, or spatial-temporal information. It is promising to design graph pooling methods specifically for these graphs, which can be beneficial to more real-world applications, such as traffic analysis and recommendation systems. 
\end{itemize}

\section{Graph Transformer} 
Though GNNs based on the message-passing paradigm have achieved impressive performance on multiple well-known tasks \citep{gilmer2017neural, xu2018powerful, wang2019planit, li2020deepergcn}, they still face some intrinsic problems due to the iterative neighbor-aggregation operation. 
Many previous works have demonstrated the two major defects of message-passing GNNs, which are known as the over-smoothing and long-distance modeling problems.
And there are lots of explanatory works trying to mine insights from these two issues. 
The over-smoothing problem can be explained in terms of various GNNs focusing only on low-frequency information \citep{bo2021beyond}, mixing information between different kinds of nodes destroying model performance \citep{chen2020measuring}, GCN is equivalent to Laplacian smoothing \citep{li2018deeper}, isotropic aggregation among neighbors leading to the same influence distribution as random walk \citep{xu2018representation}, etc.  
The inability to model long-distance dependencies of GNNs is partially due to the over-smoothing problem, because in the context of conventional neighbor-aggregation GNNs, node information can be passed over long distances only through multiple GNN layers.
Recently, \citet{alon2020bottleneck} found that this problem may also be caused by over-squashing, which means the exponential growth of computation paths with increasing distance.
Though the two basic performance bottlenecks can be tackled with elaborate message passing and aggregation strategies, the representational power of GNNs is inherently bounded by the Weisfeiler-Lehman isomorphism hierarchy \citep{morris2019weisfeiler}. Worse still, most GNNs \citep{kipf2016semi, velivckovic2017graph, gilmer2017neural} are bounded by the simplest first-order Weisfeiler-Lehman
test (1-WL). Some efforts have been dedicated to break this limitation, such as hypergraph-based \citep{feng2019hypergraph, huang2021unignn}, path-based \citep{cai2020graph, ying2021transformers}, and k-WL-based \citep{balcilar2021breaking, morris2019weisfeiler} approaches.

Among many attempts to solve these fundamental problems, an essential one is the adaptation of Transformer \citep{vaswani2017attention} for graph representation learning. 
Transformers, both the vanilla version and several variants, have been adopted with impressive results in various deep learning fields including natural language processing~\citep{vaswani2017attention, devlin2018bert}, computer vision \citep{carion2020end, zhu2020deformable}, etc. Recently, Transformer also shows powerful graph modeling abilities in many researches \citep{dwivedi2020generalization, kreuzer2021rethinking, ying2021transformers, wu2021representing, chen2022structure, ma2023rethinking, zhao2023more}. Extensive empirical results show that some chronic shortcomings in conventional GNNs can be easily overcome in Transformer-based approaches. This section gives an overview of the current progress on this kind of method. 

\subsection{Transformer}

 Transformer~\citep{vaswani2017attention} was first applied to model machine translation, but two of the key mechanisms adopted in this work, attention operation and positional encoding, are highly compatible with the graph modeling problem. 

To be specific, we denote the input of attention layer in Transformer as $\mathbf{X} = [ \mathbf{x}_0, \mathbf{x}_1, \ldots, \mathbf{x}_{n-1} ]$, $\mathbf{x}_i \in \mathbb{R}^{d}$, where $n$ is the length of input sequence and $d$ is the dimension of each input embedding $\mathbf{x}_i$. 
Then the core operation of calculating new embedding $\hat{\mathbf{x}}_i$ for each $\mathbf{x}_i$ in attention layer can be streamlined as:
\begin{equation}
    \label{eq:Transformer_attention}
    \begin{gathered}
        \mathrm{s}^h(\mathbf{x}_i, \mathbf{x}_j) = \text{NORM}_j(\mathop{\|}_{\mathbf{x}_k \in \mathbf{X}}\mathcal{Q}^h(\mathbf{x}_i)^{\mathrm{T}}\mathcal{K}^h(\mathbf{x}_k)),\\
        \mathbf{x}_i^h\ = \mathop{\sum}_{\mathbf{x}_j \in \mathbf{X}}{\mathrm{s}^h(\mathbf{x}_i,\mathbf{x}_j) \mathcal{V}^h(\mathbf{x}_j}),\\
        \hat{\mathbf{x}}_i = \text{MERGE} (\mathbf{x}_i^1, \mathbf{x}_i^2, \ldots, \mathbf{x}_i^H), 
    \end{gathered}
\end{equation}where $h \in \{0, 1,\ldots,H-1\}$ represents the attention head number. $\mathcal{Q}^h$, $\mathcal{K}^h$ and $\mathcal{V}^h$ are projection functions mapping a vector to the query space, key space and value space respectively. $\mathrm{s}^h(\mathbf{x}_i, \mathbf{x}_j)$ is score function measuring the similarity between $\mathbf{x}_i$ and $\mathbf{x}_j$. 
NORM is the normalization operation ensuring $\sum_{\mathbf{x}_j \in \mathbf{X} } \mathrm{s}^h(\mathbf{x}_i, \mathbf{x}_j) \equiv 1$ to propel the stability of the output generated by a stack of attention layers, it is usually performed as scaled softmax: $\text{NORM}(\cdot) = \text{SoftMax}(\cdot/\sqrt{d})$. 
And MERGE function is designed to combine the information extracted from multiple attention heads. 
Here, we omit further implementation details that do not affect our understanding of attention operation.


The attention process cannot encode the position information of each $\mathbf{x}_i$, which is essential in machine translation problems. So positional encoding is introduced to remedy this deficiency, and it's calculated as:
\begin{equation}
    \label{eq:Transformer_encoding}
    \begin{gathered}
    \mathbf{X}^{pos}_{i,2j} = \sin (i/10000^{2j/d}), \ 
    \mathbf{X}^{pos}_{i,2j+1} = \cos (i/10000^{2j/d}),
    \end{gathered}
\end{equation}where $i$ is the position and $j$ is the dimension. The positional encoding is added to the input before it is fed to the Transformer.

\begin{table}[t]
\renewcommand\arraystretch{0.935}
\caption{Summary of graph transformer methods.}
\label{table_graph_transformer}
\centering
\resizebox{\textwidth}{!}{
\begin{tabular}{c|cc|ccc}
\toprule
  \multirow{2}*{Method} & \multicolumn{2}{c|}{Technique} & \multicolumn{3}{c}{Capacity}\\
    & Attention Modification & Encoding Enhancement & Heterogeneous & Long Distance & > 1-WL\\
\midrule
GGT \citep{dwivedi2020generalization} & \checkmark & \checkmark & & structure only & \checkmark\\

GTSA \citep{kreuzer2021rethinking} & \checkmark & \checkmark & & \checkmark & \checkmark\\

HGT \citep{hu2020heterogeneous} & \checkmark &   & \checkmark &  & \\

G2SHGT \citep{yao2020heterogeneous} & \checkmark & & \checkmark & \checkmark & \\

HINormer \citep{mao2023hinormer}  & \checkmark & & \checkmark & \checkmark & \\

GRUGT \citep{cai2020graph} & \checkmark &  & & \checkmark & \checkmark\\

GRIT \citep{ma2023graph} & \checkmark &  & & \checkmark & \checkmark\\

Graphormer-GD \citep{zhang2023rethinking} & \checkmark &  & & \checkmark & \checkmark\\

Graphormer \citep{ying2021transformers} & \checkmark & \checkmark & & \checkmark & \checkmark\\

GSGT \citep{hussain2021edge} &  & \checkmark & & \checkmark & \checkmark\\

TMDG \citep{geisler2023transformers} &  & \checkmark & & \checkmark & \checkmark\\

Graph-BERT \citep{zhang2020graph}  & & \checkmark & & \checkmark & \checkmark\\

LRGT \citep{wu2021representing} &  & \checkmark & & \checkmark & \\

SAT \citep{chen2022structure} &  & \checkmark & & \checkmark & \checkmark\\
\bottomrule

\end{tabular}
}
\end{table}

\subsection{Overview}

From the simplified process shown in Eq. \ref{eq:Transformer_attention}, we can see that the core of the attention operation is to accomplish information transfer based on the similarity between the source and the target to be updated. It's quite similar to the message-passing process on a fully-connected graph. However, the direct application of this architecture to arbitrary graphs does not make use of structural information, so it may lead to poor performance when graph topology is important. On the other hand, the definition of positional encoding in graphs is not a trivial problem because the order or coordinates of graph nodes are underdefined.

According to these two challenges, Transformer-based methods for graph representation learning can be classified into two major categories, one considering graph structure during the attention process, and the other encoding the topological information of the graph into initial node features. We name the first one as \textit{Attention Modification} and the second one as \textit{Encoding Enhancement}. A summarization is provided in Table \ref{table_graph_transformer}. In the following discussion, if both methods are used in one paper, we will list them in different subsections, and we will ignore the multi-head trick in attention operation.

\subsection{Attention Modification}

This group of works attempts to modify the full attention operation to capture structure information. The most prevalent approach is changing the score function, which is denoted as $\mathrm{s}(\cdot, \cdot)$ in Eq. \ref{eq:Transformer_attention}. 
GGT~\citep{dwivedi2020generalization} constrains each node feature can only attend to neighbors and enables the model to represent edge feature information by rewrite $\mathrm{s}(\cdot, \cdot)$ as:
\begin{equation}
    \begin{gathered}
        \tilde{\mathrm{s}}_1(\mathbf{x}_i, \mathbf{x}_j) = \left\{
        \begin{aligned}
            &(\mathbf{W}^Q\mathbf{x}_i)^{\mathrm{T}}(\mathbf{W}^K\mathbf{x}_j \odot \mathbf{W}^E\mathbf{e}_{ji}),& \left<j,i\right> \in E \\
            &-\infty,& \text{otherwise}
        \end{aligned}
        \right., \\
        \mathrm{s}_1(\mathbf{x}_i, \mathbf{x}_j) = \text{SoftMax}_j( \mathop{\|}_{\mathbf{x}_k \in \mathbf{X}} \tilde{\mathrm{s}}_1(\mathbf{x}_i, \mathbf{x}_k)),
    \end{gathered}
\end{equation}where $\odot$ is Hadamard product and $\mathbf{W}^{Q,K,E}$ represents trainable parameter matrix. 
This approach is not efficient yet to model long-distance dependencies since only 1st-neighbors are considered. Though it adopts Laplacian eigenvectors to gather global information (see Section \ref{sec:Transformer_encoding}), but only long-distance structure information is remedied while the node and edge features are not. 
GTSA~\citep{kreuzer2021rethinking} improves this approach by combining the original graph and the full graph. Specifically, it extends $\mathrm{s}_1(\cdot, \cdot)$ to:
\begin{equation}
    \begin{gathered}
        \tilde{\mathrm{s}}_2(\mathbf{x}_i, \mathbf{x}_j) = \left\{
        \begin{aligned}
            &(\mathbf{W}^Q_1\mathbf{x}_i)^{\mathrm{T}}(\mathbf{W}^K_1\mathbf{x}_j \odot \mathbf{W}^E_1\mathbf{e}_{ji}),& \left<j,i\right> \in E \\
            &(\mathbf{W}^Q_0\mathbf{x}_i)^{\mathrm{T}}(\mathbf{W}^K_0\mathbf{x}_j \odot \mathbf{W}^E_0\mathbf{e}_{ji}),& \text{otherwise}
        \end{aligned}
        \right.,\\
        \mathrm{s}_2(\mathbf{x}_i, \mathbf{x}_j) = \left\{ 
        \begin{aligned}
            &\frac{1}{1+\lambda}\text{SoftMax}_j( \mathop{\|}_{\left< k, i \right> \in E} \tilde{\mathrm{s}}_2(\mathbf{x}_i, \mathbf{x}_k)), & \left<j,i\right> \in E \\
            &\frac{\lambda}{1+\lambda}\text{SoftMax}_j( \mathop{\|}_{\left< k, i \right> \not\in E} \tilde{\mathrm{s}}_2(\mathbf{x}_i, \mathbf{x}_k)), &\text{otherwise}
        \end{aligned}
        \right.,
    \end{gathered}
\end{equation}where $\lambda$ is a hyperparameter representing the strength of the full connection. 

Some works try to reduce information-mixing problems \citep{chen2020measuring} in heterogeneous graphs.
HGT~\citep{hu2020heterogeneous} disentangles the attention of different node types and edge types by adopting additional attention heads. It defines $\mathbf{W}_{Q,K,V}^{\tau(v)}$ for each node type $\tau(v)$ and $\mathbf{W}_E^{\phi(e)}$ for each edge type $\phi(e)$, $\tau(\cdot)$ and $\phi(\cdot)$ are type indicating function. 
G2SHGT~\citep{yao2020heterogeneous} defines four types of subgraphs, \textit{fully-connected}, \textit{connected}, \textit{default} and \textit{reverse}, to capture global, undirected, forward and backward information respectively. Each subgraph is homogeneous, so it can reduce interactions between different classes.

Path features between nodes are always treated as inductive bias added to the original score function. Let $\text{SP}_{ij} = (e_1,e_2,\ldots,e_N)$ denote the shortest path between node pair $(v_i, v_j)$. 
GRUGT~\citep{cai2020graph} uses  GRU \citep{chung2014empirical} to encode forward and backward features as: $\mathbf{r}_{ij}=\text{GRU}(\text{SP}_{ij})$, $\mathbf{r}_{ji} = \text{GRU}(\text{SP}_{ji})$. Then, the final attention score is calculated by adding up four components:
\begin{equation}
    \begin{gathered}
        \tilde{\mathrm{s}}_3(\mathbf{x}_i, \mathbf{x}_j) = (\mathbf{W}^Q\mathbf{x}_i)^{\mathrm{T}} \mathbf{W}^K\mathbf{x}_j + (\mathbf{W}^Q\mathbf{x}_i)^{\mathrm{T}} \mathbf{W}^K\mathbf{r}_{ji}+(\mathbf{W}^Q\mathbf{r}_{ij}) ^{\mathrm{T}} \mathbf{W}^K\mathbf{x}_j + (\mathbf{W}^Q\mathbf{r}_{ij})^{\mathrm{T}} \mathbf{W}^K\mathbf{r}_{ji},
    \end{gathered}
\end{equation} from front to back, which represent content-based score, source-dependent bias, target-dependent bias and universal bias respectively. Graphormer~\citep{ying2021transformers} uses both path length and path embedding to introduce structural bias as:
\begin{equation}
\begin{gathered}
    \tilde{\mathrm{s}}_4(\mathbf{x}_i, \mathbf{x}_j) = (\mathbf{W}^Q\mathbf{x}_i)^{\mathrm{T}} \mathbf{W}^K\mathbf{x}_j / \sqrt{d} + b_N + c_{ij}, \\
    c_{ij} = \frac{1}{N}\sum_{k=1}^{N} (\mathbf{e}_k)^{\mathrm{T}} \mathbf{w}^E_{k},\\
    \mathrm{s}_4(\mathbf{x}_i, \mathbf{x}_j) = \text{SoftMax}_j( \mathop{\|}_{\mathbf{x}_k \in \mathbf{X}} \tilde{\mathrm{s}}_4(\mathbf{x}_i, \mathbf{x}_k)),
\end{gathered}
\end{equation}where $b_N$ is a trainable scalar indexed by $N$, the length of $\text{SP}_{ij}$ . $\mathbf{e}_k$ is the embedding of the the edge $e_k$, and $\mathbf{w}^E_{k} \in \mathbb{R}^{d}$ is the $k$-th edge parameter. If $\text{SP}_{ij}$ does not exist, then $b_N$ and $c_{ij}$ are set to be special values. 
GRIT~\citep{ma2023graph} utilizes relative random walk probabilities as an inductive bias to encode relative path information.
Graphormer-GD~\citep{zhang2023rethinking} also incorporates relative distance as bias, and rigorously proves that this bias is crucial for determining the biconnectivity of a graph.



\subsection{Encoding Enhancement}

\label{sec:Transformer_encoding}
This kind of method intends to enhance initial node representations to enable the Transformer to encode structure information. They can be further divided into two categories, position-analogy methods and structure-aware methods.


\subsubsection{Position-analogy methods}

In Euclidean space, the Laplacian operator corresponds
to the divergence of the gradient, whose eigenfunctions are sine/cosine functions. For the graph, the Laplacian operator is the Laplacian matrix, whose eigenvectors can be considered as eigenfunctions. Hence, inspired by Eq. \ref{eq:Transformer_encoding}, position-analogy methods utilize Laplacian eigenvectors to simulate positional encoding $\mathbf{X}^{pos}$ as they are the equivalents of sine/cosine functions.

Laplacian eigenvectors can be calculated via the eigendecomposition of normalized graph Laplacian matrix $\tilde{\mathbf{L}}$:
\begin{equation}
    \tilde{\mathbf{L}} \triangleq \mathbf{I} - \mathbf{D}^{-1/2} \mathbf{A} \mathbf{D}^{-1/2} = \mathbf{U} \mathbf{\Lambda} \mathbf{U}^{\text{T}},
\end{equation} where $\mathbf{A}$ is the adjacency matrix, $\mathbf{D}$ is the degree matrix, $\mathbf{U} = [\mathbf{u}_1, \mathbf{u}_2, \ldots, \mathbf{u}_{n-1}]$ are eigenvectors and $\mathbf{\Lambda} = diag(\lambda_0, \lambda_1, \ldots, \lambda_{n-1})$ are eigenvalues.
With $\mathbf{U}$ and $\mathbf{\Lambda}$, GGT~\citep{dwivedi2020generalization} uses eigenvectors of the k smallest non-trivial eigenvalues to denote the intermediate embedding $\mathbf{X}^{mid} \in \mathbb{R}^{n \times k}$, and maps it to d-dimensional space and gets the position encoding $\mathbf{X}^{pos} \in \mathbb{R}^{n \times d}$. This process can be formalized as:
\begin{equation}
    \label{eq:Transformer_spatialE}
    \begin{gathered}
    index = \text{argmin}_{k} (\{\lambda_i| 0\le i < n \wedge \lambda_i > 0\}), \\
    \mathbf{X}^{mid} = [\mathbf{u}_{index_0}, \mathbf{u}_{index_1}, \ldots, \mathbf{u}_{index_{k-1}}]^{\text{T}}, \\
    \mathbf{X}^{pos} = \mathbf{X}^{mid} \mathbf{W}^{k \times d},
    \end{gathered}
\end{equation}where $index$ is the subscript of the selected eigenvectors. 
GTSA~\citep{kreuzer2021rethinking} puts eigenvector $\mathbf{u}_i$ on the frequency axis at $\lambda_i$ and uses sequence modeling methods to generate positional encoding. Specifically, it extends $\mathbf{X}^{mid}$ in Eq. \ref{eq:Transformer_spatialE} to $\tilde{\mathbf{X}}^{mid} \in \mathbb{R}^{n \times k \times 2 }$ by concatenating each value in eigenvectors with corresponding eigenvalue, and then positional encoding $\mathbf{X}^{pos} \in \mathbb{R}^{n \times d}$ are generated as:
\begin{equation}
    \begin{gathered}
        \mathbf{X}^{input} = \tilde{\mathbf{X}}^{mid} \mathbf{W}^{2 \times d}, \\
        \mathbf{X}^{pos} = \text{SumPooling} (\text{Transformer}(\mathbf{X}^{input}), \text{dim} = 1).
    \end{gathered}
\end{equation}

Here, $\mathbf{X}_{input} \in \mathbb{R}^{n \times k \times d}$ is equivalent to the input matrix in sequence modeling with shape $(batch\_size, length, dim)$, and can be naturally processed by Transformer. 
Since the Laplacian eigenvectors can be complex-valued for directed graph, GSGT~\citep{hussain2021edge} proposes to utilize SVD of adjacency matrix $\mathbf{A}$, which is denoted as $\mathbf{A} = \mathbf{U} \mathbf{\Sigma} \mathbf{V}^{\text{T}}$, and uses the largest $k$ singular values $\mathbf{\Sigma}_k$ and associated left and right singular vectors $\mathbf{U}_k$ and $\mathbf{V}_k^{\text{T}}$ to output $\mathbf{X}^{pos}$ as $\mathbf{X}^{pos} = [\mathbf{U}_k \mathbf{\Sigma}_k^{1/2} \| \mathbf{V}_k \mathbf{\Sigma}_k^{1/2}]$, where $\|$ is the concatenation operation. In addition to SVD, TMDG~\citep{geisler2023transformers} processes directed graphs by utilizing the Magnetic Laplacian.
All these methods above randomly flip the signs of eigenvectors or singular vectors during the training phase to promote the invariance of the models to the sign ambiguity.

\subsubsection{Structure-aware methods}

In contrast to position-analogy methods, structure-aware methods do not attempt to mathematically rigorously simulate sequence positional encoding. They use some additional mechanisms to directly calculate structure-related encoding.

Some approaches compute extra encoding $\mathbf{X}^{add}$ and add it to the initial node representation. Graphormer~\citep{ying2021transformers} proposes to leverage node centrality as an additional signal to address the importance of each node. Concretely, $\mathbf{x}_{i}^{add}$ is determined by the in-degree $\text{deg}_i^{-}$ and outdegree $\text{deg}_i^{+}$:
\begin{equation}
    \mathbf{x}_{i}^{add} = \mathcal{P}^{-}(\text{deg}_i^{-}) + \mathcal{P}^{+}(\text{deg}_i^{+}),
\end{equation}where $\mathcal{P}^{-}$ and $\mathcal{P}^{+}$ are learnable embedding function. 
Graph-BERT~\citep{zhang2020graph} employs Weisfeiler-Lehman algorithm to label node $v_i$ to a number $\text{WL}(v_i) \in \mathbb{N}$ and defines $\mathbf{x}_{i}^{add}$ as:
\begin{equation}
    \begin{gathered}
    \mathbf{x}_{i, 2j}^{add} = \sin (\text{WL}(v_i)/10000^{2j/d}), \ 
    \mathbf{x}_{i, 2j+1}^{add} = \cos (\text{WL}(v_i)/10000^{2j/d}).
    \end{gathered}
\end{equation}

The other approaches try to leverage GNNs to initialize inputs to the Transformer. LRGT~\citep{wu2021representing} applies GNN to get intermediate vectors as $\mathbf{X}'=\text{GNN}(\mathbf{X})$, and passes the concatenation of $\mathbf{X}'$ and a special vector $\mathbf{x}_{\text{CLS}}$ to Transformer layer as: $\hat{\mathbf{X}} = \text{Transformer}([\mathbf{X}'\|\mathbf{x}_{\text{CLS}}])$. Then $\hat{\mathbf{x}}_{\text{CLS}}$ can be used as the representation of the entire graph for downstream tasks. This method cannot break the 1-WL bottleneck because it uses GCN \citep{kipf2016semi} and GIN \citep{xu2018powerful} as graph encoders in the first step, which are intrinsically limited by 1-WL test. SAT~\citep{chen2022structure} improves this deficiency by using subgraph-GNN NGNN \citep{zhang2021nested} for initialization, and achieves outstanding performance. 

\subsection{Summary}

This section introduces Transformer-based approaches for graph representation learning and we provide the summary as follows:

\begin{itemize}
    \item \textbf{Techniques.} Graph Transformer methods modify two fundamental techniques in Transformer, attention operation and positional encoding, to enhance its ability to encode graph data. Typically, they introduce fully connected attention to model long-distance relationships, utilize shortest path and Laplacian eigenvectors to break 1-WL bottleneck, and separate points and edges belonging to different classes to avoid over-mixing problems.
    
    \item \textbf{Challenges and Limitations.} Though Graph Transformers achieve encouraging performance, they still face two major challenges. 
    The first challenge is the computational cost of the quadratic attention mechanism and shortest path calculation. These operations require significant computing resources and can be a bottleneck, particularly for large graphs.
    The second is the reliance of Transformer-based models on large amounts of data for stable performance. It poses a challenge when dealing with problems that lack sufficient data, especially for few-shot and zero-shot settings.
    
    \item \textbf{Future Works.} We expect efficiency improvement for Graph Transformer should be further explored. Additionally, there are some works using pre-training and fine-tuning frameworks to balance performance and complexity in downstream tasks \citep{ying2021transformers}, this may be a promising solution to address the aforementioned two challenges.
\end{itemize}

\section{Semi-supervised Learning on Graphs}

We have investigated various architectures of graph neural networks in which the parameters should be tuned by a learning objective. The most prevalent optimization approach is supervised learning on graph data. Due to the label deficiency, semi-supervised learning has attracted increasing attention in the data mining community. In detail, these methods attempt to combine graph representation learning with current semi-supervised techniques including pseudo-labeling, consistency learning, knowledge distillation and active learning.
These works can be further subdivided into node-level representation learning and graph-level representation learning. We would introduce both parts in detail as in Sec. \ref{sec:nrl} and Sec. \ref{sec:grl}, respectively. A summarization is provided in Table \ref{table_semi_supervised}.

\subsection{Node Representation Learning}\label{sec:nrl}

Typically, node representation learning follows the concept of transductive learning, which has access to test unlabeled data. We first review the simplest loss objective, i.e., node-level supervised loss. This loss exploits the ground truth of labeled nodes on graphs. The standard cross-entropy is usually adopted for optimization. In formulation,
\begin{equation}
    \mathcal{L}_{NSL} = -\frac{1}{|\mathcal{Y}^{L}|} \sum_{i\in \mathcal{Y}^{L}}\mathbf{y}_{i}^{T}\log\mathbf{p}_{i},
    \label{equ:loss_sup}
\end{equation}
where $\mathcal{Y}^{L}$ denotes the set of labeled nodes. Additionally, there are a variety of unlabeled nodes that can be used to offer semantic information. To fully utilize these nodes, a range of methods attempt to combine semi-supervised approaches with graph neural networks. Pseudo-labeling~\citep{lee2013pseudo} is a fundamental semi-supervised technique that uses the classifier to produce the label distribution of unlabeled examples and then adds appropriately labeled examples to the training set~\citep{li2022cognet,zhou2019dynamic}. Another line of semi-supervised learning is consistency regularization~\citep{laine2016temporal} that requires two examples to have identical predictions under perturbation. This regularization is based on the assumption that each instance has a distinct label that is resistant to random perturbations~\citep{feng2020graph,park2021learning}. Then, we show several representative works in detail.

\begin{table}[t]
\renewcommand\arraystretch{0.935}
\caption{Summary of methods for semi-supervised Learning on Graphs. Contrastive learning can be considered as a specific kind of consistency learning. }
\label{table_semi_supervised}
\centering
\resizebox{\textwidth}{!}{
\begin{tabular}{c|c|c|c|c|c}
\toprule
  & Approach  & Pseudo-labeling & Consistency Learning  & Knowledge Distillation & Active Learning \\
\midrule
  \multirow{5}*{Node-level} &        CoGNet~\citep{li2022cognet}  & \checkmark &  &   &  \\
           & DSGCN~\citep{zhou2019dynamic} & \checkmark &  &  & \\
           & GRAND~\citep{feng2020graph} & & \checkmark &  & \\
           & AugGCR~\citep{park2021learning} & & \checkmark  & & \\
           & HCPL~\citep{luo2023towards_node}  & \checkmark &  &   &  \\
        
\midrule
\multirow{11}*{Graph-level} & SEAL~\citep{li2019semi} & \checkmark &  &  & \checkmark  \\
& InfoGraph~\citep{sun2019infograph} &  & \checkmark  & \checkmark  & \\
& DSGC~\citep{yang2022dual} & & \checkmark  &  & \\
& ASGN~\citep{hao2020asgn} & &  & \checkmark  & \checkmark \\
& TGNN~\citep{ju2022tgnn} & & \checkmark  &  & \\
& KGNN~\citep{ju2022kgnn} & \checkmark &  &  & \\
& HGMI~\citep{li2022semi} & \checkmark & \checkmark &  & \\
& ASGNN~\citep{xie2022active} & \checkmark  &  &  & \checkmark \\
& DualGraph~\citep{luo2022dualgraph} & \checkmark & \checkmark  &  & \\
 & GLA~\citep{yue2022label} &  & \checkmark  &  & \\           
 & SS~\citep{xie2022semisupervised} & \checkmark  &  &  & \\            
\bottomrule

\end{tabular}
}
\end{table}

\textit{Cooperative Graph Neural Networks~(CoGNet)}~\citep{li2022cognet}. CoGNet is a representative pseudo-label-based GNN approach for semi-supervised node classification. It employs two GNN classifiers to jointly annotate unlabeled nodes. In particular, it calculates the confidence of each node as follows:
\begin{equation}
	CV(\mathbf{p}_i)=\mathbf{p}_i^T \log\mathbf{p}_i,
\end{equation}
where $\mathbf{p}_i$ denotes the output label distribution. Then it selects the pseudo-labels with high confidence generated from one model to supervise the optimization of the other model. In particular, the objective for unlabeled nodes is written as follows:
\begin{equation}
	\mathcal{L}_{CoGNet}= \sum_{i\in \mathcal{V}^U} \mathbf{1}_{CV(\mathbf{p}_i)>\tau } \hat{\mathbf{y}}^T_i log\mathbf{q}_i,
\end{equation}
where $\hat{\mathbf{y}}_i$ denotes the one-hot formulation of the pseudo-label $\hat{y}_i=arg max \mathbf{p}_i$ and $\mathbf{q}_i$ denotes the label distribution predicted by the other classifier. $\tau$ is a pre-defined temperature coefficient. 
This cross supervision has been demonstrated effective in \citep{chen2021semi,luo2021cimon} to prevent the provision of biased pseudo-labels. 
Moreover, it employs GNNExplainer~\citep{ying2019gnnexplainer} to provide additional information from a dual perspective. Here it measures the minimal subgraphs where GNN classifiers can still generate the same prediction. In this way, CoGNet can illustrate the entire optimization process to enhance our understanding. HCPL~\citep{luo2023towards_node} incorporates curriculum learning into pseudo-labeling in semi-supervised node classification, which can generate dynamics thresholds for reliable nodes.

\smallskip
\textit{Dynamic Self-training Graph Neural Network~(DSGCN)}~\citep{zhou2019dynamic}. DSGCN develops an adaptive manner to utilize reliable pseudo-labels for unlabeled nodes. In particular, it allocates smaller weights to samples with lower confidence with the additional consideration of class balance. The weight is formulated as:
\begin{equation}
\omega_i =\frac{1}{n_{c^{i}}} \max \left(\operatorname{RELU}\left(\mathbf{p}_{i}-\beta \cdot \mathbf{1}\right)\right),
\end{equation}
where $n_{c^{i}}$ denotes the number of unlabeled samples assigned to the class $c^i$. This technique will decrease the impact of wrong pseudo-labels during iterative training.

\smallskip
\textit{Graph Random Neural Networks~(GRAND)}~\citep{feng2020graph}. GRAND is a representative consistency learning-based method. It first adds a variety of perturbations to the input graph to generate a list of graph views. Each graph view $G^r$ is sent to a GNN classifier to produce a prediction matrix $\mathbf{P}^r=[\mathbf{p}_1^r, \cdots, \mathbf{p}_N^r]$. Then it summarizes these matrices as:
\begin{equation}
	\mathbf{P} = \frac{1}{R} \mathbf{P}^r.
\end{equation} 

To provide more discriminative information and ensure that the matrix is row-normalized, GRAND sharpens the summarized label matrix into $\mathbf{P}^{SA}$ as:
\begin{equation}
	\mathbf{P}^{SA}_{ij} = \frac{\mathbf{P}_{ij}^{1/T}}{\sum_{j'=0}\mathbf{P}_{ij'}^{1/T}},
\end{equation}
where $T$ is a given temperature parameter. Finally, consistency learning is performed by comparing the sharpened summarized matrix with the matrix of each graph view. Formally, the objective is:
\begin{equation}
	\mathcal{L}_{GRAND} = \frac{1}{R}\sum_{r=1}^R\sum_{i\in V}||\mathbf{P}^{SA}_{i} - \mathbf{P}_{i}||,
\end{equation}
here $\mathcal{L}_{GRAND} $ serves as a regularization which is combined with the standard supervised loss. 

\smallskip
\textit{Augmentation for GNNs with the Consistency Regularization~(AugGCR)}~\citep{park2021learning}. AugGCR begins with the generation of augmented graphs by random dropout and mixup of different order features. To enhance the model generalization, it borrows the idea of meta-learning to partition the training data, which improves the quality of graph augmentation. In addition, it utilizes consistency regularization to enhance the semi-supervised node classification.

\subsection{Graph Representation Learning}\label{sec:grl}

The objective of graph classification is to predict the property of the whole graph example. Assuming that the training set comprises $N^l$ and $N^u$ graph samples $\mathcal{G}^l=\{G^1,\cdots,G^{N^l}\}$ and $\mathcal{G}^u=\{G^{N^l+1},\cdots,G^{N^l+N^u}\}$, the graph-level supervised loss for labeled data can be expressed as follows:
\begin{equation}
\mathcal{L}_{GSL}=-\frac{1}{\left|\mathcal{G}^{u}\right|} \sum_{G_{j} \in \mathcal{G}^{L}}{\mathbf{y}^j}^T log \mathbf{p}^j,
\end{equation}
where $\mathbf{y}^j$ denotes the one-hot label vector for the $j$-th sample while $\mathbf{p}^j$ denotes the predicted distribution of $G^j$. When $N^u=0$, this objective can be utilized to optimize supervised methods. However, due to the shortage of labels in graph data, supervised methods cannot reach exceptional performance in real-world applications~\citep{hao2020asgn,liu2023semi,mao2023rahnet,yi2023towards}. To tackle this, semi-supervised graph classification has been developed extensively. These approaches can be categorized into pseudo-labeling-based methods, knowledge distillation-based methods and contrastive learning-based methods. Pseudo-labeling methods annotate graph instances and utilize well-classified graph examples to update the training set~\citep{li2019semi,li2022semi,ju2024focus}.
Knowledge distillation-based methods usually utilize a teacher-student architecture, where the teacher model conducts graph representation learning without label information to extract generalized knowledge while the student model focuses on the downstream task. Due to the restricted number of labeled instances, the student model transfers knowledge from the teacher model to prevent overfitting~\citep{sun2019infograph,hao2020asgn}. Another line of this topic is to utilize graph contrastive learning, which is frequently used in unsupervised learning. Typically, these methods extract topological information from two perspectives (i.e., different perturbation strategies and graph encoders), and maximize the similarity of their representations compared with those from other examples~\citep{ju2022tgnn,luo2022dualgraph,ju2022ghnn}. Active learning, as a prevalent technique to improve the efficiency of data annotation, has also been utilized for semi-supervised methods~\citep{hao2020asgn,xie2022active}. Then, we review these methods in detail.  

\smallskip
\textit{SEmi-supervised grAph cLassification~(SEAL)}~\citep{li2019semi}. SEAL treats each graph example as a node in a hierarchical graph. It builds two graph classifiers which generate graph representations and conduct semi-supervised graph classification respectively. SEAL employs a self-attention module to encode each graph into a graph-level representation, and then conducts message passing from a graph level for final classification. SEAL can also be combined with cautious iteration and active iteration. The former merely utilizes partial graph samples to optimize the parameters in the first classifier due to the potential erroneous pseudo-labels. The second combines active learning with the model, which increases the annotation efficiency in semi-supervised scenarios.  

\smallskip
\textit{InfoGraph}~\citep{sun2019infograph}. Infograph is the first contrastive learning-based method. It maximizes the similarity between summarized graph representations and their node representations. In particular, it generates node representations using the message passing mechanism and summarizes these node representations into a graph representation. Let $\Phi(\cdot,\cdot)$ denote a discriminator to distinguish whether a node belongs to the graph, and we have:

\begin{equation}
\mathcal{L}_{InfoGraph}= \sum_{j=1}^{|\mathcal{G}^l|+|\mathcal{G}^u|}\sum_{i\in \mathcal{G}_j} \left[-\operatorname{sp}\left(-\Phi\left(\mathbf{h}_i^j, \mathbf{z}^j\right)\right)\right]-\frac{1}{|N_i^j|}\sum_{i'^{j'} \in N_i^j} \left[\operatorname{sp}\left(\Phi\left(\mathbf{h}_{i'}^{j'}, \mathbf{z}^j\right)\right)\right],
\end{equation}
where $\operatorname{sp}(\cdot)$ denotes the softplus function. $N_i^j$ denotes the negative node set where nodes are not in $G^j$. This mutual information maximization formulation is originally developed for unsupervised learning and it can be simply extended for semi-supervised graph classification. In particular, InfoGraph utilizes a teacher-student architecture that compares the representation across the teacher and student networks. The contrastive learning objective serves as a regularization by combining with supervised loss. 

\smallskip
\textit{Dual Space Graph Contrastive Learning~(DSGC)}~\citep{yang2022dual}. DSGC is a representative contrastive learning-based method. It utilizes two graph encoders. The first is a standard GNN encoder in the Euclidean space and the second is the hyperbolic GNN encoder. The hyperbolic GNN encoder first converts graph embeddings into hyperbolic space and then measures the distance based on the length of geodesics. DSGC compares graph embeddings in the Euclidean space and hyperbolic space. Assuming the two GNNs are named as $f_1(\cdot)$ and $f_2(\cdot)$, the positive pair is denoted as:
\begin{equation}
\begin{array}{c}\mathbf{z}^j_{E\rightarrow H}=\exp _{\mathbf{o}}^{c}(f_1(G^j)), \\ 
\mathbf{z}^j_{H}=\exp _{\mathbf{o}}^{c}\left(f_2(G^j)\right).\end{array}
\end{equation}

Then it selects one labeled sample and $N_B$ unlabeled sample $G^j$ for graph contrastive learning in the hyperbolic space. In formulation,

\begin{equation}
\begin{aligned} \mathcal{L}_{DSGC} &=-\log \frac{\mathrm{e}^{d^H\left(\mathbf{h}^i_H, \mathbf{z}^i_{E \rightarrow H}\right) / \tau}}{\mathbf{e}^{d^H\left(\mathbf{z}^i_H, \mathbf{z}^i_{E \rightarrow H}\right) / \tau}+\sum_{i=1}^{N} \mathrm{e}^{d_{\mathbb{D}}\left(\mathbf{z}^i_{E \rightarrow H}, \mathbf{z}^j_{H}\right) / \tau}} \\ &-\frac{\lambda_{u}}{N} \sum_{i=1}^{N} \log \frac{\mathrm{e}^{d_{\mathbb{D}}^{u}\left(\mathbf{z}^j_{H}, \mathbf{z}^j_{E \rightarrow H}\right) / \tau}}{\mathrm{e}^{d_{\mathbb{D}}^{u}\left(\mathbf{z}^j_{H}, \mathbf{z}^j_{E \rightarrow H}\right) / \tau}+\mathrm{e}^{d_{\mathbb{D}}\left(\mathbf{z}^i_H, \mathbf{z}^j_{E \rightarrow H}\right) / \tau}}, \end{aligned}
\end{equation}
where $\mathbf{z}^i_{E \rightarrow H}$ and $\mathbf{z}^i_{H}$ denote the embeddings for labeled graph sample $G^i$ and $d^H(\cdot)$ denotes a distance metric in the hyperbolic space. This contrastive learning objective maximizes the similarity between embeddings learned from two encoders compared with other samples. 
Finally, the contrastive learning objective can be combined with the supervised loss to achieve effective semi-supervised contrastive learning.

\smallskip
\textit{Active Semi-supervised Graph Neural Network~(ASGN)}~\citep{hao2020asgn}. ASGN utilizes a teacher-student architecture with the teacher model focusing on representation learning and the student model targeting at molecular property prediction. In the teacher model, ASGN first employs a message passing neural network to learn node representations under the reconstruction task and then borrows the idea of balanced clustering to learn graph-level representations in a self-supervised fashion. In the student model, ASGN utilizes label information to monitor the model training based on the weights of the teacher model. In addition, active learning is also used to minimize the annotation cost while maintaining sufficient performance. 
Typically, the teacher model seeks to provide discriminative graph-level representations without labels, which transfer knowledge to the student model to overcome the potential overfitting in the presence of label scarcity. 

\smallskip
\textit{Twin Graph Neural Networks~(TGNN)}~\citep{ju2022tgnn}. TGNN also uses two graph neural networks to give different perspectives to learn graph representations. Differently, it adopts a graph kernel neural network to learn graph-level representations in virtue of random walk kernels. Rather than directly enforcing representation from two modules to be similar, TGNN exchanges information by contrasting the similarity structure of the two modules. In particular, it constructs a list of anchor graphs, $G^{a_1},G^{a_2},\cdots,G^{a_M}$, and utilizes two graph encoders to produce their embeddings, i.e., $\left\{z^{a_{m}}\right\}_{m=1}^{M} $, $\left\{w^{a_{m}}\right\}_{m=1}^{M}$. Then it calculates the similarity distribution between each unlabeled graph and anchor graphs for two modules. Formally,
\begin{equation}
    p_{m}^{j}=\frac{\exp \left(\cos \left(z^j, z^{a_{m}}\right) / \tau\right)}{\sum_{m^{\prime}=1}^{M} \exp \left(\cos \left(z^j, z^{a_{m^{\prime}}}\right) / \tau\right)},
\end{equation}
\begin{equation}
q_{m}^{j}=\frac{\exp \left(\cos \left(\mathbf{w}^{j}, \mathbf{w}^{a_{m}}\right) / \tau\right)}{\sum_{m^{\prime}=1}^{M} \exp \left(\cos \left(\mathbf{w}^{j}, \mathbf{w}^{a_{m^{\prime}}}\right) / \tau\right)}.
\end{equation}

Then, TGNN minimizes the distance between distributions from different modules as follows:
\begin{equation}
\mathcal{L}_{TGNN}=\frac{1}{\left|\mathcal{G}^{U}\right|} \sum_{G^{j} \in \mathcal{G}^{u}} \frac{1}{2}\left(D_{\mathrm{KL}}\left(\mathbf{p}^{j} \| \mathbf{q}^{j}\right)+D_{\mathrm{KL}}\left(\mathbf{q}^{j} \| \mathbf{p}^{j}\right)\right),
\end{equation}
which serves as a regularization term to combine with the supervised loss.

\subsection{Summary}

This section introduces semi-supervised learning for graph representation learning and we provide the summary as follows:

\begin{itemize}
    \item \textbf{Techniques.} Classic node classification aims to conduct transductive learning on graphs with access to unlabeled data, which is a natural semi-supervised problem. Semi-supervised graph classification aims to relieve the requirement of abundant labeled graphs. Here, a variety of semi-supervised methods have been put forward to achieve better performance under the label scarcity. Typically, they try to integrate semi-supervised techniques such as active learning, pseudo-labeling, consistency learning, and consistency learning with graph representation learning.
    \item \textbf{Challenges and Limitations.} Despite their great success, the performance of these methods is still unsatisfactory, especially in graph-level representation learning. For example, DSGC can only achieve an accuracy of 57\% in a binary classification dataset REDDIT-BINARY. Even worse, label scarcity is often accompanied by unbalanced datasets and potential domain shifts, which provides more challenges from real-world applications. 
    \item \textbf{Future Works.} In the future, we expect that these methods can be applied to different problems such as molecular property predictions. There are also works to extend graph representation learning in more realistic scenarios like few-shot learning~\citep{ma2020adaptive,chauhan2020few}. A higher accuracy is always anticipated for more advanced and effective semi-supervised techniques.  
\end{itemize}

\section{Graph Self-supervised Learning}


Besides supervised or semi-supervised methods, self-supervised learning (SSL) also has shown its powerful capability in data mining and representation embedding in recent years. In this section, we investigated Graph Neural Networks based on SSL and provided a detailed introduction to a few typical models. Graph SSL methods usually have a unified pipeline, which includes pretext tasks and downstream tasks. 
Pretext tasks help the model encoder to learn better representation, as a premise of better performance in downstream tasks.
So a delicate design of pretext task is crucial for Graph SSL. We would firstly introduce the overall framework of Graph SSL in Section~\ref{self_overall_framework}, then introduce the two kinds of pretext task design, generation-based methods and contrast-based methods respectively in Section~\ref{self_generation} and \ref{self_contrast}. A summarisation is provided in Table~\ref{table_self_supervised}.

\begin{table}[t]
\renewcommand\arraystretch{0.935}
\caption{Summary of methods for self-supervised Learning on Graphs. "PT", "CT" and "UFE" mean "Pre-training", "Collaborative Train" and "Unsupervised Feature Extracting" respectively.}
\label{table_self_supervised}
\centering
\resizebox{\textwidth}{!}{
\begin{tabular}{c|c|c|c|c|c}
\toprule
  & Approach  & Augmentation Scheme & Training Scheme & Generation Target & Objective Function \\
\midrule
  \multirow{5}*{Generation-based} & Graph Completion~\citep{you2020does}  & Feature Mask  &PT/CT &Node Feature &-\\
           & AttributeMask~\citep{jin2020self} &  Feature Mask &PT/CT  &PCA Node Feature &- \\
           & AttrMasking~\citep{hu2019strategies} &  Feature Mask &PT &Node/Edge Feature &- \\
           & MGAE~\citep{wang2017mgae} & No Augmentation & CT &Node Feature &- \\
           & GAE~\citep{kipf2016variational} & Feature Noise & UFE & Adjacency Matrix &- \\
           
\midrule
\multirow{9}*{Contrast-based} & DeepWalk~\citep{perozzi2014deepwalk} & Random Walk & UFE & - & SkipGram  \\
& LINE~\citep{tang2015line} & Random Walk & UFE & - & Jensen-Shannon \\
& GCC~\citep{qiu2020gcc} & Random Walk & PT/URL & - & InfoNCE \\
& SimGCL~\citep{yu2022graph} & Embedding Noise & UFE & - & InfoNCE \\
& SimGRACE~\citep{xia2022simgrace} & Model Noise & UFE & - & InfoNCE \\
& GCA~\citep{zhu2021graph} & \makecell{Feature Masking \&\\ Structure Adjustment} & URL & - & InfoNCE \\
& BGRL~\citep{grill2020bootstrap} & \makecell{Feature Masking \&\\ Structure Adjustment} & URL & - & BYOL \\      
\bottomrule

\end{tabular}
}
\end{table}

\subsection{Overall framework}
\label{self_overall_framework}
Consider a featured graph $\mathcal{G}$, we denote a graph encoder $f$ to learn the representation of the graph, and a pretext decoder $g$ with specific architecture in different pretext tasks. Then the pretext self-supervised learning loss can be formulated as:
\begin{equation}
    \mathcal{L}_{total} = E_{\mathcal{G}\sim\mathcal{D}}[\mathcal{L}_{ssl}(g, f, \mathcal{G})],
\end{equation}
where $\mathcal{D}$ denotes the distribution of featured graph $\mathcal{G}$. By minimizing
$\mathcal{L}_{overall}$, we can learn encoder $f$ with capacity to produce high-quality embedding.
As for downstream tasks, we denote a graph decoder $d$ which transforms the output of graph encoder $f$ into model prediction. The loss of downstream tasks can be formulated as:
\begin{equation}
    \mathcal{L}_{sup} = \mathcal{L}_{sup}(d, f, \mathcal{G}; y),
\end{equation}
where $y$ is the ground truth in downstream tasks. We can obverse that $\mathcal{L}_{sup}$ is a typical supervised loss. To ensure the model achieves wise graph representation extraction and optimistic prediction performance, $\mathcal{L}_{ssl}$ and $\mathcal{L}_{sup}$ have to be minimized simultaneously.
We introduce 3 different ways to minimize the two loss functions:

\textbf{Pre-training}. This strategy has two steps. In pre-training step, the $\mathcal{L}_{ssl}$ is minimized to get $g^*$ and $f^*$:
\begin{equation}
    g^*, f^* = \underset{g, f}{\arg\min}\mathcal{L}_{ssl}(g, f, \mathcal{D}).\label{ssl:ssl loss}
\end{equation}

Then the parameter of $f^*$ is kept to continue training in pretext supervised learning progress. The supervised loss is minimized to get the final parameters of $f$ and $d$.
\begin{equation}
    \underset{d, f}{\min}\mathcal{L}_{ssl}(d, f|_{f_0 = f^*}, \mathcal{G}; y).
\end{equation}

\textbf{Collaborative Train}. In this strategy, $\mathcal{L}_{ssl}$ and $\mathcal{L}_{sup}$ are optimized simultaneously. A hyperparameter $\alpha$ is used to balance the contribution of pretext task loss and downstream task loss. The overall minimization strategy is like the traditional supervised strategy with a pretext task regularization:
\begin{equation}
    \underset{g, f, d}{\min} [\mathcal{L}_{ssl}(g, f, \mathcal{G})+\alpha\mathcal{L}_{sup}(d, f, \mathcal{G}; y)].
\end{equation}

\textbf{Unsupervised Feature Extracting}. This strategy is similar to the Pre-training and Fine-tuning strategy in the first step to minimize pretext task loss $\mathcal{L}_{ssl}$ and get $f^*$. However, when minimizing downstream loss $\mathcal{L}_{sup}$, the encoder $f^*$ is fixed. Also, the training graph data are on the same dataset, which differs from the Pre-training and Fine-tuning strategy. The formulation is defined as:
\begin{equation}
    g^*, f^* = \underset{g, f}{\arg\min}\mathcal{L}_{ssl}(g, f, \mathcal{D}), \\
\end{equation}
\begin{equation}
    \underset{d}{\min}\mathcal{L}_{sup}(d, f^*, \mathcal{G}; y). \\
\end{equation}

\subsection{Generation-based pretext task design}
\label{self_generation}
If a model with an encoder-decoder structure can reproduce certain graph features from an incomplete or perturbed graph, it indicates the encoder has the ability to extract useful graph representation.
This motivation is derived from Autoencoder~\citep{hinton2006reducing} which originally learns on image dataset.
In such a case, Eq. \ref{ssl:ssl loss} can be rewritten as:
\begin{equation}
    \underset{g, f}{\min}\mathcal{L}_{ssl}(g (f (\hat{\mathcal{G}})), \mathcal{G}),
\end{equation}
where $f (\cdot)$ and $g (\cdot)$ stand for the representation encoder and rebuilding decoder.
However, feature information and structure information are both important compositions suitable to be rebuilt for graph datasets. So generation-based pretext can be divided into two categories: feature rebuilding and structure rebuilding. We introduce several outstanding models in the following part.

Graph Completion~\citep{you2020does} is one of the representative methods of feature rebuilding. They mask some node features to generate an incomplete graph. Then the pretext task is set as predicting the removed node features. As shown in Eq. \ref{ssl:Graph Completion}, this method can be formulated as a special case of Eq. \ref{ssl:Graph Completion}, letting $\mathcal{\hat{G}} = (A, \hat{X})$ and replacing $\mathcal{G}\xrightarrow{}X$. 
The loss function is often Mean Squared Error or Cross Entropy, depending on whether the feature is continuous or binary.
\begin{equation}
    \underset{g,f}{\min}\ \mathbf{MSE}(g (f (\hat{\mathcal{G}})), \mathbf{X}).\label{ssl:Graph Completion}
\end{equation}

Other works make some changes to the feature settings. For example, AttrMasking~\citep{hu2019strategies} aims to rebuild both node representation and edge representation, AttributeMask~\citep{jin2020self} preprocess $X$ firstly by PCA to reduce the complexity of rebuilding features.

On the other hand, MGAE~\citep{wang2017mgae} modifies the original graph by adding noise in node representation, motivated by denoising autoencoder~\citep{vincent2010stacked}. As shown in Eq. \ref{ssl:Graph Completion}, we can also consider MGAE as an implement of Eq. \ref{ssl:ssl loss} where $\mathcal{\hat{G}} = (A, \hat{X})$ and $\mathcal{G}\xrightarrow{}X$. $\hat{X}$ stands for perturbed node representation. Since the noise is independent and random, the encoder is more robust to feature input.
\begin{equation}
    \underset{g,f}{\min}\ \mathbf{BCE}(g (f (\hat{\mathcal{G}})), \mathbf{A}).\label{ssl:GAE}
\end{equation}

As for structure rebuilding methods, GAE~\citep{kipf2016variational} is the simplest instance, which can be regarded as an implement of Eq. \ref{ssl:ssl loss} where $\mathcal{\hat{G}} = \mathcal{G}$ and $\mathcal{G}\xrightarrow{}A$. $A$ is the adjacency matrix of the graph. Similar to feature rebuilding methods, GAE compresses raw node representation vectors into low-dimensional embedding with its encoder. Then the adjacency matrix is rebuilt by computing node embedding similarity. The loss function is set to the error between the ground-truth adjacency matrix and the recovered one, to help the model rebuild the correct graph structure. Other feature rebuilding methods~\citep{zang2023hierarchical} and structure rebuilding methods~\citep{tan2023s2gae, wen2023graph} are also increasingly being developed across numerous related publications.







\subsection{Contrast-Based pretext task design}
\label{self_contrast}
The mutual information maximization principle, which implements self-supervising by predicting the similarity between the two augmented views, forms the foundation of contrast-based approaches. Since mutual information represents the degree of correlation between two samples, we can maximize it in augmented pairs and minimize it in random-selected pairs.

The contrast-based graph SSL taxonomy can be formulated as Eq. \ref{ssl:contrast}. The discriminator that calculates the similarity of sample pairs is indicated by pretext decoder $g$. $\mathcal{G}^{(1)}$ and $\mathcal{G}^{(2)}$ are two variants of $G$ that have been augmented. 
Since graph contrastive learning methods differ from each other in 1) view generation, 2) MI estimation method we introduce this methodology in these two perspectives.
\begin{equation}
    \underset{g,f}{\min}
    \mathcal{L}_{ssl}(g [f (\hat{\mathcal{G}}^{(1)}), f (\hat{\mathcal{G}}^{(2)})]).\label{ssl:contrast}
\end{equation}

The domain of contrastive-based graph SSL is witnessing an expanding body of work in a growing number of methods~\citep{ji2023spatio, yang2023knowledge, zhao2023self, ho2023self} and applications~\citep{gao2023rumor, farhat2023graph, xia2023automated}.

\subsubsection{View generation.}

The traditional pipeline of contrastive learning-based models first involves augmenting the graph using well-crafted empirical methods, and then maximizing the consistency between different augmentations. Drawing from methods in the computer vision domain and considering the non-Euclidean structure of graph data, typical graph augmentation methods aim to modify the graph topologically or representationally. 

Given graph $\mathcal{G}=(A,X)$, the topologically augmentation 
methods usually modify the adjacency matrix $A$, which can be formulated as:
\begin{equation}
    \hat{A} = \mathscr{T}_A(A),
\end{equation}
where $\mathscr{T}_A(\cdot)$ is the transform function of adjacency matrix. Topology augmentation methods have many variants, in which the most popular one is edge modification, given by $\mathscr{T}_A(A) = P\circ A + Q\circ(1-A)$. $P$ and $Q$ are two matrices representing edge dropping and adding respectively. Another method, graph diffusion, connect nodes with their k-hop neighbors with specific weight, defined as: $\mathscr{T}_A(A)=\sum^\infty_{k=0}\alpha_k T^k$. where $\alpha$ and $T$ are coefficient and transition matrix. Graph diffusion method can integrate broad topological information with local structure.

On the other hand, the representative augmentation modifies the node representation directly, which can be formulated as:
\begin{equation}
    \hat{X} = \mathscr{T}_X(X),
\end{equation}
usually $\mathscr{T}_X(\cdot)$ can be a simple masking operater, a.k.a. $\mathscr{T}_X(X) = M \circ X$ and $M\in \{0,1\}^{N\times D}$. Based on such mask strategy, some methods propose ways to improve performance. GCA~\citep{zhu2021graph} preserves critical nodes while giving less significant nodes a larger masking probability, where significance is determined by node centrality.

As introduced before, the paradigm of augmentation has been proven to be effective in contrastive learning view generation. However, given the variety of graph data, it is challenging to maintain semantics properly during augmentations. To preserve the valuable nature of specific graph datasets, There are currently three mainly used methods: picking by trial-and-errors, trying laborious search, or seeking domain-specific information as guidance~\citep{luo2022clear, ju2023unsupervised,luo2023self}. Such complicated augmentation methods constrain the effectiveness and widespread application of graph contrastive learning.
So many newest works question the necessity of augmentation and seek other contrastive view generation methods. 

SimGCL~\citep{yu2022graph} is one of the outstanding works challenging the effectiveness of graph augmentation. The author finds that noise can be a substitution to augmentation to produce graph views in specific tasks such as recommendation. 
After doing an ablation study about augmentation and InfoNCE~\citep{xie2022self}, they find that the InfoNCE loss, not the augmentation of the graph, is what makes the difference. It can be further explained by the importance of distribution uniformity.
Contrastive learning enhances model representation ability by intensifying two characteristics: The alignment of features from positive samples and the uniformity of the normalized feature distribution. SimGCL directly adds random noises to node embeddings as augmentation, to control the uniformity of the representation distribution more effectively:
\begin{equation}
\begin{aligned}
    \textbf{e}^{(1)}_i = \textbf{e}_i + \epsilon^{(1)} * \mathbf{\tau}^{(1)}_i &,\quad
    \textbf{e}^{(2)}_i = \textbf{e}_i + \epsilon^{(2)} * \mathbf{\tau}^{(2)}_i,
    \\
    \epsilon &\sim \mathcal{N}(0,\sigma^2),
\end{aligned}
\end{equation}
where $\textbf{e}_i$ is a node representation in embedding space,  $\mathbf{\tau}^{(1)}_i$ and $\mathbf{\tau}^{(2)}_i$ are two random sampled unit vector. The experiment results indicate that SimGCL performs better than its graph augmentation-based competitors, while training time is significantly decreased.

SimGRACE~\citep{xia2022simgrace} is another graph contrastive learning framework without data augmentation. Motivated by the observation that despite encoder disruption, graph data can effectively maintain their semantics, SimGRACE takes GNN with its modified version as an encoder to produce two contrastive embedding views by the same graph input.
For GNN encoder $f(\cdot;\theta)$, the two contrastive embedding views $\textbf{e}, \textbf{e}'$  can be computed by:
\begin{equation}
\begin{aligned}
    \textbf{e}^{(1)} = f(\mathcal{G}; \theta),\quad
    &\textbf{e}^{(2)} = f(\mathcal{G}; \theta + \epsilon \cdot \Delta\theta),
    \\
    \Delta\theta_l &\sim \mathcal{N}(0,\sigma_l^2),
\end{aligned}
\end{equation}
where $\Delta\theta_l$ represents GNN parameter perturbation $\Delta\theta$ in the $l$th layer. SimGRACE can improve alignment and uniformity simultaneously, proving its capacity to produce high-quality embedding.


\subsubsection{MI estimation method.}

The mutual information $I(x, y)$ measures the information that x and y share, given a pair of random variables $(x, y)$. As discussed before, mutual information is a significant component of the contrast-based method by formulating the loss function. Mathematically rigorous MI is defined on the probability space, we can formulate mutual information between a pair of instances $(x_i,x_j)$ as:
\begin{equation}
\begin{aligned}
    I(x, y) &= D_{KL} (p(x,y) || p(x)p(y)) \\
    &= E_{p(x,y)} [\log{\frac{p(x,y)}{p(x)p(y)}}].\label{ssl:mi}
\end{aligned}
\end{equation}

However, directly computing Eq. \ref{ssl:mi} is quite difficult, so we introduce several different types of estimation for MI:

\textbf{InfoNCE}. Noise-contrastive estimator is a widely used lower bound MI estimator. Given a positive sample $y$ and several negative sample $y'_i$, a noise-contrastive estimator can be formulated as \citep{zhu2020deep}\citep{qiu2020gcc}:
\begin{equation}
    \mathcal{L} = -I(x,y) = -E_{p(x,y)} [\log{\frac{e^{g(x,y)}}{e^{g(x,y)} + \sum_i{e^{g(x,y'_i)}}}}],
    \label{ssl:infoNCE}
\end{equation}
usually the kernal function $g(\cdot)$ can be cosine similarity or dot product.

\textbf{Triplet Loss}. Intuitively, we can aim to create a distinct separation in the degree of similarity, ensuring that positive samples are closer together and negative samples are further apart by a certain distance. So we can define the loss function in the following manner~\citep{jiao2020sub}:
\begin{equation}
    \mathcal{L} = E_{p(x,y)} [{\max(g(x,y) - g(x,y')+\epsilon, 0)}],
\end{equation}
where $\epsilon$ is a hyperparameter. This function is straightforward to compute.

\textbf{BYOL Loss}. Estimation without negative samples is investigated by BYOL~\citep{grill2020bootstrap}. The estimator is Asymmetrically structured:
\begin{equation}
    \mathcal{L} = E_{p(x,y)} [2-2 \frac{g(x)\cdot y}{\|g(x)\|\|y\|}],
\end{equation}
note that encoder $g$ should keep the dimension of input and output the same.

\subsection{Summary}
This section introduces graph self-supervised learning and we provide the summary as follows:

\begin{itemize}
    \item \textbf{Techniques.} Differing from classic supervised and semi-supervised learning, self-supervised learning increases a model's generalization ability and robustness while decreasing reliance on labels. Graph SSL utilizes pretext tasks to extract inherent information from representation distributions. Typical Graph SSL methods can be divided into generation-based and contrast-based approaches. Generation-based methods learn an encoder with the ability to reconstruct a graph as precisely as possible, motivated by the principles of Autoencoder. Contrast-based methods, which have recently attracted significant interest, involve learning an encoder to minimize mutual information between relevant instances and maximize mutual information between unrelated instances.
    \item \textbf{Challenges and Limitations.} Although graph SSL has achieved superior performance in many tasks, its theoretical basis is not so solid. Many well-known methods are validated only through experiments, without providing theoretical explanations or mathematical proofs. It is imperative to establish a strong theoretical foundation for graph SSL.
    \item \textbf{Future Works.} In the future we expect more graph ssl methods designed essentially by theoretical proof, without dedicated designed augment process or pretext tasks by intuition. This will bring us more definite mathematical properties and a less ambiguous empirical sense. Also, graphs are a prevalent form of data representation across diverse domains, yet obtaining manual labels can be prohibitively expensive. Expanding the applications of graph SSL to broader fields is a promising avenue for future research.
\end{itemize}

\section{Graph Structure Learning}
Graph structure determines how node features propagate and affect each other, playing a crucial role in graph representation learning. In some scenarios the provided graph is incomplete, noisy, or even has no structure information at all. Recent research also finds that graph adversarial attacks (i.e., modifying a small number of node features or edges), can degrade learned representations significantly. These issues motivate graph structure learning (GSL), which aims to learn a new graph structure to produce optimal graph representations. According to how edge connectivity is modeled, there are three different approaches in GSL, namely metric-based approaches, model-based approaches, and direct approaches. Besides edge modeling, regularization is also a common trick to make the learned graph satisfy some desired properties. We first present the basic framework and regularization methods for GSL in Sec. \ref{sec:gsl_framework} and Sec. \ref{sec:regularization}, respectively, and then introduce different categories of GSL in Sec. \ref{sec:gsl_metricbased}, \ref{sec:gsl_modelbased} and \ref{sec:gsl_direct}. We summarize GSL approaches in Table \ref{tab:GSL}.

\subsection{Overall Framework}
\label{sec:gsl_framework}
We denote a graph by $\mathcal{G} = (\mathbf{A}, \mathbf{X})$, where $\mathbf{A} \in \mathbb{R}^{N \times N}$ is the adjacency matrix and $\mathbf{X} \in \mathbb{R}^{N \times M}$ is the node feature matrix with $M$ being the dimension of each node feature. A graph encoder $f_\theta$ learns to represent the graph based on node features and graph structure for task-specific objective $\mathcal{L}_{t}(f_\theta(\mathbf{A}, \mathbf{X})) $. In the GSL setting, there is also a graph structure learner which aims to build a new graph adjacency matrix $\mathbf{A}^*$ to optimize the learned representation. Besides the task-specific objective, a regularization term can be added to constrain the learned structure. So the overall objective function of GSL can be formulated as
\begin{equation}
    \min_{\theta, \mathbf{A}^*} \mathcal{L} = \mathcal{L}_t(f_\theta(\mathbf{A}^*, \mathbf{X})) + \lambda  \mathcal{L}_r(\mathbf{A}^*, \mathbf{A}, \mathbf{X}),
\end{equation}
where $\mathcal{L}_t$ is the task-specific objective, $\mathcal{L}_r$ is the regularization term and $\lambda$ is a hyperparameter for the weight of regularization.

\begin{table}
    \caption{Summary of graph structure learning methods.}
    \label{tab:GSL}
    \centering
    \resizebox{0.85\linewidth}{!} {
    \begin{tabular}{cccccc}
    \toprule
        &\multirow{2}{*}{Method} &\multirow{2}{*}{Structure Learning} & \multicolumn{3}{c}{Regularization}  \\
        & & & Sparsity &Low-rank & Smoothness \\
    \midrule
    \multirow{7}{*}{\rotatebox{90}{Metric-based}}&AGCN~\citep{li2018adaptive}&Mahalanobis distance& \\
    &GRCN~\citep{yu2021graph} & Inner product & \checkmark \\
    &CAGCN~\citep{zhu2020cagnn} & Inner product & \checkmark \\
    &GNNGUARD~\citep{zhang2020gnnguard} & Cosine similarity & \\
    &IDGL~\citep{chen2020iterative} & Cosine similarity & \checkmark & \checkmark & \checkmark \\
    &HGSL~\citep{zhao2021heterogeneous} & Cosine similarity & \checkmark \\
    &GDC~\citep{gasteiger2019diffusion}  & Graph diffusion & \checkmark \\
    \midrule
    \multirow{8}{*}{\rotatebox{90}{Model-based}} & GLN~\citep{pilco2019graph} & Recurrent blocks &\\
    &GLCN~\citep{jiang2019semi} & One-layer neural network & \checkmark & & \checkmark \\
    &NeuralSparse~\citep{zheng2020robust}& Multi-layer neural network& \checkmark \\
    &GAT~\citep{velivckovic2017graph} & Self-attention &\\
    &GaAN~\citep{zhang2018gaan} & Gated attention\\
    &hGAO~\citep{gao2019graphkdd} & Hard attention & \checkmark\\
    &VIB-GSL~\citep{sun2022graph} & Dot-product attention & \checkmark\\
    &MAGNA~\citep{wang2020multi} & Graph attention diffusion\\
    \midrule
    \multirow{4}{*}{\rotatebox{90}{Direct}} & GLNN~\citep{gao2020exploring}& MAP estimation&\checkmark & & \checkmark \\
    &Pro-GNN~\citep{jin2020graph} & Direct optimization & \checkmark & \checkmark & \checkmark \\
    &GSML~\citep{wan2021graph}& Bilevel optimization &\checkmark \\
    &LSD-GNN~\citep{franceschi2019learning}& Bilevel optimization & \\
    &BGCNN~\citep{zhang2019bayesian}& Bayesion optimization \\
    &VGCN~\citep{elinas2020variational}& Stochastic variational inference & \\
    \bottomrule
    \end{tabular}}
\end{table}

\subsection{Regularization}
\label{sec:regularization}
The goal of regularization is to constrain the learned graph to satisfy some properties by adding some penalties to the learned structure. The most common properties used in GSL are sparsity, low lank, and smoothness.

\subsubsection{Sparsity}
Noise or adversarial attacks will introduce redundant edges into graphs and degrade the quality of graph representation. An effective technique to remove unnecessary edges is sparsity regularization, i.e., adding a penalty on the number of nonzero entries of the adjacency matrix ($\ell_0$-norm)~\citep{yu2021graph,zhao2021heterogeneous,zheng2020robust,wan2021graph}:
\begin{equation}
    \mathcal{L}_{sp} = \lVert \mathbf{A} \rVert_0,
\end{equation}
however, $\ell_0$-norm is not differentiable so optimizing it is difficult, and in many cases $\ell_1$-norm is used instead as a convex relaxation. Other methods to impose sparsity include pruning and discretization~\citep{zhu2020cagnn,franceschi2019learning}. These processes are also called postprocessing since they usually happen after the adjacency matrix is learned. Pruning removes part of the edges according to some criteria~\citep{zhu2020cagnn}. For example, edges with weights lower than a threshold, or those not in the top-K edges of nodes or graphs. Discretization is applied to generate graph structure by sampling from some distribution~\citep{franceschi2019learning}. Compared to directly learning edge weights, sampling enjoys the advantage of controlling the generated graph, but has issues during optimizing since sampling itself is discrete and hard to optimize. Reparameterization and Gumbel-softmax are two useful techniques to overcome such issues, and are widely adopted in GSL.

\subsubsection{Low Rank}
In real-world graphs, similar nodes are likely to group together and form communities, which should lead to a low-rank adjacency matrix. Recent work also finds that adversarial attacks tend to increase the rank of the adjacency matrix quickly~\citep{chen2020iterative,jin2020graph}. Therefore, low-rank regularization is also a useful tool to make graph representation learning more robust:
\begin{equation}
    \mathcal{L}_{lr} = Rank(\mathbf{A}).
\end{equation}

It is hard to minimize matrix rank directly. A common technique is to optimize the nuclear norm, which is a convex envelope of the matrix rank:
\begin{equation}
    \mathcal{L}_{nc} = \lVert \mathbf{A} \rVert_{*} = \sum_i^N \sigma_i,
\end{equation}
where $\sigma_i$ are singular values of $\mathbf{A}$. Entezari et al. replaces the learned adjacency matrix with rank-r approximation by singular value decomposition (SVD) to achieve robust graph learning against adversarial attacks. 

\subsubsection{Smoothness}
A common assumption is that connected nodes share similar features, or in other words, the graph is ``smooth'' as the difference between local neighbors is small~\citep{chen2020iterative,jiang2019semi,gao2020exploring,jin2020graph}. The following metric is a natural way to measure graph smoothness:
\begin{equation}
    \mathcal{L}_{sm} = \frac{1}{2}\sum_{i,j=1}^N A_{ij}(x_i-x_j)^2 = tr(\mathbf{X}^\top \mathbf{(D-A)X}) = tr(\mathbf{X}^\top \mathbf{LX}),
\end{equation}
where $\mathbf{D}$ is the degree matrix of $\mathbf{A}$ and $\mathbf{L} = \mathbf{D-A}$ is called graph Laplacian. A variant is to use the normalized graph Laplacian $\widehat{\mathbf{L}} = \mathbf{D}^{-\frac{1}{2}}\mathbf{LD}^{-\frac{1}{2}}$.

\subsection{Metric-based Methods}
\label{sec:gsl_metricbased}
Metric-based methods measure the similarity between nodes as the edge weights. They follow the basic assumption that similar nodes tend to have connections with each other. We show some representative works

\smallskip
\textit{Adaptive Graph Convolutional Neural Networks~(AGCN)}~\citep{li2018adaptive}. AGCN learns a task-driven adaptive graph during training to enable a more generalized and flexible graph representation model. After parameterizing the distance metric between nodes, AGCN is able to adapt graph topology to the given task. It proposes a generalized Mahalanobis distance between two nodes with the following formula:
\begin{equation}
    \mathbb{D}(x_i, x_j) = \sqrt{(x_i - x_j)^\top M (x_i - x_j)},
\end{equation}
where $M = W_d W_d^\top$ and $W_d$ is the trainable weights to minimize task-specific objective. Then the Gaussian kernel is used to obtain the adjacency matrix:
\begin{align}
    \mathbb{G}_{ij} &= \exp(-\mathbb{D}(x_i, x_j)/(2\sigma^2)),  \\
    \hat{A} &= normalize(\mathbb{G}).
\end{align}

\smallskip
\textit{Graph-Revised Convolutional Network~(GRCN)}~\citep{yu2021graph}. GRCN uses a graph revision module to predict missing edges and revise edge weights through joint optimization on downstream tasks. It first learns the node embedding with GCN and then calculates pair-wise node similarity with the dot product as the kernel function.
\begin{align}
    Z &= GCN_g(A, X), \\
    S_{ij} &= \left \langle z_i,z_j \right \rangle.
\end{align}

The revised adjacency matrix is the residual summation of the original adjacency matrix $\hat{A} = A + S$. GRCN also applies a sparsification technique on the similarity matrix $S$ to reduce computation cost:
\begin{equation}
    S^{(K)}_{ij} = \left\{
    \begin{aligned}
        S_{ij}, &~~S_{ij} \in topK(S_i)\\
        0, &~~S_{ij} \notin topK(S_i)
    \end{aligned}
    \right..
\end{equation}

Threshold pruning is also a common strategy for sparsification. For example, CAGCN~\citep{zhu2020cagnn} uses dot product to measure node similarity, and refines the graph structure by removing edges between nodes whose similarity is less than a threshold $\tau_r$ and adding edges between nodes whose similarity is greater than another threshold $\tau_a$.

\smallskip
\textit{Defending Graph Neural Networks
against Adversarial Attacks~(GNNGuard)}~\citep{zhang2020gnnguard}. GNNGuard measures similarity between a node $u$ and its neighbor $v$ in the $k$-th layer by cosine similarity and normalizes node similarity at the node level within the neighborhood as follows:
\begin{align}
  s_{uv}^k &= \frac{h_u^k \odot h_v^k}{\Vert h_u^k \Vert_2 \Vert h_v^k \Vert_2}, \\
  \alpha_{uv}^k &= \left\{
    \begin{aligned}
        &s_{uv}^k / \sum\nolimits_{v \in \mathcal{N}_u} s_{uv}^k \times \hat{N}_u^k / (\hat{N}_u^k + 1), &~~if~~u \ne v\\
        &1/(\hat{N}_u^k + 1), &~~if~~u = v
    \end{aligned}
    \right.,
\end{align}
where $\mathcal{N}_u$ denotes the neighborhood of node $u$ and $\hat{N}_u^k = \sum\nolimits_{v \in \mathcal{N}_u} \Vert s_{uv}^k \Vert_0$. To stabilize GNN training, it also proposes a layer-wise graph memory by keeping part of the information from the previous layer in the current layer. Similar to GNNGuard, IDGL~\citep{chen2020iterative} uses multi-head cosine similarity and mask edges with node similarity smaller than a non-negative threshold, and HGSL~\citep{zhao2021heterogeneous} generalizes this idea to heterogeneous graphs.

\smallskip
\textit{Graph Diffusion Convolution~(GDC)}~\citep{gasteiger2019diffusion}. GDC replaces the original adjacency matrix with generalized graph diffusion matrix $\mathbf{S}$:
\begin{equation}
    \mathbf{S} = \sum_{k=0}^\infty \theta_k \mathbf{T}^k,
\end{equation}
where $\theta_k$ is the weighting coefficient and $\mathbf{T}$ is the generalized transition matrix. To ensure convergence, GDC further requires that $\sum_{k=0}^\infty \theta_k = 1$ and the eigenvalues of $\mathbf{T}$ lie in $[0, 1]$. The random walk transition matrix $\mathbf{T}_{rw} = \mathbf{AD}^{-1}$ and the symmetric transition matrix $\mathbf{T}_{sym} = \mathbf{D}^{-1/2}\mathbf{AD}^{-1/2}$ are two examples. This new graph structure allows graph convolution to aggregate information from a larger neighborhood. The graph diffusion acts as a smoothing operator to filter out underlying noise. However, in most cases graph diffusion will result in a dense adjacency matrix $S$, so sparsification technology like top-k filtering and threshold filtering will be applied to graph diffusion. Following GDC, there are some other graph diffusion proposed. For example, AdaCAD~\citep{lim2021class} proposes Class-Attentive Diffusion, which further considers node features and aggregates nodes probably of the same class among K-hop neighbors. Adaptive diffusion convolution~(ADC)~\citep{zhao2021adaptive} learns the optimal neighborhood size via optimizing a bi-level problem.
                                                                 
\subsection{Model-based Methods}
\label{sec:gsl_modelbased}
Model-based methods parameterize edge weights with more complex models like deep neural networks. Compared to metric-based methods, model-based methods offer greater flexibility and expressive power. 

\smallskip
\textit{Graph Learning Network~(GLN)}~\citep{pilco2019graph}. GLN proposes a recurrent block to first produce intermediate node embeddings and then merge them with adjacency information as the output of this layer to predict the adjacency matrix for the next layer. Specifically, it uses convolutional graph operations to extract node features, and creates a local-context embedding based on node features and the current adjacency matrix:
\begin{align}
    &H_{int}^{(l)} = \sum_{i=1}^k \sigma_l (\tau (A^{(l)})H^{(l)}W_i^{(l)}), \\
    &H_{local}^{(l)} = \sigma_l (\tau (A^{(l)}) H_{int}^{(l)} U^{(l)}),
\end{align}
where $W_i^{(l)}$ and $U^{(l)}$ are the learnable weights. GLN then predicts the next adjacency matrix as follows:
\begin{equation}
    A^{(l+1)} = \sigma_l (M^{(l)} \alpha_l (H_{local}^{(l)}) {M^{(l)}}^\top).
\end{equation}

Similarly, GLCN~\citep{jiang2019semi} models graph structure with a softmax layer over the inner product between the difference of node features and a learnable vector. NeuralSparse~\citep{zheng2020robust} uses a multi-layer neural network to generate a learnable distribution from which a sparse graph structure is sampled. PTDNet~\citep{luo2021learning} prunes graph edges with a multi-layer neural network and penalizes the number of non-zero elements to encourage sparsity. 

\smallskip
\textit{Graph Attention Networks~(GAT)}~\citep{velivckovic2017graph}. Besides constructing a new graph to guide the message passing and aggregation process of GNNs, many recent researchers also leverage the attention mechanism to adaptively model the relationship between nodes. GAT is the first work to introduce the self-attention strategy into graph learning. In each attention layer, the attention weight between two nodes is calculated as the Softmax output on the combination of linear and non-linear transform of node features: 

\begin{align}
    e_{ij} &= a(\mathbf{W}\vec{h}_i, \mathbf{W}\vec{h}_j), \\
    \alpha_{ij} &= \frac{exp(e_{ij})}{\sum_{k\in \mathcal{N}_i} exp(e_{ik})},
\end{align}
where $\mathcal{N}_i$ denotes the neighborhood of node $i$,$\mathbf{W}$ is learnable linear transform and $a$ is pre-defined attention function. In the original implementation of GAT, $a$ is a single-layer neural network with $\mathrm{LeakyReLU}$:
\begin{equation}
    a(\mathbf{W}\vec{h}_i, \mathbf{W}\vec{h}_j) = \mathrm{LeakyReLU}(\vec{\mathrm{a}}^\top[\mathbf{W}\vec{h}_i || \mathbf{W}\vec{h}_j]).
\end{equation} 

The attention weights are then used to guide the message-passing phase of GNNs:
\begin{equation}
    \vec{h}^\prime_i = \sigma(\sum_{j \in \mathcal{N}_i}\alpha_{ij} \mathbf{W}\vec{h}_j),
\end{equation}
where $\sigma$ is a nonlinear function. It is beneficial to concatenate multiple heads of attention together to get a more stable and generalizable model, so-called multi-head attention. The attention mechanism serves as a soft graph structure learner which captures important connections within node neighborhoods. Following GAT, many recent works propose more effective and efficient graph attention operators to improve performance. GaAN~\citep{zhang2018gaan} adds a soft gate at each attention head to adjust its importance. MAGNA~\citep{wang2020multi} proposes a novel graph attention diffusion layer to incorporate multi-hop information. One drawback of graph attention is that the time and space complexities are both $O(N^3)$. hGAO~\citep{gao2019graphkdd} performs hard graph attention by limiting node attention to its neighborhood. VIB-GSL~\citep{sun2022graph} adopts the information bottleneck principle to guide feature masking in order to drop task-irrelevant information and preserve actionable information for the downstream task. 

\subsection{Direct Methods}
\label{sec:gsl_direct}
Direct methods treat edge weights as free learnable parameters. These methods enjoy more flexibility but are also more difficult to train. The optimization is usually carried out in an alternating way, i.e., iteratively updating the adjacency matrix $\mathbf{A}$ and the GNN encoder parameters $\theta$.  

\smallskip
\textit{GLNN}~\citep{gao2020exploring}. GLNN uses MAP estimation to learn an optimal adjacency matrix for a joint objective function including sparsity and smoothness. Specifically, it targets at finding the most probable adjacency matrix $\hat{A}$ given graph node features $x$:
\begin{equation}
    \tilde{A}_{MAP}(x) = \mathop{\mathrm{argmax}}\limits_{\hat{A}} f(x|\hat{A})g(\hat{A}),
\end{equation}
where $f(x|\hat{A})$ measures the likelihood of observing $x$ given $\hat{A}$, and $g(\hat{A})$ is the prior distribution of $\hat{A}$. GLNN uses sparsity and property constraint as prior, and defines the likelihood function $f$ as:
\begin{align}
    f(x|\hat{A}) &= exp(-\lambda_0 x^\top L x) \\
                 &= exp(-\lambda_0 x^\top (I - \hat{A}) x),
\end{align}
where $\lambda_0$ is a parameter. This likelihood imposed a smoothness assumption on the learned graph structure. Some other works also model the adjacency matrix in a probabilistic manner. Bayesian GCNN~\citep{zhang2019bayesian} adopts a Bayesian framework and treats the observed graph as a realization from a family of random graphs. Then it estimates the posterior probablity of labels given the observed graph adjacency matrix and features with Monte Carlo approximation. VGCN~\citep{elinas2020variational} follows a similar formulation and estimates the graph posterior through stochastic variational inference. Pro-GNN~\citep{jin2020graph} learns a clean graph structure from perturbed data and optimizes parameters for a robust GNN, leveraging properties like sparsity, low rank, and feature smoothness.

\smallskip
\textit{Graph Sparsification via Meta-Learning~(GSML)}~\citep{wan2021graph}. GSML formulates GSL as a meta-learning problem and uses bi-level optimization to find the optimal graph structure. The goal is to find a sparse graph structure that leads to high node classification accuracy at the same time given labeled and unlabeled nodes. To achieve this, GSML makes the inner optimization as training on the node classification task, and targets the outer optimization at the sparsity of the graph structure, which formulates the following bi-level optimization problem:
\begin{align}
    \hat{G}^* &= \mathop{\mathrm{min}}\limits_{\hat{G} \in \Phi(G)} L_{sps} (f_{\theta^*}(\hat{G}), Y_U), \\
    s.t. ~&~ \theta^* = \mathop{\mathrm{arg min}}\limits_{\theta} L_{train} (f_\theta(\hat{G}), Y_L).
\end{align}

In this bi-level optimization problem, $\hat{G} \in \Phi(G)$ are the meta-parameters and optimized directly without parameterization. Similarly, LSD-GNN~\citep{franceschi2019learning} also uses bi-level optimization. It models graph structure with a probability distribution over the graph and reformulates the bi-level program in terms of the continuous distribution parameters.

\subsection{Summary}
In this section, we provide the summary as follows:
\begin{itemize}
    \item \textbf{Techniques.} GSL aims to learn an optimized graph structure for better graph representations. It is also used for more robust graph representation against adversarial attacks. According to the way of edge modeling, we categorize GSL into three groups: metric-based methods, model-based methods, and direct methods. Regularization is also a commonly used principle to make the learned graph structure satisfy specific properties including sparsity, low-rank and smoothness.
    \item \textbf{Challenges and Limitations.} 
     Since there is no way to access the ground truth or optimal graph structure as training data, the learning objective of GSL is either indirect (e.g., performance on downstream tasks) or manually designed (e.g., sparsity and smoothness). Therefore, the optimization of GSL is difficult and the performance is not satisfying. In addition, many GSL methods are based on homophily assumption, i.e., similar nodes are more likely to connect with each other. However, many other types of connection exist in the real world which impose great challenges for GSL. 
    \item \textbf{Future Works.} In the future we expect more efficient and generalizable GSL methods to be applied to large-scale and heterogeneous graphs. Most existing GSL methods focus on pair-wise node similarities and thus struggle to scale to large graphs. Besides, they often learn homogeneous graph structure, but in many scenarios graphs are heterogeneous.
\end{itemize}


\section{Social Analysis}
In the real world, there usually exist complex relations and interactions between people and multiple entities. Taking people, concrete things, and abstract concepts in society as nodes and taking the diverse, changeable, and large-scale connections between data as links, we can form massive and complex social information as social networks~\citep{tabassum2018social,camacho2020four}. 
Compared with traditional data structures such as texts and forms, modeling social data as graphs has many benefits. Especially with the arrival of the "big data" era, more and more heterogeneous information is interconnected and integrated, and it is difficult and uneconomical to model this information with a traditional data structure. The graph is an effective implementation for information integration, as it can naturally incorporate different types of objects and their interactions from heterogeneous data sources~\citep{shi2016survey,moscato2021survey}.  A summarization of social analysis applications is provided in Table \ref{tab:SA}.

\begin{table*}[t]
    \caption{A summarization of social analysis applications}
    \label{tab:SA}
    \centering
    \resizebox{1\linewidth}{!}{
    \begin{tabular}{cccm{2cm}<{\centering}m{6cm}}
    \toprule
       Social networks  &  Node type & Edge type  & Applications & \multicolumn{1}{c}{References}  \\
    \midrule
      \multirow{8}{*}{\makecell{Academic\\Social\\Network}}  & \multirow{8}{2cm}{\textit{Author}, \textit{Publication}, \textit{Venue}, \textit{Organization}, \textit{Keyword}} &  \multirow{8}{2cm}{\textit{Authorship}, \textit{Co-Author}, \textit{Advisor-advisee}, \textit{Citing}, \textit{Cited}, \textit{Co-Citing}, \textit{Publishing}} & \makecell{Classification/\\Clustering} &Paper/author classification~\citep{dong2017metapath2vec,wang2019heterogeneous,zhang2019heterogeneous,qiao2020tree}, name disambiguation~\citep{zhang2018name,qiao2019unsupervised,chen2020conna,ma2023author} \\
      \cmidrule{4-5} 
      & &  &  \makecell{Relationship\\prediction} & Co-authorship ~\citep{chuan2018link,cho2018link,zhu2023predicting}, citation relationship ~\citep{yu2012citation,jiang2018cross,wang2020deep}, advisor-advisee relationship~\citep{liu2019shifu2,zhao2018identifying,luo2023impact} \\\cmidrule{4-5} 
      & &  & \makecell{Recommen-\\dation}  & Collaborator recommendation~\citep{liu2018context,kong2017exploring,kong2016exploiting}, paper recommendation~\citep{bai2019scientific,sugiyama2010scholarly,dai2023heterogeneous}, venue recommendation~\citep{yu2018pave,margaris2019handling}\\
        \midrule
      \multirow{6}{*}{\makecell{Social\\Media\\Network}}  & \multirow{6}{2cm}{\textit{User}, \textit{Blog}, \textit{Article}, \textit{Image}, \textit{Video}} &  \multirow{6}{2cm}{\textit{Following}, \textit{Like}, \textit{Unlike}, \textit{Clicked}, \textit{Viewed}, \textit{Commented}, \textit{Reposted}} & \makecell{Anomaly\\detection} & Malicious attacks~\citep{sun2020deepdom,liu2018heterogeneous,sadhasivam2023malicious}, emergency detection~\citep{bian2020rumor,dahou2023social,li2023detecting}, and robot discovery~\citep{feng2021botrgcn,lu2023sybilhp} \\\cmidrule{4-5} 
      & &  &  \makecell{Sentiment\\analysis} & Customer feedback~\citep{rosa2018knowledge, zhang2014explicit,uma2023opinion}, public events~\citep{unankard2014predicting, manguri2020twitter,bouadjenek2023user} \\\cmidrule{4-5} 
      & &  &  \makecell{Influence\\analysis} & Important node finding~\citep{domingos2001mining,richardson2002mining}, information diffusion modeling~\citep{panagopoulos2020multi, keikha2020influence,zhang2022network, kumar2022influence}   \\
      \midrule
      \multirow{5}{*}{\makecell{Location-based\\Social\\Network}}  & \multirow{5}{2cm}{\textit{Restaurant}, \textit{Cinema}, \textit{Mall}, \textit{Parking}} &  \multirow{5}{2cm}{\textit{Friendship}, \textit{Check-in}} & POI recommendation  & 
Spatial/temporal influence~\citep{si2019adaptive, wang2022graph,zhao2020go}, social relationship~\citep{xu2021novel,long2023decentralized}, textual information~\citep{xu2021exploring,wang2023point,wang2023statrl}  \\ \cmidrule{4-5} 
         & &  &  \makecell{Urban\\computing} & Traffic congestion prediction~\citep{jiang2022graph,xiong2018predicting}, urban mobility analysis~\citep{yildirimoglu2018identification,cao2021resolving}, event detection~\citep{yu2021deep,sofuoglu2022gloss}\\
    \bottomrule
    \end{tabular}}
\end{table*}

\subsection{Concepts of Social Networks}
A social network is usually composed of multiple types of nodes, link relationships, and node attributes, which inherently include rich structural and semantic information. 
Specifically, a social network can be homogeneous or heterogeneous and directed or undirected in different scenarios. Without loss of generality, we define the social network as a directed heterogeneous graph $G=\{V,E,\mathcal{T},\mathcal{R}\}$, where $V=\{n_i\}^{|V|}_{i=1}$ is the node set, $E=\{e_i\}^{|E|}_{i=1}$ is the edge set, $\mathcal{T}=\{t_i\}^{|\mathcal{T}|}_{i=1}$ is the node type set, and $\mathcal{R}=\{r_i\}^{|\mathcal{R}|}_{i=1}$ is the edge type set. 
Each node $n_i\in V$ is associated with a node type mapping: $\phi_n(n_i) = t_j: V\longrightarrow \mathcal{T}$ and each edge $e_i\in E$ is associated with a node type mapping: $\phi_e(e_i) = r_j: E\longrightarrow \mathcal{R}$. 
A node $n_i$ may have a feature set, where the feature space is specific for the node type. 
An edge $e_i$ is also represented by node pairs $(n_j,n_k)$ at both ends and can be directed or undirected with relation-type-specific attributes. If $|\mathcal{T}|=1$ and $|\mathcal{R}|=1$, the social network is a homogeneous graph; otherwise, it is a heterogeneous graph.

Almost any data produced by social activities can be modeled as social networks, for example, the academic social network produced by academic activities such as collaboration and citation, the online social network produced by user following and followed on social media, and the location-based social network produced by human activities on different locations.
Based on constructing social networks, researchers have new paths to data mining, knowledge discovery, and multiple application tasks on social data. 
Exploring social networks also brings new challenges. One of the critical challenges is how to succinctly represent the network from the massive and heterogeneous raw graph data, that is, how to learn continuous and low-dimensional social network representations, so as to researchers can efficiently perform advanced machine learning techniques on the social network data for multiple application tasks, such as analysis, clustering, prediction, and knowledge discovery.
Thus, graph representation learning on the social network becomes the foundational technique for social analysis. 

\subsection{Academic Social Network}

Academic collaboration is a common and important behavior in academic society, and also a major way for scientists and researchers to innovate and breakthrough scientific research, which leads to social relationship between scholars.
The academic data generated by academic collaboration usually contains a large number of interconnected entities with complex relationships~\citep{kong2019academic,zhou2020survey}.
Normally, in an academic social network, the node type set consists of \textit{Author}, \textit{Publication}, \textit{Venue}, \textit{Organization}, \textit{Keyword}, etc., and the relation set consists of \textit{Authorship}, \textit{Co-Author}, \textit{Advisor-advisee}, \textit{Citing}, \textit{Cited}, \textit{Co-Citing}, \textit{Publishing}, \textit{Co-Word}, etc.
Note that in most social networks,
each relation type always connects two fixed node types with a fixed direction. For example, the relation \textit{Authorship} points from the node type \textit{Author} to \textit{Publication}, and the \textit{Co-Author} is an undirected relation between two nodes with type \textit{Author}. 
Based on the node and relation types in an academic social network, one can divide it into multiple categories. For example, the co-author network with nodes of \textit{Author} and relations of \textit{Co-Author}, the citation network with nodes of \textit{Publication} and relation of \textit{Citing}, and the academic heterogeneous information graph with multiple academic node and relation types. 
Many research institutes and academic search engines, such as Aminer\footnote{https://www.aminer.cn/}, DBLP\footnote{https://dblp.uni-trier.de/}, Microsoft Academic Graph (MAG)\footnote{https://www.microsoft.com/en-us/research/project/microsoft-academic-graph/}, have provided open academic social network datasets for research purposes.

There are multiple applications of graph representation learning on the academic social network. Roughly, they can be divided into three categories--academic entity classification/clustering, academic relationship prediction, and academic resource recommendation.

\begin{itemize}
    \item Academic entities usually belong to different classes of research areas. Research of academic entity classification and clustering aims to categorize these entities, such as papers and authors, into different classes~\citep{dong2017metapath2vec,wang2019heterogeneous,zhang2019heterogeneous,qiao2020tree,ju2023glcc,yi2023redundancy}. In literature, academic networks such as Cora, citepSeer, and Pubmed~\citep{sen2008collective} have become the most widely used benchmark datasets for examining the performance of graph representation learning models on paper classification. Also, the author name disambiguation problem~\citep{zhang2018name,qiao2019unsupervised,chen2020conna,ma2023author} is also essentially a node clustering task on co-author networks and is usually solved by the graph representation learning technique.
    \item Academic relationship prediction represents the link prediction task on various academic relations. Typical applications are co-authorship prediction~\citep{chuan2018link,cho2018link,zhu2023predicting} and citation relationship prediction~\citep{yu2012citation,jiang2018cross,wang2020deep}. Existing methods learn representations of authors and papers and use the similarity between two nodes to predict the link probability. Besides, some work~\citep{liu2019shifu2,zhao2018identifying,luo2023impact} studies the problem of advisor-advisee relationship prediction in the collaboration network.
    \item Various academic recommendation systems have been introduced to retrieve academic resources for users from large amounts of academic data in recent years. For example, collaborator recommendation~\citep{liu2018context,kong2017exploring,kong2016exploiting} benefit researchers by finding suitable collaborators under particular topics; paper recommendation~\citep{bai2019scientific,sugiyama2010scholarly,dai2023heterogeneous} help researchers find relevant papers on given topics; venue recommendation~\citep{yu2018pave,margaris2019handling} help researchers choose appropriate venues when they submit papers.
\end{itemize}

\subsection{Social Media Network}
With the development of the Internet in decades, various online social media have emerged in large numbers and greatly changed people's traditional social models. 
People can establish friendships with others beyond the distance limit and share interests, hobbies, status, activities, and other information among friends.
These abundant interactions on the Internet form large-scale complex social media networks, also named online social networks. 
Usually, in an academic social network, the node type set consists of \textit{User}, \textit{Blog}, \textit{Article}, \textit{Image}, \textit{Video}, etc., and the relation type set consists of \textit{Following}, \textit{Like}, \textit{Unlike}, \textit{Clicked}, \textit{Viewed}, \textit{Commented}, \textit{Reposted}, etc. 
The main property of a social media network is that it usually contains multi-mode information on the nodes, such as video, image, and text. Also, the relations are more complex and multiplex, including the explicit relations such as \textit{Like} and \textit{Unlike} and the implicit relations such as \textit{Clicked}. The social media network can be categorized into multiple types based on their media categories. For example, the friendship network, the movie review network, and the music interacting network are extracted from different social media platforms. In a broad sense, the user-item networks in online shopping system can also be viewed as social media networks as they also exist on the Internet and contains rich interactions by people. There are many widely used data sources for social media network analysis, such as Twitter, Facebook, Weibo, YouTube, and Instagram. 

The mainstream application research on social media networks via graph representation learning techniques mainly includes anomaly detection, sentiment analysis, and influence analysis. 
\begin{itemize}
    \item Anomaly detection aims to find strange or unusual patterns in social networks, which has a wide range of application scenarios, such as malicious attacks~\citep{sun2020deepdom,liu2018heterogeneous,sadhasivam2023malicious}, emergency detection~\citep{bian2020rumor,dahou2023social}, and robot discovery~\citep{feng2021botrgcn,lu2023sybilhp} in social networks. Unsupervised anomaly detection usually learns a reconstructed graph to detect those nodes with higher reconstructed error as the anomaly nodes~\citep{ahmed2021graph,zhao2022graph}; Supervised methods model the problem as a binary classification task on the learned graph representations~\citep{meng2021semi,zheng2019addgraph}.
    \item Sentiment analysis, also named as opinion mining, is to mine the sentiment, opinions, and attitudes, which can help enterprises understand customer feedback on products~\citep{rosa2018knowledge, zhang2014explicit,uma2023opinion} and help the government analyze the public emotion and make rapid response to public events~\citep{unankard2014predicting, manguri2020twitter,bouadjenek2023user}. The graph representation learning model is usually combined with RNN-based~\citep{zhang2019aspect,chen2020joint} or Transformer-based~\citep{albadani2022transformer,tang2020dependency} text encoders to incorporate both the user relationship and textual semantic information.
    \item Influence analysis usually aims to find several nodes in a social network to initially spread information such as advertisements, so as to maximize the final spread of information~\citep{domingos2001mining,richardson2002mining}. The core challenge is to model the information diffusion process in the social network. Deep learning methods~\citep{panagopoulos2020multi, keikha2020influence,zhang2022network, kumar2022influence} usually leverage graph neural networks to learn node embeddings and diffusion probabilities between nodes.
\end{itemize}

\subsection{Location-based Social Network}
Locations are the fundamental information of human social activities. With the wide availability of mobile Internet and GPS positioning technology, people can easily acquire their precise locations and socialize with their friends by sharing their historical check-ins on the Internet. This opens up a new avenue of research on location-based social network analysis, which gathered significant attention from the user, business, and government perspectives. 
Usually, in a location-based social network, the node type set consists of \textit{User}, and \textit{Location}, also named \textit{Point of Interest}(POI) in the recommendation scenario containing multiple categories such as \textit{Restaurant}, \textit{Cinema}, \textit{Mall}, \textit{Parking}, etc. The relation type set consists of \textit{Friendship}, \textit{Check-in}.
Also, those node and relation types that exist in traditional social media networks can be included in a location-based social network.   
The difference with other social networks, the main location-based social networks are spatial and temporal, making the graph representation learning more challenging.
For example, in a typical social network constructed for the POI recommendation, the user nodes are connected with each other by their friendship. The location nodes are connected by user nodes with the relations feature of timestamps. The location nodes also have a spatial relationship with each other and own have complex features, including categories, tags, check-in counts, number of users check-in, etc.
There are many location-based social network datasets, such as Foursquare\footnote{https://foursquare.com/}, Gowalla\footnote{https://www.gowalla.com/}, and Waze\footnote{https://www.waze.com/live-map/}. Also, many social media such as Twitter, Instagram, and Facebook can provide location information.

The research of graph representation learning on location-based social networks can be divided into two categories: POI recommendation for business benefits and urban computing for public management.

\begin{itemize}
    \item POI recommendation is one of the research hotspots in the field of location-based social networks and recommendation systems in recent years~\citep{islam2022survey, werneck2020survey,ju2022kernel}, which aim to utilize historical check-ins of users and auxiliary information to recommend potential favor places for users from a large of location points. Existing researches mainly integrate four essential characteristics, including spatial influence, temporal influence~\citep{si2019adaptive, wang2022graph,zhao2020go}, social relationship~\citep{xu2021novel,long2023decentralized}, and textual information~\citep{xu2021exploring,wang2023point,wang2023statrl}.
    \item Urban computing is defined as a process of analysis of the large-scale connected urban data created from city activities of vehicles, human beings, and sensors~\citep{paulos2004familiar, paulos2004ubicomp,silva2019urban}. Besides the local-based social network, the urban data also includes physical sensors, city infrastructure, traffic roads, and so on. Urban computing aims to improve the quality of public management and life quality of people living in city environments. Typical applications including traffic congestion prediction~\citep{jiang2022graph,xiong2018predicting}, urban mobility analysis~\citep{yildirimoglu2018identification,cao2021resolving}, event detection~\citep{yu2021deep,sofuoglu2022gloss}.
\end{itemize}

\subsection{Summary}
This section introduces social analysis by graph representation learning and we provide the summary as follows:

\begin{itemize}
    \item \textbf{Techniques.} 
    Social networks, generated by human social activities, such as communication, collaboration, and social interactions, typically involve massive and heterogeneous data, with different types of attributes and properties that can change over time. 
    Thus, social network analysis is a field of study that explores the techniques to understand and analyze the complex attributes, heterogeneous structures, and dynamic information of social networks. Social network analysis typically learns low-dimensional graph representations that capture the essential properties and patterns of the social network data, which can be used for various downstream tasks, such as classification, clustering, link prediction, and recommendation.

    \item \textbf{Chanllenges and Limitations.}
    Despite the structural heterogeneity in social networks (nodes and relations have different types), with the technological advances in social media, the node attributes have become more heterogeneous now, containing text, video, and images. 
    Also, the large-scale problem is a pending issue in social network analysis. The data in the social network has increased exponentially in past decades, containing a high density of topological links and a large amount of node attribute information, which brings new challenges to the efficiency and effectiveness of traditional network representation learning on the social network. 
    Lastly, social networks are often dynamic, which means the network information usually changes over time, and this temporal information plays a significant role in many downstream tasks, such as recommendations. This brings new challenges to representation learning on social networks in incorporating temporal information.
    
    \item \textbf{Future Works.}
     Recently, multi-modal big pre-training models that can fuse information from different modalities have gained increasing attention~\citep{qiao2022rpt,radford2021learning}. These models can obtain valuable information from a large amount of unlabeled data and transfer it to various downstream analysis tasks. Moreover, Transformer-based models have demonstrated better effectiveness than RNNs in capturing temporal information.
     In the future, there is potential for introducing multi-modal big pre-training models in social network analysis.  Also, it is important to make the models more efficient for network information extraction and use lightweight techniques like knowledge distillation to further enhance the applicability of the models. These advancements can lead to more effective social network analysis and enable the development of more sophisticated applications in various domains.

\end{itemize}

\def \m{\mathbf{m}}
\def \q{\mathbf{q}}
\def \w{\mathbf{w}}
\def \r{\mathbf{r}}
\def \a{\mathbf{a}}
\def \b{\mathbf{b}}
\def \d{\mathbf{d}}
\def \x{\mathbf{x}}
\def \y{\mathbf{y}}
\def \s{\mathbf{s}}
\def \e{\mathbf{e}}
\def \u{\mathbf{u}}
\def \bmu{\mathbf{u}}
\def \t{\mathbf{t}}
\def \z{\mathbf{z}}
\def \h{\mathbf{h}}
\def \bu{\mathbf{u}}

\def \BA{\mathbf{A}}
\def \BK{\mathbf{K}}
\def \BB{\mathbf{B}}
\def \BX{\mathbf{X}}
\def \BY{\mathbf{Y}}
\def \BI{\mathbf{I}}
\def \BC{\mathbf{C}}
\def \BD{\mathbf{D}}
\def \BE{\mathbf{E}}
\def \BH{\mathbf{H}}
\def \BL{\mathbf{L}}
\def \BM{\mathbf{M}}
\def \BG{\mathbf{G}}
\def \BP{\mathbf{P}}
\def \BQ{\mathbf{Q}}
\def \BS{\mathbf{S}}
\def \BU{\mathbf{U}}
\def \BV{\mathbf{V}}
\def \BW{\mathbf{W}}
\def \BZ{\mathbf{Z}}
\def \bmA{\mathbf{A}}
\def \bmH{\mathbf{H}}
\def \bmL{\mathbf{L}}
\def \bmX{\mathbf{X}}
\def \bmY{\mathbf{Y}}
\def \bmI{\mathbf{I}}
\def \bmW{\mathbf{W}}
\def \bmZ{\mathbf{Z}}
\newcommand{\CAT}{\kw{CONCAT}}
\newcommand{\AGG}{\text{AGG}}
\newcommand{\SUM}{\kw{SUM}}
\newcommand{\FLAT}{\kw{Flatten}}
\newcommand{\agt}{\text{AGG}}
\newcommand{\agg}{\text{AGG}}
\newcommand{\mlp}{\text{MLP}}

\section{Molecular Property Prediction}
Molecular Property Prediction is an essential task in computational drug discovery and cheminformatics. Traditional quantitative structure property/activity relationship (QSPR/QSAR) approaches are based on either SMILES or fingerprints \citep{mikolov2013distributed,xu2017seq2seq,zhang2018seq3seq}, largely overlooking the topological features of the molecules. To address this problem, graph representation learning has been widely applied to molecular property prediction. A molecule can be represented as a graph where nodes stand for atoms and edges stand for atom-bonds (ABs). Graph-level molecular representations are learned via the message passing mechanism to incorporate the topological information. The representations are then utilized for the molecular property prediction tasks. 

Specifically, a molecule is denoted as a topological graph $\mathcal{G}=(\mathcal{V}, \mathcal{E})$, where $\mathcal{V}=\{v_i | i=1,\ldots,|\mathcal{G}|\}$ is the set of nodes representing atoms. A feature vector $\mathbf{x}_i$ is associated with each node $v_i$ indicating its type such as Carbon, Nitrogen. $\mathcal{E}=\{e_{i j}| i, j=1,\ldots,|\mathcal{G}|\}$ is the set of edges connecting two nodes (atoms) $v_i$ and $v_j$ representing atom bonds. Graph representation learning methods are used to obtain the molecular representation $\mathbf{h}_{\mathcal{G}}$. Then downstream classification or regression layers $f(\cdot)$ are applied to predict the probability of target property of each molecule $y=f(\mathbf{h}_{\mathcal{G}})$.

In Section \ref{sec:property}, we introduce 4 types of molecular properties graph representation learning can be treated and their corresponding datasets. Section \ref{sec:gnn} reviews the graph representation learning backbones applied to molecular property prediction. Strategies for training the molecular property prediction methods are listed in Section \ref{sec:training}.

\subsection{Molecular Property Categorization}
\label{sec:property}
Plenty of molecular properties can be predicted by graph-based methods. We follow \citet{wieder2020a} to categorize them into 4 types: quantum chemistry, physicochemical properties, biophysics, and biological effect.

Quantum chemistry is a branch of physical chemistry focused on the application of quantum mechanics to chemical systems, including conformation, partial charges and energies. QM7, QM8, QM9 \citep{wu2018moleculenet}, COD \citep{ruddigkeit2012enumeration} and CSD \citep{groom2016the} are datasets for quantum chemistry prediction.

Physicochemical properties are the intrinsic physical and chemical characteristics of a substance, such as bioavailability, octanol solubility, aqueous solubility and hydrophobicity. ESOL, Lipophilicity and Freesolv \citep{wu2018moleculenet} are datasets for physicochemical properties prediction.

Biophysics properties are about the physical underpinnings of biomolecular phenomena, such as affinity, efficacy and activity. PDBbind \citep{wang2005the}, MUV, and HIV \citep{wu2018moleculenet} are biophysics property prediction datasets.

Biological effect properties are generally defined as the response of an organism, a population, or a community to changes in its environment, such as side effects, toxicity and ADMET. Tox21, toxcast \citep{wu2018moleculenet} and PTC \citep{toivonen2003statistical} are biological effect prediction datasets.

Moleculenet \citep{wu2018moleculenet} is a widely-used benchmark dataset for molecule property prediction. It contains over 700,000 compounds tested on different properties. For each dataset, they provide a metric and a splitting
pattern. Among the datasets, QM7, OM7b, QM8, QM9, ESOL, FreeSolv, Lipophilicity and PDBbind are regression tasks, using MAE or RMSE as the evaluation metrics. Other tasks such as tox21 and toxcast are classification tasks, using AUC as evaluation metric.

\subsection{Molecular Graph Representation Learning Backbones}
\label{sec:gnn}
Since node attributes and edge attributes are crucial to molecular representation, most works use GNN instead of traditional graph representation learning methods as backbones, since many GNN methods consider edge information. Existing GNNs designed for the general domain can be applied to molecular graphs. Table \ref{tab:gnn} summarizes the GNNs used for molecular property prediction and the types of properties they can be applied to predict.

\begin{table*}[t]
\centering
\caption{Summary of GNNs in molecular property prediction.}
\label{tab:gnn}
\resizebox{0.85\linewidth}{!} {
\begin{tabular}{l cccc}
    \toprule
    Type & Spatial/Specrtal & Method & Application \\
    \midrule
    Reccurent GNN & - & R-GNN & Biological effect \citep{scarselli2008the} \\
    \midrule
    Reccurent GNN & - & GGNN & \makecell{Quantum chemistry \citep{mansimov2019molecular}, \\ Biological effect \citep{withnall2020building,altae2017low,feinberg2018potentialnet}} \\
    \midrule
    Reccurent GNN & - &  IterRefLSTM & Biophysics \citep{altae2017low}, Biological effect \citep{altae2017low} \\
    \midrule
    Convolutional GNN & Spatial/Specrtal & GCN & \makecell{Quantum chemistry \citep{yang2019analyzing,liao2019lanczosnet,withnall2020building},\\ pysicochemical properties \citep{ryu2018deeply,duvenaud2015convolutional,coley2017convolutional},\\ Biophysics \citep{duvenaud2015convolutional,yang2019analyzing,bouritsas2022improving}\\
    Biological effect \citep{li2017learning,wu2018moleculenet}}\\
    \midrule
    Convolutional GNN & Specrtal & LanczosNet & Quantum chemistry \citep{liao2019lanczosnet}\\
    \midrule
    Convolutional GNN & Specrtal & ChebNet & \makecell{Physicochemical properties, \\Biophysics, Biological effect \citep{li2018adaptive}}\\
    \midrule
    Convolutional GNN & Spatial & GraphSAGE & \makecell{Physicochemical properties \citep{hu2019strategies}, \\Biophysics \citep{errica2019fair,liang2020mxpool,chen2020can}, \\ Biological effect \citep{hu2019strategies,ma2019graph}}\\
    \midrule
    Convolutional GNN & Spatial & GAT & \makecell{Physicochemical properties \citep{hu2019strategies, ahmad2023attention}, \\Biophysics \citep{chen2020can,bouritsas2022improving}, \\ Biological effect \citep{hu2019strategies}}\\
    \midrule
    Convolutional GNN & Spatial & DGCNN & \makecell{Biophysics \citep{chen2019powerful}, Biological effect \citep{zhang2018end}}\\
    \midrule
    Convolutional GNN & Spatial & GIN & \makecell{Physicochemical properties \citep{hu2019strategies,bouritsas2022improving}, \\Biophysics \citep{hu2019strategies,hu2020open},\\ Biological effect \citep{hu2019strategies}}\\
    \midrule
    Convolutional GNN & Spatial & MPNN & Physicochemical \citep{ma2020multi}\\
    \midrule
    Transformer & - & MAT & Physicochemical, Biophysics \citep{Jastrz2017}\\
    \bottomrule
\end{tabular}
}
\end{table*}

Furthermore, many works customize their GNN structure by considering the chemical domain knowledge. 
\begin{itemize}
    \item First, the \textbf{chemical bonds} and molecule interaction are taken into consideration carefully. For example, Ma et al. \citep{ma2020multi} use an additional edge GNN to model the chemical bonds separately. Specifically, given an edge $(v,w)$, they formulate an Edge-based GNN  as:
     \begin{align}
     {\m}_{vw}^{(k)} &= \agt_{\text{edge}}(\{{\h}_{vw}^{(k-1)}, {\h}_{uv}^{(k-1)}, {\x}_u | u \in \mathcal{N}_v \setminus w\}),\quad
     {\h}_{vw}^{(k)} = \mlp_{\text{edge}}(\{ {\m}_{vw}^{(k-1)}, {\h}_{vw}^{(0)} \}),
     \label{equ_bondmpnn}
    \end{align}
 where ${\h}_{vw}^{(0)}=\sigma({\BW}_{\text{ein}}{\e}_{vw})$ is the input state of the Edge-based GNN,  ${\BW}_{\text{ein}}\in \mathbb{R}^{d_{\text{hid}}\times d_e}$ is the input weight matrix.
    PotentialNet \citep{feinberg2018potentialnet} further uses different message passing operations for different edge types. DGNN-DDI \citep{ma2023dual} leverage dual graph neural networks to model the interaction between two molecules.
    \item Second, \textbf{motifs} in molecular graphs play an important role in molecular property prediction. GSN \citep{bouritsas2022improving} leverage substructure encoding to construct a topologically-aware message-passing method. Each node $v$ updates its state $\mathbf{h}^t_v$ by combining its previous state with the aggregated messages:
\begin{align}\label{eq:gsn}
    \mathbf{h}^{t+1}_v &=  \mathrm{UP}^{t+1}\big(\mathbf{h}^{t}_v, \mathbf{m}^{t+1}_v\big),\\
    \mathbf{m}^{t+1}_v &=\left\{
        \begin{array}{l}
         M^{t+1}([\mathbf{h}^{t}_v, \mathbf{h}^{t}_u, \mathbf{x}^V_v, \mathbf{x}^V_u, \mathbf{e}_{u,v}]_{u\in \mathcal{N}(v)}) \ (\textbf{GSN-v})\\
         \quad \quad \quad \quad \quad \quad \text{ or }\\
         M^{t+1}([\mathbf{h}^{t}_v, \mathbf{h}^{t}_u, \mathbf{x}^E_{u,v}, \mathbf{e}_{u,v}]_{u\in \mathcal{N}(v)}) \ (\textbf{GSN-e})
                 \end{array} 
                 \right.,\label{eq:msg_fn}
\end{align}
where $\mathbf{x}^V_v, \mathbf{x}^V_u, \mathbf{x}^E_{u,v}, \mathbf{e}_{u,v}$ contains the substructure information associated with nodes and edges, $[]$ denotes a multiset.
Yu et al. \citep{yu2022molecular} constructs a heterogeneous graph using motifs and molecules. Motifs and molecules are both treated as nodes and the edges model the relationship between motifs and graphs, for example, if a graph contains a motif, there will be an edge between them. MGSSL \citep{zhang2021motif} leverages a
retrosynthesis-based algorithm BRICS and additional rules to find the motifs and combines motif layers with atom layers. It is a hierarchical
framework jointly modeling atom-level information and motif-level information. Aouichaoui et al. \citep{aouichaoui2023combining} introduce group-contribution-based attention to highlight the most substructures in molecules.

    \item Third, \textbf{different feature modalities} have been used to improve molecular graph embedding. Lin et al. \citep{lin2022pisces} combine SMILES modality and graph modality with contrastive learning. Zhu et al. \citep{zhu2022unified} encode 2D molecular graph and 3D molecular conformation with a unified Transformer. It uses a unified model to learn  3D conformation generation given 2D graph and 2D graph generation given 3D conformation. Cremer et al.\citep{cremer2023equivariant} use a Equivariant Graph Neural Networks to represent the 3D information of molecules. Liu et al. \citep{liu2023interpretable} consider molecular chirality and design a chirality-aware molecular convolution module.
    \item Finally, \textbf{knowledge graph and literature} can provide additional knowledge for molecular property prediction. Fang et al. \citep{fang2022molecular} introduce a chemical element knowledge graph to summarize microscopic associations between elements and augment the molecular graph based on the knowledge graph, and a knowledge-aware message-passing network is used to encode the augmented graph. MuMo \citep{su2022molecular} introduces biomedical literature to guide molecular property prediction. It pretrains a GNN and a language model on paired data of molecules and literature mentions via contrastive learning:
    \begin{equation}
    \ell_i^{(\mathbf{z}_i^G, \mathbf{z}_i^T)} = -\log \frac{\exp{(\emph{sim}(\mathbf{z}_i^G,\mathbf{z}_i^T)/\tau)}}{\sum_{j=1}^{N}\exp{(\emph{sim}(\mathbf{z}_i^G,\mathbf{z}_j^T)/\tau})},
\end{equation}
where $\mathbf{z}_i^G,\mathbf{z}_i^T$ are the representation of molecule and its corresponding literature.
Zhao et al. \citep{zhao2023gimlet} propose a unified Transformer architecture to jointly model molecule graph and the corresponding bioassay description.
\end{itemize}

\subsection{Training strategies}
\label{sec:training}
Despite the encouraging performance achieved by GNNs, the traditional supervised training scheme of GNNs faces a severe limitation: The scarcity of available molecules with desired properties. Although there are a large number of molecular graphs in public databases such as PubChem, labeled molecules are hard to acquire due to the high cost of wet-lab experiments and quantum chemistry calculations. Directly training GNNs on such limited molecules in a supervised way is prone to over-fitting and lack of generalization. To address this issue, few-shot learning and self-supervised learning are widely used in molecular property prediction.

\textbf{Few-shot learning.} Few-shot learning aims at generalizing to a task with a small labeled data set. The prediction of each property is treated as a single task. Metric-based and optimization-based few-shot learning have been adopted for molecular property prediction. Metric-based few-shot learning is similar to nearest neighbors and kernel density estimation, which learns a metric or distance function over objects. IterRefLSTM \citep{altae2017low} leverages matching network \citep{vinyals2016matching} as the few-shot learning framework, calculating the similarity between support samples and query samples. Optimization-based few-shot learning optimizes a meta-learner for parameter initialization which can be fast adapted to new tasks. Meta-MGNN~\citep{guo2021few} adopts MAML \citep{finn2017model} to train a parameter initialization to adapt to different tasks and use self-attentive task weights for each task. PAR \citep{wang2021property} also uses MAML framework and learns an adaptive relation graph among molecules for each task.  

\textbf{Self-supervised learning.} Self-supervised learning can pre-train a GNN model with plenty of unlabeled molecular graphs and transfer it to specific molecular property prediction tasks. Self-supervised learning contains generative methods and predictive methods. Predictive methods design prediction tasks to capture the intrinsic data features. Pre-GNN \citep{hu2019strategies} exploits both node-level and graph-level prediction tasks including context prediction, attribute masking, graph-level property prediction and structural similarity prediction. MGSSL \citep{zhang2021motif} provides a motif-based generative pre-training framework making topology prediction and motif generation iteratively. Contrastive methods learn graph representations by pulling views from the same graph close and pushing views from different graphs apart. Different views of the same graph are constructed by graph augmentation or leveraging the 1D SMILES and 3D structure. MolCLR \citep{wang2022molecular} augments molecular graphs by atom masking, bond deletion and subgraph removal and maximizes the agreement between the original molecular graph and augmented graphs. Fang et al. \citep{fang2022molecular} uses a chemical knowledge graph to guide the graph augmentation. SMICLR \citep{pinheiro2022smiclr} uses contrastive learning across SMILES and 2D molecular graphs. GeomGCL~\citep{li2022geomgcl} leverages graph contrastive learning to capture the geometry of the molecule across 2D and 3D views. Jiang et al. \citep{jiang2023gode} and Fang et al. \citep{fang2023knowledge} integrate molecule graphs with chemical knowledge graph and fuse the two modalities with contrastive learning. Self-supervised learning can also be combined with few-shot learning to fully leverage the hierarchical information in the training set \citep{ju2023few}.

\subsection{Summary}

This section introduces graph representation learning in molecular property prediction and we provide the summary as follows:

\begin{itemize}
    \item \textbf{Techniques.} For molecular property prediction, a molecule is represented as a graph whose nodes are atoms and edges are atom-bonds (ABs). GNNs such as GCN, GAT, and GraphSAGE are adopted to learn the graph-level representation. The representations are then fed into a classification or regression head for the molecular property prediction tasks. Many works guide the model structure design with medical domain knowledge including chemical bond features, motif features, different modalities of molecular representation, chemical knowledge graph and literature. Due to the scarcity of available molecules with desired properties, few-shot learning and contrastive learning are used to train molecular property prediction models, so that the model can leverage the information in large unlabeled dataset and can be adapted to new tasks with a few examples.
    \item \textbf{Challenges and Limitations.} Despite the great success of graph representation learning in molecular property prediction, the methods still have limitations: 1) Few-shot molecular property prediction are not fully explored. 2) Most methods depend on training with labeled data, but neglect the chemical domain knowledge.
    \item \textbf{Future Works.} In the future, we expect that: 1) More few-shot learning and zero-shot learning methods are studied for molecular property prediction to solve the data scarcity problem. 2) Heterogeneous data can be fused for molecular property prediction. There are a large amount of heterogeneous data about molecules such as knowledge graphs, molecule descriptions and property descriptions. They can be considered to assist molecular property prediction. 3) Chemical domain knowledge can be leveraged for the prediction model. For example, when we perform affinity prediction, we can consider molecular dynamics knowledge.
\end{itemize}

\section{Molecular Generation}

Molecular generation is pivotal to drug discovery, where it serves a fundamental 
role in downstream tasks like molecular docking~\citep{meng2011molecular} and 
virtual screening~\citep{walters1998virtual}. The goal of molecular generation 
is to produce chemical structures that satisfy a specific molecular profile, 
e.g., novelty, binding affinity, and SA scores. Traditional methods have relied 
on 1D string formats like SMILES~\citep{gomez2018automatic} and SELFIES~\citep{krenn2020self}. 
With the recent advances in graph representation learning, numerous graph-based 
methods have also emerged, where molecular graph $\mathcal{G}$ can naturally embody 
both 2D topology and 3D geometry. While recent literature reviews
~\citep{meyers2021novo, du2022molgensurvey} have covered the general topics of 
molecular design, this chapter is dedicated to the applications of graph 
representation learning in the molecular generation task. Molecular generation is 
intrinsically a \textit{de novo} task, where molecular structures are generated 
from scratch to navigate and sample from the vast chemical space. Therefore, this 
chapter does not discuss tasks that restrict chemical structures 
\textit{a priori}, such as docking~\citep{ganea2021independent, stark2022equibind} 
and conformation generation~\citep{shi2021learning, zhu2022direct}. 

\subsection{Taxonomy for molecular featurization methods}

This section categorizes the different methods to feature molecules. The taxonomy 
presented here is unique to the task of molecular generation, owing to the 
various modalities of molecular entities, complex interactions with 
other bio-molecular systems and formal knowledge from the laws of chemistry and 
physics.  

\textbf{2D topology \textit{vs.} 3D geometry. }
Molecular data are multi-modal by nature. For one thing, a molecule can be unambiguously 
represented by its 2D topological graph $\mathcal{G}_\mathrm{2D}$, where atoms are nodes 
and bonds are edges. $\mathcal{G}_\mathrm{2D}$ can be encoded by canonical MPNN models like 
GCN~\citep{kipf2016semi}, GAT~\citep{velivckovic2017graph}, and R-GCN~\citep{schlichtkrull2018modeling}, in ways similar to tasks like social 
networks and knowledge graphs. A typical example of this line of work is GCPN~\citep{you2018graph}, 
a graph convolutional policy network generating molecules with desired properties such as 
synthetic accessibility and drug-likeness.  

For another, the 3D conformation of a molecule can be accurately depicted by its 
3D geometric graph $\mathcal{G}_\mathrm{3D}$, which incorporates 3D atom coordinates. 
In 3D-GNNs like SchNet~\citep{schutt2018schnet} and OrbNet~\citep{qiao2020orbnet}, 
$\mathcal{G}_\mathrm{3D}$ is organized into a $k$-NN graph or a radius graph 
according to the Euclidean distance between atoms. It is justifiable to approximate 
$\mathcal{G}_\mathrm{3D}$ as a 3D extension to $\mathcal{G}_\mathrm{2D}$, since 
covalent atoms are closest to each other in most cases. However, $\mathcal{G}_\mathrm{3D}$ 
can also find a more long-standing origin in the realm of computational chemistry
~\citep{frisch2016gaussian}, where both covalent and non-covalent atomistic 
interactions are considered to optimize the potential surface and simulate molecular 
dynamics. Therefore, $\mathcal{G}_\mathrm{3D}$ more realistically represents the 
molecular geometry, which makes a good fit for protein pocket binding and 3D-QSAR 
optimization~\citep{verma20103d}. 

Molecules can rotate and translate, affecting their position in the 3D space. 
Therefore, it is ideal to encode these molecules with GNNs equivariant/invariant 
to roto-translations, which can be $\sim10^3$ times more efficient than data augmentation~\citep{geiger2022e3nn}. 
Equivariant GNNs can be based on irreducible representation~\citep{thomas2018tensor, anderson2019cormorant, fuchs2020se, batzner2021e3equivariant, brandstetter2022geometric}, 
regular representation~\citep{finzi2020generalizing, hutchinson2021lietransformer}, 
or scalarization~\citep{schutt2018schnet, Klicpera2020Directional, liu2022spherical, kohler2020equivariant, jing2021learning, satorras2021en, huang2022constrained, pmlr-v139-schutt21a, tholke2022equivariant, klicpera2021gemnet}, 
which are explained in more detail in \citep{han2022geometrically}. 
Recent works like GraphVF~\citep{sun2023graphvf} and MolCode~\citep{zhang2023equivariant} have been incorporating $\mathcal{G}_\mathrm{2D}$ and $\mathcal{G}_\mathrm{3D}$ 
to accurately capture the relationship between structure and properties in molecular design in a unified way.

\textbf{Unbounded \textit{vs.} binding-based. }
Earlier works have aimed to generate \textbf{unbounded} molecules in either 2D 
or 3D space, striving to learn good molecular representations through this task. 
In the 2D scenario, GraphNVP~\citep{madhawa2019graphnvp} first introduces a flow-based model to 
learn an invertible transformation between the 2D chemical space and the latent 
space. GraphAF~\citep{shi2020graphaf} further adopts an autoregressive generation scheme to 
check the valence of the generated atoms and bonds. In the 3D scenario, G-SchNet 
~\citep{gebauer2019symmetry} first proposes to utilize $\mathcal{G}_\mathrm{3D}$ (instead of 3D density 
grids) as the generation backbone. It encodes $\mathcal{G}_\mathrm{3D}$ via SchNet, 
and uses an auxiliary token to generate atoms on the discretized 3D space 
autoregressively. G-SphereNet~\citep{luo2022autoregressive} uses symmetry-invariant representations 
in a spherical coordinate system (SCS) to generate atoms in the continuous 3D 
space and preserve equivariance. 

Unbounded models adopt certain techniques to optimize specific properties of the 
generated molecules. GCPN and GraphAF use scores like logP, QED, and chemical 
validity to tune the model via reinforcement learning. EDM~\citep{hoogeboom2022equivariant} 
can generate 3D molecules with property $c$ by re-training the diffusion model 
with $c$'s feature vector concatenated to the E(n) equivariant dynamics function 
$\hat{\boldsymbol{\epsilon}}_t=\phi\left(\boldsymbol{z}_t,[t, c]\right)$. cG-SchNet
~\citep{gebauer2022inverse} adopts a conditioning network architecture to jointly 
target multiple electronic properties during conditional generation without the 
need to re-train the model. RetMol~\citep{wang2022retrieval} uses a retrieval-based 
model for controllable generation. 

On the other hand, \textbf{binding-based} methods generate drug-like molecules 
(aka. ligands) according to the binding site (aka. binding pocket) of a protein 
receptor. Drawing inspirations from the lock-and-key model for enzyme action
~\citep{fischer1894einfluss}, works like LiGAN~\citep{ragoza2022generating} and 
DESERT~\citep{long2022zero} uses 3D density grids to fit the density surface 
between the ligand and the receptor, encoded by 3D-CNNs. Meanwhile, a growing 
amount of literature has adopted $\mathcal{G}_\mathrm{3D}$ for representing 
ligand and receptor molecules, because $\mathcal{G}_\mathrm{3D}$ more accurately 
depicts molecular structures and atomistic interactions both within and between 
the ligand and the receptor. Representative works include 3D-SBDD~\citep{luo20213d}, 
GraphBP~\citep{liu2022generating}, Pocket2Mol~\citep{peng2022pocket2mol}, and DiffSBDD
~\citep{schneuing2022structure}. GraphBP 
shares a similar workflow with G-SphereNet, except that the receptor atoms are also 
incorporated into $\mathcal{G}_\mathrm{3D}$ to depict the 3D geometry at the binding 
pocket. 

\textbf{Atom-based \textit{vs.} fragment-based. } 
Molecules are inherently 
hierarchical structures. At the atomistic level, molecules are represented by 
encoding atoms and bonds. At a coarser level, molecules can also be represented 
as molecular fragments like functional groups or chemical sub-structures. Both 
the composition and the geometry are fixed within a given fragment, e.g., the 
planar peptide-bond ($\text{--CO--NH--}$) structure. Fragment-based generation effectively 
reduces the degree of freedom (DOF) of chemical structures, and injects well-established 
knowledge about molecular patterns and reactivity. JT-VAE~\citep{jin2018junction} 
decomposes 2D molecular graph $\mathcal{G}_\mathrm{2D}$ into a junction-tree 
structure $\mathcal{T}$, which is further encoded via tree message-passing. 
DeepScaffold~\citep{li2019deepscaffold} expands the provided molecular scaffold 
into 3D molecules. L-Net~\citep{li2021learning} adopts a graph U-Net architecture and devises 
a custom three-level node clustering scheme for pooling and unpooling operations 
in molecular graphs. A number of works have also emerged lately for fragment-based 
generation in the binding-based setting, including 
FLAG~\citep{zhang2022molecule} and FragDiff~\citep{peng2023pocketspecific}. FLAG uses a regression-based approach to sequentially decide the 
type and torsion angle of the next fragment to be placed at the binding site, 
and finally optimizes the molecule conformation via a pseudo-force field. 
FragDiff also adopts a sequential generation process but uses a diffusion model 
to determine the type and pose of each fragment in one go. 

\subsection{Generative methods for molecular graphs}

For a molecular graph generation process, the model first learns a latent distribution 
$P(Z|\mathcal{G})$ characterizing the input molecular graphs. A new molecular graph $\hat{\mathcal{G}}$ is 
then generated by sampling and decoding from this learned distribution. 
Various models have been adopted to generate molecular graphs, including generative 
adversarial network (GAN), variational auto-encoder (VAE), 
normalizing flow (NF), diffusion model (DM), and autoregressive model (AR). 


\textbf{Generative adversarial network (GAN). } 
GAN~\citep{goodfellow2020generative} is trained to discriminate real data $\boldsymbol{x}$ 
from generated generated data $\boldsymbol{z}$, with the training object formalized as 
\begin{equation}
    \min _G \max _D \mathcal{L}(D, G)  = \mathbb{E}_{\boldsymbol{x} \sim p_{\text {data }}}[\log D(\boldsymbol{x})]+\mathbb{E}_{\boldsymbol{z} \sim p(\boldsymbol{z})}[\log (1-D(G(\boldsymbol{z})))], 
\end{equation}
where $G(\cdot)$ is the generator function and $D(\cdot)$ is the discriminator 
function. For example, MolGAN~\citep{de2018molgan} encodes $\mathcal{G}_\mathrm{2D}$ with R-GCN, 
trains $D$ and $G$ with improved W-GAN~\citep{arjovsky2017wasserstein}, and uses reinforcement learning to 
generate attributed molecules, where the score function is assigned from RDKit~\citep{landrum2013rdkit} 
and chemical validity. 

\textbf{Varaitional auto-encoder (VAE). } 
In VAE~\citep{kingma2013auto}, the decoder parameterizes the 
conditional likelihood distribution $p_\theta(\boldsymbol{x}|\boldsymbol{z})$, 
and the encoder parameterizes an approximate posterior distribution 
$q_\phi(\boldsymbol{z}|\boldsymbol{x}) \approx p_\theta(\boldsymbol{z}|\boldsymbol{x})$. 
The model is optimized by the evidence lower bound (ELBO), consisting of the reconstruction 
loss term and the distance loss term: 
\begin{equation}
    \max _{\theta, \phi}\mathcal L_{\theta, \phi}(\boldsymbol{x}):=\mathbb{E}_{\boldsymbol{z} \sim q_\phi(\cdot | \boldsymbol{x})}\left[\ln \frac{p_\theta(\boldsymbol{x}, \boldsymbol{z})}{q_\phi(\boldsymbol{z}|\boldsymbol{x})}\right]=\ln p_\theta(\boldsymbol{x})-D_\mathrm{K L}\left(q_\phi(\cdot|\boldsymbol{x}) \| p_\theta(\cdot |\boldsymbol{x})\right).
\end{equation}

Maximizing ELBO is equivalent to simultaneously maximizing the log-likelihood of 
the observed data, and minimizing the divergence of the approximate posterior 
${q_{\phi }(\cdot |x)}$ from the exact posterior ${ p_{\theta }(\cdot |x)}$. 
Representative works along this thread include JT-VAE~\citep{jin2018junction}, 
GraphVAE~\citep{simonovsky2018graphvae}, and CGVAE~\citep{liu2018constrained} for 
the 2D generation task, and 3DMolNet~\citep{nesterov20203dmolnet} for the 3D generation task. 

\textbf{Autoregressive model (AR). } 
Autoregressive model is an umbrella definition for any model that sequentially 
generates the components (atoms or fragments) of a molecule. ARs better capture 
the interdependency within the molecular structure and allows for explicit valency 
check. For each step in AR, the new component can be produced using different techniques: 
\begin{itemize}
    \item Regression/classification, such is the case with 3D-SBDD~\citep{luo20213d}, Pocket2Mol~\citep{peng2022pocket2mol}, etc. 
    \item Reinforcement learning, such is the case with L-Net~\citep{li2021learning}, DeepLigBuilder~\citep{li2021structure}, etc. 
    \item Probabilistic models like normalizing flow and diffusion. 
\end{itemize}

\textbf{Normalizing flow (NF). } Based on the change-of-variable theorem, NF~\citep{rezende2015variational} 
constructs an invertible mapping $f$ between a complex data distribution $\boldsymbol{x} \sim X$: 
and a normalized latent distribution $\boldsymbol{z} \sim Z$. Unlike VAE, which 
has juxtaposed parameters for encoder and decoder, the flow model uses the same set of 
parameter for encoding and encoding: reverse flow $f^{-1}$ for generation, and forward flow $f$ for training:  
\begin{align}
    \max _f \log p(\boldsymbol{x}) & =\log p_K\left(\boldsymbol{z}_K\right) \\
    & =\log p_{K-1}\left(\boldsymbol{z}_{K-1}\right)-\log \left|\operatorname{det}\left(\frac{d f_K\left(\boldsymbol{z}_{K-1}\right)}{d \boldsymbol{z}_{K-1}}\right)\right| \\
    & =\ldots \\
    & =\log p_0\left(\boldsymbol{z}_0\right)-\sum_{i=1}^K \log \left|\operatorname{det}\left(\frac{d f_i\left(\boldsymbol{z}_{i-1}\right)}{d \boldsymbol{z}_{i-1}}\right)\right|,
\end{align}
where $f = f_K \circ f_{K-1}\circ ... \circ f_1$ is a composite of $K$ blocks of transformation. 
While GraphNVP~\citep{madhawa2019graphnvp} generates the molecular graph with NF in one go, 
following works tend to use the autoregressive flow model, including GraphAF~\citep{shi2020graphaf}, 
GraphDF~\citep{luo2021graphdf}, GraphBP~\citep{liu2022generating} and SiamFlow~\citep{tan2023target}. 

\textbf{Diffusion model (DM). } 
Diffusion models~\citep{sohl2015deep, song2019generative, ho2020denoising} define 
a Markov chain of diffusion steps to slowly add random noise to data $\boldsymbol{x}_0 \sim q(\boldsymbol{x})$: 
\begin{align}
    q(\boldsymbol{x}_t \vert \boldsymbol{x}_{t-1}) & = \mathcal{N}(\boldsymbol{x}_t; \sqrt{1 - \beta_t} \boldsymbol{x}_{t-1}, \beta_t\boldsymbol{I}), \\
    q(\boldsymbol{x}_{1:T} \vert \boldsymbol{x}_0) & = \prod^T_{t=1} q(\boldsymbol{x}_t \vert \boldsymbol{x}_{t-1}).
\end{align}

They then learn to reverse the diffusion process to construct desired data samples 
from the noise: 
\begin{align}
    p_\theta(\boldsymbol{x}_{0:T}) & = p(\boldsymbol{x}_T) \prod^T_{t=1} p_\theta(\boldsymbol{x}_{t-1} \vert \boldsymbol{x}_t),  \\
    p_\theta(\boldsymbol{x}_{t-1} \vert \boldsymbol{x}_t) & = \mathcal{N}(\boldsymbol{x}_{t-1}; \boldsymbol{\mu}_\theta(\boldsymbol{x}_t, t), \boldsymbol{\Sigma}_\theta(\boldsymbol{x}_t, t)), 
\end{align}
while the models are trained using a variational lower bound. 
Diffusion models have been applied to generate unbounded 3D molecules in EDM~\citep{hoogeboom2022equivariant} 
and GCDM~\citep{morehead2023geometry}, 
and binding-specific ligands in DiffSBDD~\citep{schneuing2022structure}, DiffBP~\citep{lin2022diffbp} and TargetDiff~\citep{guan20233d}. Diffusion can 
also be applied to generate molecular fragments in autoregressive models, as is the case 
with FragDiff~\citep{peng2023pocketspecific}.

{
\begin{table*}[t]
\centering
\caption{Summary of molecular generation models. }
\label{tab:molgen_summary}
\resizebox{0.77\textwidth}{!}{
\begin{tabular}{l ccccc}
\toprule
Model & 2D/3D & \makecell{Binding- \\ based} & \makecell{Fragment-\\ based} & \makecell{GNN\\ Backbone} & \makecell{Generative\\ Model} \\
\midrule  

GCPN~\citep{you2018graph} & 2D & & & GCN~\citep{kipf2016semi} & GAN \\
MolGAN~\citep{de2018molgan} & 2D & & & R-GCN~\citep{schlichtkrull2018modeling} & GAN \\
DEFactor~\citep{assouel2018defactor} & 2D & & & GCN & GAN \\
GraphVAE~\citep{simonovsky2018graphvae} & 2D & & & ECC~\citep{simonovsky2017dynamic} & VAE  \\
MDVAE~\citep{du2022interpretable} & 2D & & & GGNN~\citep{li2015gated} & VAE \\
JT-VAE~\citep{jin2018junction} & 2D & & $\checkmark$ & MPNN~\citep{gilmer2017neural} & VAE \\
CGVAE~\citep{liu2018constrained} & 2D & & & GGNN & VAE \\
DeepScaffold~\citep{li2019deepscaffold} & 2D & & $\checkmark$ & GCN & VAE \\
GraphNVP~\citep{madhawa2019graphnvp} & 2D & & & R-GCN & NF \\
MoFlow~\citep{zang2020moflow} & 2D & & & R-GCN & NF \\
GraphAF~\citep{shi2020graphaf} & 2D & & & R-GCN & NF + AR \\
GraphDF~\citep{luo2021graphdf} & 2D & & & R-GCN & NF + AR  \\
L-Net~\citep{li2021learning} & 3D & & $\checkmark$ & g-U-Net~\citep{gao2019graph} & AR \\
G-SchNet~\citep{gebauer2019symmetry} & 3D & & & SchNet~\citep{schutt2018schnet} & AR \\
GEN3D~\citep{roney2021generating} & 3D & & & EGNN~\citep{satorras2021en} & AR \\
G-SphereNet~\citep{luo2022autoregressive} & 3D & & & SphereNet~\citep{liu2022spherical} & NF + AR \\
EDM~\citep{hoogeboom2022equivariant} & 3D & & & EGNN & DM \\
GCDM~\citep{morehead2023geometry} & 3D & & & GCPNet~\citep{morehead2022geometry} & DM \\
3D-SBDD~\citep{luo20213d} & 3D & $\checkmark$ & & SchNet & AR \\
Pocket2Mol~\citep{peng2022pocket2mol} & 3D & $\checkmark$ & & GVP~\citep{jing2020learning} & AR \\
FLAG~\citep{zhang2022molecule} & 3D & $\checkmark$ & $\checkmark$ & SchNet & AR \\
GraphBP~\citep{liu2022generating} & 3D & $\checkmark$ & & SchNet & NF + AR \\
SiamFlow~\citep{tan2023target} & 3D & $\checkmark$ & & R-GCN & NF  \\
DiffBP~\citep{lin2022diffbp} & 3D & $\checkmark$ & & EGNN & DM \\
DiffSBDD~\citep{schneuing2022structure} & 3D & $\checkmark$ & & EGNN & DM \\
TargetDiff~\citep{guan20233d} & 3D & $\checkmark$ & & EGNN & DM \\
FragDiff~\citep{peng2023pocketspecific} & 2D + 3D & $\checkmark$ & $\checkmark$ & MPNN & DM + AR \\
GraphVF~\citep{sun2023graphvf} & 2D + 3D & $\checkmark$ & $\checkmark$ & SchNet  & NF + AR \\
MolCode~\citep{zhang2023equivariant} & 2D + 3D & $\checkmark$ &  & EGNN & NF + AR \\

\bottomrule
\end{tabular}
}
\end{table*}
}

\subsection{Summary and prospects}
We wrap up this chapter with Table~\ref{tab:molgen_summary}, which profiles existing molecular generation models according to their taxonomy for molecular featurization, the GNN backbone, and
the generative method. This chapter covers the critical topics of molecular generation, which also elicit valuable insights into the promising directions for future research. We summarize these important aspects as follows. 

\textbf{Techniques.} Graph neural networks can be flexibly leveraged to encode molecular features on different representation levels and across different problem settings. Canonical GNNs like GCN~\citep{kipf2016semi}, GAT~\citep{velivckovic2017graph}, and R-GCN~\citep{schlichtkrull2018modeling} have been widely adopted to model 2D molecular graphs, while 3D equivariant GNNs have also been effective in modeling 3D molecular graphs. 
In particular, this 3D approach can be readily extended to binding-based scenarios, where the 3D geometry of the binding protein receptor is considered alongside the ligand geometry \textit{per se}. 
Fragment-based models like JT-VAE~\citep{jin2018junction} and L-Net~\citep{li2021learning} can also effectively capture the hierarchical molecular structure. Various generative methods have also been effectively incorporated into the molecular setting, including generative adversarial network (GAN), variational auto-encoder (VAE), autoregressive model (AR), normalizing flow (NF), and diffusion model (DM). These models have been able to generate valid 2D molecular topologies and realistic 3D molecular geometries, greatly accelerating the search for drug candidates. 

\textbf{Challenges and Limitations. } While there has been an abundant supply of unlabelled molecular structural and geometric data~\citep{irwin2012zinc, spackman2022basis, francoeur2020three}, the number of labeled molecular data over certain critical biochemical properties like toxicity~\citep{gayvert2016data} and solubility~\citep{delaney2004esol} remain very limited. On the other hand, existing models have heavily relied on expert-crafted metrics to evaluate the quality of the generated molecules, such as QED and Vina~\citep{eberhardt2021autodock}, rather than actual wet lab experiments. 

\textbf{Future Works.} Besides the structural and geometric attributes described in this chapter, an even more extensive array of data can be applied to aid molecular generation, including chemical reactions and medical ontology. These data can be organized into a heterogeneous knowledge graph to aid the extraction of high-quality molecular representations. Furthermore, high throughput experimentation (HTE) should be adopted to realistically evaluate the synthesizablity and druggability of the generated molecules in the wet lab. Concrete case studies, such as the design of potential inhibitors to SARS-CoV-2~\citep{li2021structure}, would be even more encouraging, bringing new insights into leveraging these molecular generative models to facilitate the design and fabrication of potent and applicable drug molecules in the pharmaceutical industry. 

Integrating Large Language Models (LLMs) like GPT-4~\citep{openai2023gpt4} with graph-based representations offers a promising new direction in molecular generation. Recent studies like those by \citep{jablonka2023leveraging} and \citep{guo2023indeed} highlight LLMs' potential in chemistry, especially in low-data scenarios. While current LLM-based approaches in this domain, including those by \citep{mazuz2023molecule} and \citep{bagal2021molgpt}, predominantly utilize textual SMILES strings, their potential is somewhat constrained by the limits of text-only inputs. The emerging trend, exemplified by \citep{liu2023git}, is to leverage multi-modal data, integrating graph, image, and text, which could more comprehensively capture the intricacies of molecular structures. This approach marks a significant shift towards utilizing graph-based information alongside traditional text, enhancing the capability of LLMs in molecular generation. Such advances suggest that future research should focus more on exploiting the synergy between graph-based molecular representations and the evolving landscape of LLMs to address complex challenges in chemistry and material sciences.


\section{Recommender Systems}

The use of graph representation learning in recommender systems has been drawing increasing attention as one of the key strategies for addressing the issue of information overload. With their strong ability to capture high-order connectivity between graph nodes, deep graph representation learning has been shown to be beneficial in enhancing recommendation performance across a variety of recommendation scenarios.

Typical recommender systems take the observed interactions between users and items and their fixed features as input, and are intended for making proper predictions on which items a specific user is probably interested in. To formulate, given an user set $\mathcal{U}$, an item set $\mathcal{I}$ and the interaction matrix between users and items $X\in\{0,1\}^{\left|\mathcal{U}\right|\times\left|\mathcal{I}\right|}$, where $X_{u,v}$ indicates there is an observed interaction between user $u$ and item $i$. The target of GNNs on recommender systems is to learn representations $h_u,h_i\in\mathbb{R}^d$ for given $u$ and $i$. The preference score can further be calculated by a similarity function:
\begin{equation}
    \hat x_{u,i}=f(h_u,h_i),
\end{equation}
where $f(\cdot,\cdot)$ is the similarity function, e.g. inner product, cosine similarity, multi-layer perceptrons that takes the representation of $u$ and $i$ and calculate the preference score $\hat x_{u,i}$.

When it comes to adapting graph representation learning in recommender systems, a key step is to construct graph-structured data from the interaction set $X$. Generally, a graph is represented as $\mathcal{G}=\{\mathcal{V},\mathcal{E}\}$ where $\mathcal{V},\mathcal{E}$ denotes the set of vertices and edges respectively. According to the construction of $\mathcal{G}$, we can categorize the existing works as follows into three parts which are introduced in the following subsections. A summary is provided in Table~\ref{tab:recommender_system}.

\begin{table*}[t]
\centering
\caption{Summary of graph models for recommender systems. }
\label{tab:recommender_system}
\small
\resizebox{\textwidth}{!}{
\begin{tabular}{l cccccc}
\toprule
Model & Recommendation Task & Graph Structure & Graph Encoder & Representation \\
\midrule  
GC-MC \citep{berg2017graph} & Matrix Completion & User-Item Graph & GCN & Last-Layer \\
NGCF \citep{wang2019neural} & Collaborative Filtering & User-Item Graph & GCN+Affinity & Concatenation  \\
MMGCN \citep{wei2019mmgcn} & Micro-Video & Multi-Modal Graph & GCN & Last-Layer \\
LightGCN \citep{he2020lightgcn} & Collaborative Filtering & User-Item Graph & LGC & Mean-Pooling\\
DGCF \citep{wang2020disentangled} & Collaborative Filtering & User-Item Graph & Dynamic Routing & Mean-Pooling \\
CAGCN \citep{wang2023collaboration} &  Collaborative Filtering & User-Item Graph & GCN+CIR & Mean-Pooling \\
\midrule
SR-GNN \citep{wu2019session} & Session-based & Transition Graph & GGNN & Soft-Attention\\
GC-SAN \citep{wu2019session,xu2019graphrec} & Session-based & Session Graph & GGNN & Self-Attention\\
FGNN \citep{qiu2019rethinking}  & Session-based & Session Graph & GAT & Last-Layer \\
GAG \citep{qiu2020gag} & Session-based & Session Graph & GCN & Self-Attention \\
GCE-GNN \citep{wang2020global} & Session-based & Transition+Global & GAT & Sum-Pooling\\
\midrule
HyperRec \citep{wang2020next} & Sequence-based & Sequential HyperGraph & HGCN & Self-Attention \\
DHCF \citep{ji2020dual} & Collaborative Filtering & Dual HyperGraph & JHConv & Last-Layer \\
MBHT \citep{yang2022multi} & Sequence-based & Learnable HyperGraph & Transformer &  Cross-View Attention \\
HCCF \citep{xia2022hypergraph} & Collaborative Filtering & Learnable HyperGraph & HGCN & Last-Layer \\
H$^3$Trans \citep{xu2023correlative} & Sequence-based & Hierarchical HyperGraph & Message-passing & Last-Layer \\
STHGCN \citep{yan2023spatio} & POI Recommendation & Spatio-temporal HyperGraph & HGCN & Mean-Pooling \\

\bottomrule
\end{tabular}
}
\end{table*}

\subsection{User-Item Bipartite Graph}
\subsubsection{Graph Construction}
A undirected bipartite graph where the vertex set $\mathcal{V}=\mathcal{U}\cup\mathcal{I}$ and the undirected edge set $\mathcal{E}=\{(u,i)|u\in\mathcal{U}\land i\in\mathcal{I}\}$. Under this case the graph adjacency can be directly obtained from the interaction matrix, thus the optimization target on the user-item bipartite graph is equivalent to collaborative filtering tasks such as MF \citep{koren2009matrix} and SVD++ \citep{koren2008factorization}.

There have been plenty of previous works that applied GNNs on the constructed user-item bipartite graphs. GC-MC \citep{berg2017graph} firstly applies graph convolution networks to user-item recommendation and optimizes a graph autoencoder (GAE) to reconstruct interactions between users and items. NGCF \citep{wang2019neural} introduces the concept of Collaborative Filtering (CF) into graph-based recommendations by modeling the affinity between neighboring nodes on the interaction graph. MMGCN \citep{wei2019mmgcn} extends the graph-based recommendation to multi-modal scenarios by constructing different subgraphs for each modal. LightGCN \citep{he2020lightgcn} improves NGCF by removing the non-linear activation functions and simplifying the message function. With the development of disentangled representation learning, there are works like DGCF \citep{wang2020disentangled} that introduce disentangled graph representation learning to represent users and items from multiple disentangled perspectives. Additionally, having realized the limitation of the existing message-passing scheme in capturing collaborative signals, CAGCN \citep{wang2023collaboration} proposes Common Interacted Ratio (CIR) as a recommendation-oriented topological metric for GNN-based recommender models.

\subsubsection{Graph Propagation Scheme}
A common practice is to follow the traditional message-passing networks (MPNNs) and design the graph propagation method accordingly. GC-MC adopts vanilla GCNs to encode the user-item bipartite graph. NGCF enhances GCNs by considering the affinity between users and items. The message function of NGCF from node $j$ to $i$ is formulated as:
\begin{equation}
\begin{cases}
    m_{i\leftarrow j}=\frac{1}{\sqrt{|\mathcal{N}_i||\mathcal{N}_j|}}(W_1e_j+W_2(e_i\odot e_j)) \\
    m_{i\leftarrow i}=W_1e_i
\end{cases},
\end{equation}
where $W_1,W_2$ are trainable parameters, $e_i$ represents $i$'s representation from previous layer. The matrix form can be further provided by:
\begin{equation}
\label{eq:ngcf}
    E^{(l)}=\text{LeakyReLU}((\mathcal{L}+I)E^{(l-1)}W_1^{(l)}+\mathcal{L}E^{(l-1)}\odot E^{(l-1)}W_2^{(l)}),
\end{equation}
where $\mathcal{L}$ represents the Laplacian matrix of the user-item graph. The element-wise product in Eq. \ref{eq:ngcf} represents the affinity between connected nodes, containing the collaborative signals from interactions.

However, the notable heaviness and burdensome calculation of NGCF's architecture hinder the model from making faster recommendations on larger graphs. LightGCN solves this issue by proposing Light Graph Convolution (LGC), which simplifies the convolution operation with:
\begin{equation}
    e_i^{(l+1)}=\sum_{j\in\mathcal{N}_i}\frac{1}{\sqrt{|\mathcal{N}_i||\mathcal{N}_j|}}e_j^{(l)}.
\end{equation}

When an interaction takes place, e.g. a user clicks a particular item, there could be multiple intentions behind the observed interaction. Thus it is necessary to consider the various disentangled intentions among users and items. DGCF proposes the cross-intent embedding propagation scheme on the graph, inspired by the dynamic routing algorithm of capsule networks \citep{sabour2017dynamic}. To formulate, the propagation process maintains a set of routing logits $\tilde{S}_k(u,i)$ for each user $u$. The weighted sum aggregator to get the representation of $u$ can be defined as:
\begin{equation}
    u_k^t=\sum_{i\in\mathcal{N}_u}\mathcal{L}_k^t(u,i)\cdot i_k^0
\end{equation}
for $t$-th iteration, where $\mathcal{L}_k^t(u,i)$ denotes the Laplacian matrix of $S_k^t(u,i)$, formulated as:
\begin{equation}
    \mathcal{L}_k^t(u,i)=\frac{S_k^t}{\sqrt{[\sum_{i'\in\mathcal{N}_u}S_k^t(u,i')]\cdot [\sum_{u'\in\mathcal{N}_i}S_k^t(u',i)]}}.
\end{equation}

\subsubsection{Node Representations}
After the graph propagation module outputs node-level representations, there are multiple methods to leverage node representations for recommendation tasks. A plain solution is to apply a readout function on layer outputs like the concatenation operation used by NGCF:
\begin{equation}
    e^*=Concat(e^{(0)},...,e^{(L)})=e^{(0)}\Vert ...\Vert e^{(L)}.
\end{equation}

However, the readout function among layers would neglect the relationship between the target item and the current user. A general solution is to use the attention mechanism \citep{vaswani2017attention} to reweight and aggregate the node representations. SR-GNN adapts soft-attention mechanism to model the item-item relationship:
\begin{equation}
    \begin{split}
        \alpha_i&= \textbf{q}^T\sigma(W_1e_t+W_2e_i+c),\\
        s_g &=\sum_{i=1}^{n-1}\alpha_ie_i,
    \end{split}
\end{equation}
where $\textbf{q},\ W_1,\ W_2$ are trainable matrices.

Some methods focus on exploiting information from multiple graph structures. A common practice is contrastive learning, which maximizes the mutual information between hidden representations from several views. HCCF leverage InfoNCE loss as the estimator of mutual information, given a pair of representation $z_{i},\Gamma_{i}$ for node $i$, controlled by temperature parameter $\tau$:
\begin{equation}
    \mathcal{L}_{InfoNCE}(i)=-\log\frac{\exp(cosine(z_i,\Gamma_i))/\tau}{\sum_{i'\neq i}\exp(cosine(z_i,\Gamma_{i'}))/\tau}.
\end{equation}

Besides InfoNCE, there exist several other ways to combine node representations from different views. For instance, MBHT applies an attention mechanism to fuse multiple semantics, DisenPOI adapts bayesian personalized ranking loss (BPR) \citep{rendle2012bpr} as a soft estimator for contrastive learning, and KBGNN applies pair-wise similarities to ensure the consistency from two views.

\subsection{Transition Graph}
\subsubsection{Transition Graph Construction}
Since sequence-based recommendation (SR) is one of the fundamental problems in recommender systems, some researches focus on modeling the sequential information with GNNs. A commonly applied way is to construct transition graphs based on each given sequence according to the clicking sequence by a user. To formulate, given a user $u$'s clicking sequence $s_u=[i_{u,1},i_{u,2},...,i_{u,n}]$ containing $n$ items, noting that there could be duplicated items, the sequential graph is constructed via $\mathcal{G}_s=\{\text{SET}(s_u),\mathcal{E}\}$, where $\forall \left<i_j,i_k\right> \in \mathcal{E}$ indicates there exists a successive transition from $i_j$ to $i_k$. Since $\mathcal{G}_s$ are directed graphs, a widely used way to depict graph connectivity is by building the connection matrix $A_s\in\mathbb{R}^{n\times 2n}$. $A_s$ is the combination of two adjacency matrices $A_s=[A_s^{(in)}; A_s^{(out)}]$, which denotes the normalized node degrees of incoming and outgoing edges in the session graph respectively.

The proposed transition graphs that obtain user behavior patterns have been demonstrated important to session-based recommendations \citep{li2017neural,liu2018stamp}. SR-GNN and GC-SAN \citep{wu2019session,xu2019graphrec} propose to leverage transition graphs and apply attention-based GNNs to capture the sequential information for session-based recommendation. FGNN \citep{qiu2019rethinking} formulates the recommendation within a session as a graph classification problem to predict the next item for an anonymous user. GAG \citep{qiu2020gag} and GCE-GNN \citep{wang2020global} further extend the model to capture global embeddings among multiple session graphs.

\subsubsection{Session Graph Propagation}
Since the session graphs are directed item graphs, there have been multiple session graph propagation methods to obtain node representations on session graphs.

SR-GNN leverages Gated Graph Neural Networks (GGNNs) to obtain sequential information from a given session graph adjacency $A_s=[A_s^{(in)};A_s^{(out)}]$ and item embedding set $\{e_i\}$:
\begin{align}
a_t&=A_s[e_1,...,e_{t-1}]^TH+b,\\
z_t&=\sigma(W_za_t+U_ze_{t-1}),\\
r_t&=\sigma(W_ra_t+U_re_{t-1}),\\
\tilde{e_t}&=\tanh(W_oa_t+U_o(r_t\odot e_{t-1})),\\
e_t&=(1-z_t)\odot e_{t-1}+z_t\tilde{e_t},
\end{align}
where $W$s and $U$s are trainable parameters. GC-SAN extend GGNN by calculating initial state $a_t$ separately to better exploit transition information:
\begin{equation}
    a_t=Concat(A_s^{(in)}([e_1,...,e_{t-1}W_a^{(in)}]+b^{(in)}),A_s^{(out)}([e_1,...,e_{t-1}W_a^{(out)}]+b^{(out)})).
\end{equation}

\subsection{HyperGraph}
\subsubsection{Hypergraph Topology Construction}
Motivated by the idea of modeling hyper-structures and high-order correlation among nodes, hypergraphs \citep{feng2019hypergraph} are proposed as extensions of the commonly used graph structures. For graph-based recommender systems, a common practice is to construct hyper structures among the original user-item bipartite graphs. To be specific, an incidence matrix of a graph with vertex set $\mathcal{V}$ is presented as a binary matrix $H\in\{0,1\}^{|\mathcal{V}|\times |\mathcal{E}|}$, where $\mathcal{E}$ represents the set of hyperedges. Each entry $h(v,e)$ of $H$ depicts the connectivity between vertex $v$ and hyperedge $e$:
\begin{equation}
h(v,e)=
    \begin{cases}
    1\ if\ v\in e \\
    0\ if\ v\notin e
    \end{cases}.
\end{equation}

Given the formulation of hypergraphs, the degrees of vertices and hyperedges of $H$ can then be defined with two diagonal matrices $D_v\in\mathbb{N}^{|\mathcal{V}|\times|\mathcal{V}|}$ and $D_e\in\mathbb{N}^{|\mathcal{E}|\times|\mathcal{E}|}$, where
\begin{equation}
    D_v(i;i)=\sum_{e\in\mathcal{E}}h(v_i,e),\ \ \ \ 
    D_e(j;j)=\sum_{v\in\mathcal{V}}h(v,e_j).
\end{equation}

The development of Hypergraph Neural Networks (HGNNs) \citep{feng2019hypergraph,zhou2006learning,huang2015learning} have shown to be capable of capturing the high-order connectivity between nodes. HyperRec \citep{wang2020next} firstly attempts to leverage hypergraph structures for sequential recommendation by connecting items with hyperedges according to the interactions with users during different time periods. DHCF \citep{ji2020dual} proposes to construct hypergraphs for users and items respectively based on certain rules, to explicitly capture the collaborative similarities via HGNNs. MBHT \citep{yang2022multi} combines hypergraphs with a low-rank self-attention mechanism to capture the dynamic heterogeneous relationships between users and items. HCCF \citep{xia2022hypergraph} uses the contrastive information between hypergraph and interaction graph to enhance the recommendation performance. To extend the model's ability to multi-domain categories of items, H$^3$Trans \citep{xu2023correlative} incorporates two hyperedge-based modules and leverages hierarchical hypergraph propagation to transfer from domains. STHGCN \citep{yan2023spatio} formulates a spatio-temporal hypergraph structure for POI recommendation.

\subsubsection{Hyper Graph Message Passing}
With the development of HGNNs, previous works have proposed different variants of HGNN to better exploit hypergraph structures. A classic high-order hyper convolution process on a fixed hypergraph $\mathcal{G}=\{\mathcal{V},\mathcal{E}\}$ with hyper adjacency $H$ is given by:
\begin{equation}
    g\star X=D_v^{-1/2}HD_e^{-1}H^TD_v^{-1/2}X\Theta,
\end{equation}
where $D_v,\ D_e$ are degree matrices of nodes and hyperedges, $\Theta$ denotes the convolution kernel. For hyper adjacency matrix $H$, DHCF refers to a rule-based hyperstructure via k-order reachable rule, where nodes in the same hyperedge group are k-order reachable to each other:
\begin{equation}
    A_u^k=\min(1,\text{power}(A\cdot A^T,k)),
\end{equation}
where $A$ denotes the graph adjacency matrix. By considering the situations where $k=1,2$, the matrix formulation of the hyper connectivity of users and items is calculated with:
\begin{equation}
    \begin{cases}
        H_u=A\Vert(A(A^TA)) \\
        H_i=A^T\Vert(A^T(AA^T))
    \end{cases},
\end{equation}
which depicts the dual hypergraphs for users and items.

HCCF proposes to construct a learnable hypergraph to depict the global dependencies between nodes on the interaction graph. To be specific, the hyperstructure is factorized with two low-rank embedding matrices to achieve model efficiency:
\begin{equation}
    H_u=E_u\cdot W_u,\ H_v=E_v\cdot W_v.
\end{equation}

\subsection{Other Graphs}
Since there are a variety of recommendation scenarios, several tailored designed graph structures have been proposed accordingly, to better exploit the domain information from different scenarios. For instance, CKE \citep{zhang2016collaborative} and MKR \citep{wang2019multi} introduce Knowledge graphs to enhance graph recommendation. GSTN \citep{wang2022graph}, KBGNN \citep{ju2022kernel}, DisenPOI \citep{qin2022disenpoi} and Diff-POI \citep{qin2023diffusion} propose to build geographical graphs based on the distance between Point-of-Interests (POIs) to better model the locality of users' visiting patterns. TGSRec \citep{fan2021continuous} and DisenCTR \citep{wang2022disenctr} empower the user-item interaction graphs with temporal sampling between layers to obtain sequential information from static bipartite graphs.

\subsection{Summary}
This section introduces the application of different kinds of graph neural networks in recommender systems and can be summarized as follows:

\begin{itemize}
    \item \textbf{Graph Constructions.} There are multiple options for constructing graph-structured data for a variety of recommendation tasks. For instance, the user-item bipartite graphs reveal the high-order collaborative similarity between users and items, and the transition graph is suitable for encoding sequential information in clicking history. These diversified graph structures provide different views for node representation learning on users and items, and can be further used for downstream ranking tasks.
    \item \textbf{Challenges and Limitations.} Though the superiority of graph-structured data and GNNs against traditional methods has been widely illustrated, there are still challenges unsolved. For example, the computational cost of graph methods is normally expensive and thus unacceptable in real-world applications. The data sparsity and cold-started issue in graph recommendation remains to be explored as well.
    \item \textbf{Future Works.} In the future, an efficient solution for applying GNNs in recommendation tasks is expected. There are also some attempts \citep{fan2021continuous,wang2022disenctr,qin2024learning} on incorporating temporal information in graph representation learning for sequential recommendation tasks.
\end{itemize}

\section{Traffic Analysis}
Intelligent Transportation Systems (ITS) are essential for safe, reliable, and efficient transportation in smart cities, serving the daily commuting and traveling needs of millions of people. To support ITS, advanced modeling and analysis techniques are necessary, and Graph Neural Networks (GNNs) are a promising tool for traffic analysis. GNNs can effectively model spatial correlations, making them well-suited for analyzing complex transportation networks. As such, GNNs have garnered significant interest in the traffic domain for their ability to provide insights into traffic patterns and behaviors~\citep{li2024survey}.

In this section, we first conclude the main GNN research directions in the traffic domain, and then we summarize the typical graph construction processes in different traffic scenes and datasets. Finally, we list the classical GNN workflows for dealing with tasks in traffic networks. A summary is provided in Table \ref{tab:traffic_summary}.

\subsection{Research Directions in Traffic Domain}
We summarize main GNN research directions in the traffic domain as follows,
\begin{itemize}
    \item \textbf{Traffic Flow Forecasting}. Traffic flow forecasting plays an indispensable role in ITS \citep{ran2012modeling, dimitrakopoulos2010intelligent}, which involves leveraging spatial-temporal data collected by various sensors to gain insights into future traffic patterns and behaviors. Classic methods, like autoregressive integrated moving average (ARIMA) \citep{box1970distribution}, support vector machine (SVM) \citep{hearst1998support} and recurrent neural networks (RNN) \citep{connor1994recurrent} can only model time series separately without considering their spatial connections. To address this issue, graph neural networks (GNNs) have emerged as a powerful approach for traffic forecasting due to their strong ability of modeling complex graph-structured correlations \citep{jiang2022graph, xie2020urban, bui2021spatial, zhao2023dynamic, rao2022fogs, LI2023119374, pmlr-v206-oskarsson23a}. 
    \item \textbf{Trajectory Prediction}. Trajectory prediction is a crucial task in various applications, such as autonomous driving and traffic surveillance, which aims to forecast future positions of agents in the traffic scene. However, accurately predicting trajectories can be challenging, as the behavior of an agent is influenced not only by its own motion but also by interactions with surrounding objects. To address this challenge, Graph Neural Networks (GNNs) have emerged as a promising tool for modeling complex interactions in trajectory prediction \citep{mohamed2020social, cao2021spectral, zhou2021ast, sun2020recursive}. By representing the scene as a graph, where each node corresponds to an agent and the edges capture interactions between them, GNNs can effectively capture spatial dependencies and interactions between agents. This makes GNNs well-suited for predicting trajectories that accurately capture the behavior of agents in complex traffic scenes.
    \item \textbf{Traffic Anomaly Detection.} Anomaly detection is an essential support for ITS. There are lots of traffic anomalies in daily transportation systems, for example, traffic accidents, extreme weather and unexpected situations. Handling these traffic anomalies timely can improve the service quality of public transportation. The main trouble of traffic anomaly detection is the highly twisted spatial-temporal characteristics of traffic data. The criteria and influence of traffic anomaly vary among locations and times. GNNs have been introduced and achieved success in this domain \citep{deng2022graph, chen2021learning, zhang2022automatic, deng2021graph}.
    \item \textbf{Others.} Traffic demand prediction targets at estimating the future number of traveling at some location. It is of vital and practical significance in the resource scheduling for ITS. By using GNNs, the spatial dependencies of demands can be revealed \citep{yao2018deep, yang2020using}. What is more, urban vehicle emission analysis is also considered in recent work, which is closely related to environmental protection and gains increasing researcher attention \citep{xu2020spatiotemporal}.
\end{itemize}

\begin{table*}[t]
\centering
\caption{Summary of graph models for traffic analysis. }
\label{tab:traffic_summary}
\resizebox{0.95\textwidth}{!}{
\begin{tabular}{l cccccc}
\toprule
Models & Tasks & Adjcency matrices  & GNN types & Temporal modules\\
\midrule  
STGCN\citep{yu2017spatio} & Traffic Flow Forecasting & Fixed Matrix & GCN & TCN\\
DCRNN\citep{li2017diffusion} & Traffic Flow Forecasting & Fixed Matrix & ChebNet & RNN \\
AGCRN \citep{bai2020adaptive} & Traffic Flow Forecasting & Dynamic Matrix & GCN & GRU\\
ASTGCN \citep{guo2019attention} & Traffic Flow Forecasting & Fixed Matrix & GAT & Attention\&TCN \\ 
GraphWaveNet \citep{wu2019graph} & Traffic Flow Forecasting & Dynamic Matrix & GCN & Gated-TCN \\
STSGCN \citep{song2020spatial} & Traffic Flow Forecasting & Dynamic Matrix & GCN & Cropping \\ 
LSGCN \citep{huang2020lsgcn} & Traffic Flow Forecasting & Fixed Matrix & GAT & GLU \\
GAC-Net \citep{song2020graph} & Traffic Flow Forecasting & Fixed Matrix & GAT & Gated-TCN \\ 
STGODE \citep{fang2021spatial} & Traffic Flow Forecasting & Fixed Matrix & Graph ODE & TCN \\
STG-NCDE \citep{choi2022graph} & Traffic Flow Forecasting & Dynamic Matrix & GCN & NCDE \\
DDGCRN \citep{WENG2023109670} & Traffic Flow Forecasting & Dynamic Matrix & GAT & RNN \\
MS-ASTN \citep{wang2020multicikm} & OD Flow Forecasting & OD Matrix & GCN & LSTM \\

Social-STGCNN \citep{mohamed2020social} & Trajectory Prediction & Fixed Matrix & GCN & TXP-CNN \\
RSBG \citep{sun2020recursive} & Trajectory Prediction & Dynamic Matrix & GCN & LSTM \\
ATG \citep{10.1145/3534678.3542671} & Trajectory Prediction & Fixed Matrix & GODE & NODE \\
STGAN \citep{deng2022graph} & Anomaly Detection & Fixed Matrix & GCN & GRU \\
DMVST-VGNN \citep{jin2020deep} & Traffic Demand Prediction & Fixed Matrix & GAT & GLU \\
DST-GNN \citep{huang2022dynamical} & Traffic Demand Prediction & Dynamic Matrix & GCN & Transformer \\
TC-SGC \citep{pan2022traffic} & Traffic Speed Prediction & Fixed Matrix & GCN & GRU \\ 

\bottomrule
\end{tabular}
}
\end{table*}

\subsection{Traffic Graph Construction}
\smallskip
\subsubsection{Traffic Graph}
The traffic network is represented as a graph $\mathcal{G}=(V, E, A)$, where $V$ is the set of $N$ traffic nodes, $E$ is the set of edges, and $A \in \mathbb{R}^{N\times N}$ is an adjacency matrix representing the connectivity of $N$ nodes. In the traffic domain, $V$ usually represents a set of physical nodes, like traffic stations or traffic sensors. The features of nodes typically depend on the specific task. Take traffic flow forecasting as an example. The features are the traffic flows, i.e., the historical time series of nodes. The traffic flow can be represented as a flow matrix $X \in \mathbb{R}^{N\times T}$, where $N$ is the number of traffic nodes and $T$ is the length of historical series, and $X_{nt}$ denotes the traffic flow of node $n$ at time $t$. The goal of traffic flow forecasting is to learn a mapping function $f$ to predict the traffic flow during future $T'$ steps given the historical $T$ step information, which can be formulated as follows:
\begin{equation}
    \left[ X_{:, t-T+1}, X_{:, t-T+2}, \cdots, X_{:, t}; \mathcal{G}\right] \stackrel{f}{\longrightarrow} \left[ X_{:, t+1}, X_{:, t+2}, \cdots, X_{:, t+T'}\right].
\end{equation}

\subsubsection{Graph Construction} Constructing a graph to describe the interactions among traffic nodes, i.e., the design of the adjacency matrix $A$, is the key part of traffic analysis. The mainstream designs can be divided into two categories, fixed matrix and dynamic matrix. 

\textbf{Fixed matrix.} Lots of works assume that the correlations among traffic nodes are fixed and constant over time, and they design a fixed and pre-defined adjacency matrix to capture the spatial correlation. Here we list several common choices of fixed adjacency matrix. 

The \textbf{connectivity matrix} is the most natural construction way. It relies on the support of road map data. The element of the connectivity matrix is defined as 1 if two nodes are physically connected and 0 otherwise. This binary format is convenient to construct and easy to interpret. 

The \textbf{distance-based matrix} is also a common choice, which shows the connection between two nodes more precisely. The elements of the matrix are defined as the function of distance between two nodes (driving distance or geographical distance). A typical way is to use the threshold Gaussian function as follows, 
\begin{equation}
    A_{ij} = \left\{
    \begin{array}{cr}
         \exp(-\frac{d_{ij}^2}{\sigma^2}), &d_{ij} < \epsilon \\
         0, &d_{ij} > \epsilon  \\
    \end{array}
    \right. ,
\end{equation}
where $d_{ij}$ is the distance between node $i$ and $j$, and $\sigma$ and $\epsilon$ are two hyperparameters to control the distribution and the sparsity of the matrix.

Another kind of fixed adjacency matrix is the \textbf{similarity-based matrix}. In fact, a similarity matrix is not an adjacency matrix to some extent. It is constructed according to the similarity of two nodes, which means the neighbors in the similarity graph may be far away in the real world. There are various similarity metrics. For example, many works measure the similarity of two nodes by their functionality, e.g., the distribution of surrounding points of interest (POIs). The assumption behind this is that nodes that share similar functionality may share similar traffic patterns. We can also define the similarity through the historical flow patterns. To compute the similarity of two-time series, a common practice is to use Dynamic Time Wrapping (DTW) algorithm \citep{muller2007dynamic}, which is superior to other metrics due to its sensitivity to shape similarity rather than point-wise similarity. 
Specifically, given two time series $X=(x_1, x_2, \cdots, x_n)$ and $Y=(y_1, y_2, \cdots, y_n)$, DTW is a dynamic programming algorithm defined as 
\begin{equation}\label{4}
  D(i, j) = dist(x_i,y_j) + \min \left( D(i-1,j),D(i, j-1),D(i-1,j-1)\right),
\end{equation}
where $D(i,j)$ represents the shortest distance between subseries $X=(x_1, x_2, \cdots, x_i)$ and $Y=(y_1, y_2, \cdots, y_j)$, and $dist(x_i,y_j)$ is some distance metric like absolute distance. As a result, $DTW(X,Y)=D(n,n)$ is set as the final distance between $X$ and $Y$, which better reflects the similarity of the two-time series compared to the Euclidean distance.

\textbf{Dynamic matrix.} The pre-defined matrix is sometimes unavailable and cannot reflect complete information of spatial correlations. The dynamic adaptive matrix is proposed to solve the issue. The dynamic matrix is learned from input data automatically. To achieve the best prediction performance, the dynamic matrix will manage to infer the hidden correlations among nodes, more than those physical connections. 

A typical practice is learning adjacency matrix from node embeddings \citep{bai2020adaptive}. Let $E_A \in \mathbb{R}^{N\times d}$ be a learnable node embedding dictionary, where each row of $E_A$ represents the embedding of a node, $N$ and $d$ denote the number of nodes and the dimension of embeddings respectively. The graph adjacency matrix is defined as the similarities among node embeddings, 
\begin{equation}
    D^{-\frac{1}{2}}AD^{-\frac{1}{2}}=softmax\left(ReLU(E_A\cdot E_A^T)\right),
    \label{eq:agcrn}
\end{equation}
where $softmax$ function is to perform row-normalization, and $D^{-\frac{1}{2}}AD^{-\frac{1}{2}}$ is the Laplacian matrix.

\subsection{Typical GNN Frameworks in Traffic Domain}

\smallskip
\emph{Spatial Temporal Graph Convolution Network~(STGCN)}~\citep{yu2017spatio}. STGCN is a pioneering work in the spatial-temporal GNN domain. It utilizes graph convolution to capture spatial features, and deploys a gated causal convolution to extract temporal patterns. Specifically, the graph convolution and temporal convolution are defined as follows, 
\begin{align}
    \Theta *_{\mathcal{G}} x &= \theta (I_n + D^{-\frac{1}{2}}AD^{-\frac{1}{2}})x = \theta ( \Tilde{D}^{-\frac{1}{2}}\Tilde{A}\Tilde{D}^{-\frac{1}{2}})x, \\
    \Gamma *_{\mathcal{T}} y &= P \odot \sigma(Q),
\end{align}
where $\Theta$ is the parameter of graph convolution, $P$ and $Q$ are the outputs of a 1-d convolution along the temporal dimension. The sigmoid gate $\sigma(Q)$ controls how the states of $P$ are relevant for discovering hidden temporal patterns. In order to fuse features from both spatial and temporal
dimension, the spatial convolution layer and the temporal convolution layer are combined to construct a spatial temporal block to jointly deal with graph-structured time series, and more blocks can be stacked to achieve a more scalable and complex model.  

\smallskip
\emph{Diffusion Convolutional Recurrent Neural Network~(DCRNN)}~\citep{li2017diffusion}. DCRNN is a representative solution combining graph convolution networks with recurrent neural networks. It captures spatial dependencies by bidirectional random walks on the graph. The diffusion convolution operation on a graph is defined as:
\begin{equation}
    X *_{\mathcal{G}} f_{\theta} = \sum_{k=0}^K \left( \theta_{k, 1}(D_O^{-1}A)^k + \theta_{k, 2}(D_I^{-1}A)^k\right)X,
\end{equation}
where $\theta$ are parameters for the convolution filter, and $D_O^{-1}A, D_I^{-1}A$ represent the bidirectional diffusion processes respectively. In term of temporal dependency, DCRNN utilizes Gated Recurrent Units (GRU), and replace the linear transformation in the GRU with the diffusion convolution as follows, 
\begin{align}
    r^{(t)} &= \sigma(\Theta_r *_{\mathcal{G}}[X^{(t)}, H^{(t-1)}] + b_r), \\
    u^{(t)} &= \sigma(\Theta_u *_{\mathcal{G}}[X^{(t)}, H^{(t-1)}] + b_u), \\
    C^{(t)} &= \tanh(\Theta_C *_{\mathcal{G}}[X^{(t)}, (r^{(t)}\odot H^{(t-1)}] + b_c), \\
    H^{(t)} &= u^{(t)}\odot H^{(t-1)} + (1-u^{(t)})\odot C^{(t)},
\end{align}
where $X^{(t)}, H^{(t)}$ denote the input and output at time $t$, $r^{(t)}, u^{(t)}$ are the reset and update gates respectively, and $\Theta_r, \Theta_u, \Theta_C$ are parameters of convolution filters. Moreover, DCRNN employs a sequence-to-sequence architecture to predict future series. Both the encoder and the decoder are constructed with diffusion convolutional recurrent layers. The historical time series are fed into the encoder and the predictions are generated by the decoder. The scheduled sampling technique is utilized to solve the discrepancy problem between training and test distribution.

\smallskip
\emph{Adaptive Graph Convolutional Recurrent Network~(AGCRN)}~\citep{bai2020adaptive}. The focuses of AGCRN are two-fold. On the one hand, it argues that the temporal patterns are diversified and thus parameter-sharing for each node is inferior; on the other hand, it proposes that the pre-defined graph may be intuitive and incomplete for the specific prediction task. To mitigate the two issues, it designs a Node Adaptive Parameter Learning (NAPL) module to learn node-specific patterns for each traffic series, and a Data Adaptive Graph Generation (DAGG) module to infer the hidden correlations among nodes from data and to generate the graph during training. Specifically, the NAPL module is defined as follows, 
\begin{align}
    Z = (I_n+D^{-\frac{1}{2}}AD^{-\frac{1}{2}})XE_{\mathcal{G}}W_{\mathcal{G}} + E_{\mathcal{G}}b_{\mathcal{G}},
\end{align}
where $X\in \mathbb{R}^{N\times C}$ is the input feature, $E_{\mathcal{G}} \in \mathbb{R}^{N\times d}$ is a node embedding dictionary, $d$ is the embedding dimension ($d << N$), $W_{\mathcal{G}}\in \mathbb{R}^{d\times C\times F}$ is a weight pool. The original parameter $\Theta$ in the graph convolution is replaced by the matrix production of $E_{\mathcal{G}}W_{\mathcal{G}}$, and the same operation is applied for the bias. This can help the model to capture node-specific patterns from a pattern pool according to the node embedding.
The DAGG module has been introduced in \ref{eq:agcrn}. The whole workflow of AGCRN is formulated as follows, 
\begin{align}
    \Tilde{A} &= softmax(ReLU(EE^T)), \\
    z^{(t)} &= \sigma(\Tilde{A}[X^{(t)}, H^{(t-1)}]EW_z+Eb_z, \\
    r^{(t)} &= \sigma(\Tilde{A}[X^{(t)}, H^{(t-1)}]EW_r+Eb_r, \\
    \hat{H}^{(t)} &= \tanh (\Tilde{A}[X, r^{(t)}\odot H^{(t-1)}]EW_h+Eb_h, \\
    H^{(t)} &= z^{(t)}\odot H^{(t-1)} + (1-z^{(t)})\odot \hat{H}^{(t)}.
\end{align}

\smallskip
\emph{Attention Based Spatial-Temporal Graph Convolutional Networks~(ASTGCN)}~\citep{guo2019attention}. ASTGCN introduces two kinds of attention mechanisms into spatial-temporal forecasting, i.e., spatial attention and temporal attention. Spatial attention is defined as the following,
\begin{gather}
    S = V_S \sigma\left((XW_1)W_2(W_3X)^T+b_S\right), \\
    S_{i,j}'=\frac{\exp (S_{i,j})}{\sum_{j=1}^N\exp (S_{i,j})},
\end{gather}
where $S'$ is the attention score, and $W_1, W_2, W3$ are learnable parameters. A similar construction is applied for temporal attention. Besides the attention mechanism, ASTGCN also introduces multi-component fusion to enhance the prediction ability. The input of ASTGCN consists of three parts, the recent segments, the daily-periodic segments and the weekly-periodic segment. The three segments are processed by the main model independently and fused with learnable weights at last:
\begin{equation}
    Y = W_h\odot Y_h + W_d \odot Y_d + W_w \odot Y_w,
\end{equation}
where $Y_h, Y_d, Y_w$ denotes the predictions of different segments respectively. 

\section{Summary}
This section introduces graph models for traffic analysis and we provide a summary as follows:
\begin{itemize}
    \item \textbf{Techniques.} Traffic analysis is a classical spatial temporal data mining task, and graph models play a vital role for extracting spatial correlations. Typical procedures include graph construction, spatial dimension operations, temporal dimension operations and information fusion. There are multiple implementations for each procedure, each implementation has its strengths and weaknesses. By combining different implementations, various kinds of traffic analysis models can be created. Choosing the right combination of procedures and implementations is critical for achieving accurate and reliable traffic analysis results.
    \item \textbf{Challenges and Limitations.} Despite the remarkable success of graph representation learning in traffic analysis, there are still several challenges that need to be addressed in current studies. Firstly, external data, such as weather and calendar information, are not well-utilized in current models, despite their close relation to traffic status. The challenge lies in how to effectively fuse heterogeneous data to improve traffic analysis accuracy. Secondly, the interpretability of models has been underexplored, which could hinder their deployment in real-world transportation systems. Interpretable models are crucial for building trust and understanding among stakeholders, and more research is needed to develop models that are both accurate and interpretable. Addressing these challenges will be critical for advancing the state-of-the-art in traffic analysis and ensuring the deployment of effective transportation systems.
    \item \textbf{Future Works.} In the future, we anticipate that more data sources will be available for traffic analysis, enabling a more comprehensive understanding of real-world traffic scenes. From data collection to model design, there is still a lot of work to be done to fully leverage the potential of GNNs in traffic analysis. In addition, we expect to see more creative applications of GNN-based traffic analysis, such as designing traffic light control strategies, which can help to improve the efficiency and safety of transportation systems. To achieve these goals, it is necessary to continue advancing the development of GNN-based models and exploring new ways to fuse diverse data sources. Additionally, there is a need to enhance the interpretability of models and ensure their applicability to real-world transportation systems. We believe that these efforts will contribute to the continued success of traffic analysis and the development of intelligent transportation systems.
\end{itemize}

\section{Discussion}
In this section, we systematically compare the strengths and weaknesses of different components within both GNN architectures and learning paradigms. This comprehensive exploration aims to shed light on how these components can be harnessed to better serve the field of deep graph representation learning.

\begin{table}[t]
\renewcommand\arraystretch{0.935}
\caption{Comparison of different components in GNN architectures.}
\label{table_GNN_architectures}
\centering
\resizebox{0.85\textwidth}{!}{
\begin{tabular}{c|ccccc}
\toprule
Method & Advantages & Disadvantages \\
\midrule
Graph Convolutions & \makecell{Local information aggregation, \\parameter sharing} & \makecell{Limited global context, \\sensitivity to graph structure} \\
\midrule
Graph Kernel Neural Networks & \makecell{Kernel trick benefits, \\embedding similarity} & \makecell{Computational complexity, \\limited scalability} \\
\midrule
Graph Pooling & \makecell{Hierarchical representation, \\dimensionality reduction} & \makecell{Information loss, \\pooling strategy sensitivity} \\
\midrule
Graph Transformer & \makecell{Attention mechanism, \\parallelization}	 & \makecell{Computational intensity, \\limited interpretability} \\
\bottomrule
\end{tabular}
}
\end{table}

\subsection{GNN Architectures}
Here, we summarize the advantages and disadvantages of different components in GNN architectures in Table \ref{table_GNN_architectures}.

\smallskip
\textbf{Graph Convolutions}
\begin{itemize}
\item \textbf{Advantages:} 
(i) \emph{Local Information Aggregation:} Graph convolutions excel at capturing intricate local neighborhood information, making them highly effective for tasks where node representations heavily depend on nearby nodes.
(ii) \emph{Parameter Sharing:} Shared weights in graph convolutions enable the model to generalize robustly across different regions of the graph, enhancing efficiency.
\item \textbf{Disadvantages:} 
(i) \emph{Limited Global Context:} One limitation lies in the potential struggle to capture long-range dependencies in graphs, leading to information loss on more distant nodes.
(ii) \emph{Sensitivity to Graph Structure:} The performance of graph convolutions can be influenced by irregularities in graph structures, impacting their adaptability.
\end{itemize}

\textbf{Graph Kernel Neural Networks}
\begin{itemize}
\item \textbf{Advantages:} 
(i) \emph{Kernel Trick Benefits:} Leveraging kernel methods, graph kernel neural networks effectively learn on structured data, providing flexibility in capturing complex relationships.
(ii) \emph{Embedding Similarity:} These networks encode similarity measures between nodes, capturing nuanced graph structure information crucial for certain applications.
\item \textbf{Disadvantages:} 
(i) \emph{Computational Complexity:} Graph kernel methods may face challenges related to computational expense, particularly when dealing with large graphs.
(ii) \emph{Limited Scalability:} Scalability issues arise when extending these methods to graphs of varying sizes and structures.
\end{itemize}

\textbf{Graph Pooling}
\begin{itemize}
\item \textbf{Advantages:} 
(i) \emph{Hierarchical Representation:} Graph pooling contributes to creating hierarchical representations, enabling the model to capture information at multiple levels of granularity, fostering a more comprehensive understanding.
(ii) \emph{Dimensionality Reduction:} Effectively reduces computational load by downsampling the graph, making it more computationally tractable.
\item \textbf{Disadvantages:} 
(i) \emph{Information Loss:} Aggregating information during pooling may lead to a loss of fine-grained details, potentially impacting the model's performance on certain downstream tasks.
(ii) \emph{Pooling Strategy Sensitivity:} The choice of pooling strategy may significantly affect performance, emphasizing the importance of thoughtful strategy selection.
\end{itemize}

\textbf{Graph Transformer}
\begin{itemize}
\item \textbf{Advantages:} 
(i) \emph{Attention Mechanism:} Graph transformers leverage attention mechanisms to capture global dependencies and relationships effectively, enabling the model to consider the entire graph when making predictions.
(ii) \emph{Parallelization:} The attention mechanism allows for parallelization of computations, enhancing computational efficiency, particularly on hardware optimized for parallel processing.
\item \textbf{Disadvantages:} 
(i) \emph{Computational Intensity:} Transformers can be computationally intensive, especially with large graphs, requiring careful consideration of resource constraints.
(ii) \emph{Limited Interpretability:} The attention mechanism, while powerful, may lack interpretability, making it challenging to understand the model's decision-making process.
\end{itemize}

\begin{table}[t]
\renewcommand\arraystretch{0.935}
\caption{Comparison of different components in learning paradigms.}
\label{table_learning_paradigms}
\centering
\resizebox{1\textwidth}{!}{
\begin{tabular}{c|ccccc}
\toprule
Method & Advantages & Disadvantages \\
\midrule
\makecell{Supervised/Semi-Supervised Learning \\on Graphs} & \makecell{Utilization of labeled data, \\task-specific objectives} & \makecell{Data dependency, \\generalization challenges} \\
\midrule
Graph Self-Supervised Learning & \makecell{Data efficiency, \\transferability} & \makecell{Pretext task design, \\computationally intensive pre-training} \\
\midrule
Graph Structure Learning & \makecell{Incorporation of inherent graph properties, \\robustness to label sparsity} & \makecell{Limited task specificity, \\sensitivity to graph noises} \\
\bottomrule
\end{tabular}
}
\end{table}

\subsection{Learning Paradigms}
Here, we summarize the advantages and disadvantages of different components in learning paradigms in Table \ref{table_learning_paradigms}.

\smallskip
\textbf{Supervised/Semi-Supervised Learning on Graphs}
\begin{itemize}
\item \textbf{Advantages:} 
(i) \emph{Utilization of Labeled Data:} Supervised learning effectively utilizes labeled data for direct prediction, while semi-supervised learning leverages both labeled and unlabeled data, providing a flexible approach.
(ii) \emph{Task-Specific Objectives:} Clear task objectives guide the learning process, allowing for precise model training.
\item \textbf{Disadvantages:} 
(i) \emph{Data Dependency:} These paradigms heavily rely on the availability of labeled data, potentially limiting their applicability in scenarios with sparse labels.
(ii) \emph{Generalization Challenges:} Generalizing to unseen nodes or graphs can be challenging, particularly in dynamic or evolving graph structures.
\end{itemize}

\textbf{Graph Self-Supervised Learning}
\begin{itemize}
\item \textbf{Advantages:} 
(i) \emph{Data Efficiency:} Utilizes unlabeled data for pre-training, making efficient use of available resources and potentially reducing the need for extensive labeled datasets.
(ii) \emph{Transferability:} Pre-trained models can be fine-tuned for downstream tasks, enhancing generalization across various applications.
\item \textbf{Disadvantages:} 
(i) \emph{Pretext Task Design:} The effectiveness of self-supervised learning depends on the design of pretext tasks, requiring careful consideration of task relevance.
(ii) \emph{Computationally Intensive Pre-training:} The pre-training phase can be computationally intensive, especially when dealing with large and complex graphs.
\end{itemize}

\textbf{Graph Structure Learning}
\begin{itemize}
\item \textbf{Advantages:} 
(i) \emph{Incorporation of Inherent Graph Properties:} Focuses on exploiting the intrinsic structure of graphs, enhancing representation learning by considering the unique properties of graph data.
(ii) \emph{Robustness to Label Sparsity:} Less dependent on labeled data, making it suitable for scenarios with limited labels, and potentially more robust in the face of sparse data.

\item \textbf{Disadvantages:} 
(i) \emph{Limited Task Specificity:} May not be inherently task-specific and might not outperform task-tailored methods on certain applications that demand specialized representations.
(ii) \emph{Sensitivity to Graph Noises:} Performance may degrade in the presence of noisy or irregular graph structures, necessitating preprocessing steps for noise reduction.
\end{itemize}

\section{Future Directions}


In this section, we outline some prospective future directions of deep graph representation learning based on the above cornerstone, taxonomy, and real-world applications. We also outline a few more directions closer to the theoretical side.

\subsection{Application-Inspired Directions}

Since deep graph representation has been widely used these years, many problems have been solved while many others have arisen. While we observe many real-world applications, we conclude plenty of challenging problems that are not yet solved. Here in this subsection, we outline a few.

\subsubsection{Fairness in Graph Representation Learning}
One common aspect to care about is the fairness concern. Fairness, by definition, refers to the protected features that do not infect the outcome. In general, a data set's fairness refers to that the protected features are not influencing the data distribution. A model's fairness, on the other hand, refers to the concern that the output of our algorithms should not be affected by certain protected features. The protected features can be race, gender, etc.


Similar to the fairness challenge in many other fields of machine learning~\citep{chouldechova2018frontiers,mehrabi2021survey}, graph representation learning can easily suffer from bias from the data sets that inherit stereotypes from the real world. 
As graph representation has become increasingly popular in recent years, researchers are getting fairness into their sights~\citep{ma2021subgroup,dong2021individual}.

Different from i.i.d. data, graph data contains a lot of relational information that would bring about new challenges to bias detection~\citep{dai2021say,dong2022edits}.
Besides, another challenge is that better algorithms are needed to prevent the model from inheriting the input data's biases~\citep{ma2021subgroup,dong2022edits}.
Researchers have been working on various fairness notions regarding graph-representation learning. Such as group fairness, individual fairness, and application-specific fairness~\citep{dong2023fairness}. Application-specific fairness mostly refers to that in the field of recommender systems~\citep{fu2020fairness,li2021user} and knowledge graphs~\citep{fisher2019measuring}.

Fairness concern in graph mining is crucial in real-world applications. Researchers have already found significant gender and ethnic group bias in existing recommender systems~\citep{lambrecht2019algorithmic,sweeney2013discrimination}. Not to mention more sensitive domains of applications, such as loan approval and criminal justice~\citep{sarkar2020mitigating,agarwal2021towards}.

\subsubsection{Robustness in Graph Representation Learning}
Real-world data is always noisy, containing many different kinds of disruptions, and does not end up being a perfectly normal distribution. In the worst case, some noise can potentially prevent a model from learning the correct knowledge. 
Better robustness refers to the model having a better chance of reaching a relatively good and stable outcome while input is being manipulated. If a model is not robust enough, the performance can not be relied on.
Therefore, robustness is another important yet challenging consideration in deep graph representation learning. 
It is especially true when given that researchers have already found that existing graph representation learning models are vulnerable to adversarial data samples in general~\citep{chen2020survey,bojchevski2019adversarial,zugner2020certifiable,yuan2023alex,yuan2023learning,wang2022disencite}.

Again, similar to many other machine learning approaches that aim at solving real-world problems~\citep{carlini2017towards}, improving the robustness of deep graph representation models is a nontrivial direction. Either enhancing the models' robustness or conducting adversarial attacks to challenge the robustness of graph representations, are promising directions to go for~\citep{tang2020transferring,geisler2021robustness,gunnemann2022graph}.

There have been studies on enhancing graph neural networks' robustness, from different perspectives~\citep{chen2020survey,xu2021robustness}. There they introduced some metrics to measure the robustness of graph models, such as classification margin, and adversarial gap~\citep{xu2021robustness}.

One of the most common solutions to enhance robustness is to avoid using malicious samples. Starting from the preprocessing steps, some researchers have already studied how to get rid of the poisoned data. For example, \citep{xu2023edog} worked on sampling sub-graphs from the poisoned training data set and then detected outliers so that they could filter the adversarial edges. In particular, for example, since some attacker models are known as adding high-rank high-rank perturbations to data, in order to get rid of those attacks, \citep{entezari2020all} ``vaccinate'' data sets in pre-processing steps, by using low-rank approximation matrices. Some other works have a more general measurement of which edges to eliminate, such as using the Jaccard Similarity score to measure an edge's end nodes and remove the suspicious edges whose score is not high enough~\citep{wu2019adversarial}.

Moving one step ahead, it is natural to consider conducting anomaly detection. There has already been a lot of anomaly detection work done on static graphs~\citep{jiang2019anomaly,wang2021one}. These works are of great importance since many traditional anomaly detection methods fail to take link connections into consideration and behave poorly on graph data.
Nonetheless, anomaly detection works on dynamic graphs are much fewer~\citep{du2017deeplog}. Robustness on dynamic graphs has been of increasing importance in recent years~\citep{arp2022and} and is also worth being studied.

However, despite the efforts to eliminate malicious samples at an early stage, adversarial attacks can still easily reach the training stage. There we need more careful design of the model to make it robust when facing poisoned data directly on its own. 
Some researchers are focusing on adversarial training, by either utilizing adversarial mini-max objectives ~\citep{jin2021adversarial,li2020deeprobust}, or adding adversarial samples intentionally so as to learn how to deal with them~\citep{zhang2021backdoor,deng2019batch,yang2022pa}. 
Some other researchers also propose to relieve the threat of poisoned data by using a variance-based attention mechanism, under the assumption that fake links and nodes should have higher uncertainty in prediction results~\citep{zhu2019robust}.

\subsubsection{Adversarial Reprogramming}
With the emergence of pre-trained graph neural network models~\citep{hu2019strategies,hu2020gpt,qiu2020gcc}, introducing adversarial reprogramming~\citep{elsayed2018adversarial,zheng2021adversarial} into deep graph representation learning becomes another possibility as well. The major difference between adversarial reprogramming and adversarial attack \citep{zhao2020object, thys2019fooling, huang2017adversarial} lies in whether or not there is a particular target after putting some adversarial samples against the model.
An adversarial attack requires some small modifications to the input data samples. An adversarial attack is considered successful once the result is influenced. However, under the adversarial reprogramming settings, the task succeeds if and only if the influenced results can be used for another desired task.

This is to say, without changing much on the model's inner structure or fine-tuning its parameters, we might be able to use some pre-trained graph models for some other tasks that were not planned to be solved by these models in the first place. In other deep learning fields, adversarial reprogramming problems are normally solved by having the input carefully encoded, and output cleverly mapped.
On some graph data sets, such as chemical data sets and biology data sets, pre-trained models are already available. Therefore, there is a possibility that adversarial reprogramming could be applied in the future.

\subsubsection{Generalizing to Out of Distribution Data}
In order to perform better on unobserved data sets, in the ideal case, the representation we learn should better be able to generalize to some out-of-distribution (OOD) data. Being out-of-distribution is not identical to being misclassified. The misclassified samples are coming from the same distribution of the training data but the model fails to classify it correctly, while out-of-distribution refers to the case where the sample comes from a distribution other than the training data~\citep{hendrycks2016baseline, luo2023towards_graph,ju2023zero}. Being able to generalize to out-of-distribution data will greatly enhance a model's reliability in real life. And studying out-of-distribution generalized graph representation~\citep{li2022ood} is an opening field~\citep{li2022out}.
This is partly because of, currently, even the problem of detecting out-of-distribution data samples is not fully conquered yet~\citep{hendrycks2016baseline}.

In order to do something on the out-of-distribution data samples, we need to detect which samples belong to this type first. Detecting OOD samples itself is somewhat similar to novelty detection, or outlier detection problems~\citep{pimentel2014review}. Their major difference is whether or not a well-performed model conducting the original tasks remains part of our goal. Novelty detection cares only about figuring out who are the outliers; OOD detection requires our model to detect the outliers while keeping the performance unharmed at the same time.

\subsubsection{Interpretability in Graph Representation Learning}
Interpretability concern is another limitation that exists when researchers try to apply deep graph representation learning to some of the emerging application fields.
For instance, in the field of computational social science, researchers are urging more efforts in integrating explanation and prediction together~\citep{hofman2021integrating}. So as drug discovery, being able to explain why such a structure is chosen instead of another option, is very important~\citep{jimenez2020drug}. 
Generally speaking, neural networks are completely in black-box mode to human knowledge without making efforts to make them interpretable and explainable. Although more and more tasks are being handled by deep learning methods in many fields, the tool remains mysterious to most human beings. Even an expert in deep learning can not easily explain to you how the tasks are performed and what the model has learned from the data. This situation reduces the trustworthiness of the neural network models, prevents a human from learning more from the models' results, and even limits the potential improvements of the models themselves, without sufficient feedback to human beings.

Seeking for better interpretability is not only some personal interests of companies and researchers, in fact, as more and more ethical concerns arose since more and more black-box decisions were made by AI algorithms, interpretability has become a legal requirement~\citep{goodman2017european}.

Various approaches have been applied, serving the goal of better interpretability~\citep{zhang2021survey}. There we find existing works that provide either ad-hoc explanations after the results come out, or those actively change the model structure to provide better explanations; explanations by providing similar examples, highlighting some attributes to the input features, by making sense of some hidden layers and extract semantics from them, or by extracting logical rules; we also see local explanations that explain some particular samples, global explanations that explain the network as a whole, or hybrid.
Most of those existing directions make sense in a graph representation learning setting. 

Not a consensus has been reached on what are the best methods of making a model interpretable. Researchers are still actively exploring every possibility, and thus there are plenty of challenges and interesting topics in this direction.

\subsubsection{Causality in Graph Representation Learning}

Most of the existing studies~\citep{wu2020comprehensive,zhou2020graph} on graph representation learning is good at handling predictive (i.e. node classification, graph classification, link prediction) or descriptive (e.g. shortest path, centrality) tasks on static graphs. However, graph data, such as social networks, and citation networks, can change over time. To analyze their dynamic features, there is an emerging need to analyze causality on graphs. That is to say, apply some treatment to the system, and predict the following outcome.

In recent years, there have been increasing research works focusing on combining causality and machine learning models~\citep{madumal2020explainable,hu2021causal,richens2020improving, luo2023rignn}.
It is widely believed that making good use of causality will help models gain higher performances. However, finding the right way to model causality in many real-world scenarios remains challenging.
In real-world data sets, there usually exist unobserved confounders (i.e. the variables that affect both treatment assignment and outcome). To mitigate the confounder bias, most of the causality models, such as TARNet (treatment-agnostic representation network) which learns the representation of confounders, and CFR (counterfactual regression) which uses representation balancing techniques to minimize the distribution distance between the confounders' representations of the treatment and controlled groups~\citealp{shalit2017estimating},  are built upon strong ignorability assumption, meaning that there is no other confounder except for those who are already included in the observed features~\citep{ma2022learning}. On the other hand, models such as CEVAE (causal effect variational autoencoder) did not rely on the strong ignorability assumption. Instead, it assumes that it can infer the hidden confounders from the observed features~\citep{louizos2017causal}.

In particular, graph data is especially challenging for causal inference, due to the following few reasons~\citep{ma2022learning}: (1) multi-modality of the data; (2) non-neglectable hidden confounders; (3) complicated forms e.g. dynamic graphs; (4) network inference among individuals that no longer fulfill the assumptions of many traditional causal effect models; (5) graph-structured treatment.

Something to note is that the most common kind of graph that comes along the causal study, called ``causal graph'', is not necessarily identical to the kind of graphs we are studying in deep graph representation learning. Causal graphs are the kind of graphs whose nodes are factors and links represent causal relations. Up till now, they are among the most reliable tools for causal inference study.
Traditionally, causal graphs are defined by human experts. Recent works have shown that neural networks can help with scalable causal graph generation~\citep{xu2019scalable}. From this perspective, the story can be the other side around: besides using causal relations to enhance graph representation learning, it is also possible to use graph representation learning strategies to help with causal study. For example, a knowledge graph can be used to diagnose the root cause of performance anomaly in cloud applications~\citep{qiu2020causality}. Many researchers have already worked on mining the causality in graph-structured data and found this direction appealing~\citep{guo2020learning,ma2022assessing}.

\subsubsection{Emerging Application Fields}
Besides the above-mentioned directions solving existing challenges in the deep learning world, there are many emerging fields of application that naturally come along with graph-structured data.

For instance, the emerging fields of social network analysis that help with traditional social science or political science studies, and drug discovery that helps with medical science. Due to the nature of the data, such as the social network interactions (i.e. follow, retweet, reply), and drug molecule structures (i.e. atoms and bounds in between them), can be easily depicted as graph-structured data. Therefore, deep graph representation learning has much to do in these fields~\citep{abbas2021social,zhu2022survey,gaudelet2021utilizing,yang2023poisoning}.

Some basic problems on the social network are easily solved using graph representation learning strategies. Those basic problems include node classification, link prediction, graph classification, and so on. In practice, those problem settings could refer to real-world problems such as ideology prediction, interaction prediction, analyzing a social group, etc.
However, social network data typically has many unique features that could potentially stop the general-purposed models from performing well. For instance, social media data can be sparse, incomplete, and can be extremely imbalanced~\citep{zhao2021graphsmote}.
On the one hand, social media platforms themselves never have a clear and consistent topic.
On the other hand, people have clear goals when studying social media data, such as controversy detection~\citep{benslimane2022text}, rumor detection~\citep{takahashi2012rumor,hamidian2019rumor}, misinformation and dis-information detection~\citep{di2021fake}, or studying the dynamics of the system~\citep{kipf2018neural,luo2023graph,luo2023hope}.
The real-world social media data is naturally unlabeled, making even data annotation (i.e. labeling) itself a challenging task to deal with.
There are still a lot of open quests to be conquered, which deep graph representation learning can help with.

As for drug discovery, researchers have some interest in other perspectives beyond simply proposing a set of potentially functional structures, which is widely seen today. The other perspectives include having more interpretable results from the model's proposals~\citep{preuer2019interpretable,jimenez2020drug}, and considering synthetic accessibility~\citep{xie2021mars}.
These directions are important, in answer to some doubt on AI from the society~\citep{goodman2017european}, as well as from the tradition of chemistry studies~\citep{schneider2020rethinking}. Similar to the challenges we faced when combining social science and neural networks, chemical science would also prefer the black-box AI models to be interpretable instead. 
Some chemical scientists would also prefer AI tools to provide them with synthetic routes instead of the targeting structure itself. In practice, proposing new molecule structures is usually not the bottleneck, but synthesizing is. There are already some existing works focusing on conquering this problem~\citep{empel2019artificial,ishida2022ai}. But so far there is a gap between chemical experiments and AI tools, indicating that there is still plenty of improvement to be made.

Some chemistry researchers also found it useful to have material data better organized, given that the molecule structures are becoming increasingly complex, and a massive amount of research papers are describing the material's features from different aspects~\citep{walsh2023community}. This direction might be more closely related to knowledge base or even database systems. But in a way, given that the polymer structure is typically a node-link graph, graph representation learning might be able to help with dealing with such issues.

Besides, we also realize that most of the machine-generated molecule structures will be regarded as useless from a chemical scientist's view. The root cause is that, from a human expert's perspective, some properties (i.e. features) of the molecules are obvious, but to a machine learning model, it is not. A well-trained human expert will realize that some molecule structures are not stable under room temperature and standard atmosphere pressure. But according to the existing benchmarks and the available labeled data sets, a machine learning model could regard that particular structure as the best option.

In a way, we believer that, as more and more wet experiments are involved, researchers will gradually realize the importance of combining expertise knowledge with the graph representation learning models. It will enhance the interpretability and trustworthiness of many applications.

\subsection{Theory-Driven Directions}

Some other future directions dig into the root of graph theory. More specifically, it focuses on some fundamental improvements in neural network structure design, or better ways of expressing the graph representations. These directions require background knowledge of their mathematical backgrounds. All in all, breakthroughs in these directions might not end up with immediate impact, but every study in these directions has the potential to change the entire field, sooner or later.

\subsubsection{Mathematical Proof of Feasibility}

It has been a long-lasting problem that most of the existing deep learning approaches lack mathematical proof of their learnability, bound, etc~\citep{bouzerdoum1993neural,bartlett2017spectrally,liu2023fimo}. This problem relates to the difficulty of providing theoretical proof on a complicated structure like neural network~\citep{grohs2021proof}.

Currently, most of the theoretical proof aims at figuring out theoretical bounds~\citep{harvey2017nearly,bartlett2019nearly,karpinski1997polynomial}. There are multiple types of bounds with different problem settings. Such as: given a known architecture of the model, with input data satisfying particular normal distribution, prove that training will converge, and provide the estimated number of iterations. Most of these architectures being studied are simple, such as those made of multi-layer perceptron (MLP), or simply studying the updates of parameters in a single fully-connected layer.

In the field of deep graph representation learning, neural network architectures are typically much more complex than MLPs. Graph neural networks (GNNs), since the very beginning~\citep{defferrard2016convolutional,kipf2016semi}, involve a lot of approximation and simplification of mathematical theorems. Nowadays, most researchers rely heavily on the experimental results. No matter how wild an idea is, as long as it finally works out in an experiment, say, being able to converge and the results are acceptable, the design is acceptable. All these practices make the entire field somewhat experiments-oriented or experience-oriented, while there remains a huge gap between the theoretical proof and the frontier of deep graph representation.

It will be more than beneficial to the whole field if some researchers can push forward these theoretical foundations. However, these problems are incredibly challenging.

\subsubsection{Combining Spectral Graph Theory}

Down to the theory foundations, the idea of graph neural networks~\citep{shuman2013emerging,defferrard2016convolutional,kipf2016semi} initially comes from spectral graph theory~\citep{chung1997spectral}. In recent years, many researchers have investigated possible improvements in graph representation learning strategies via utilizing spectral graph theory~\citep{chen2020bridging,yang2021supergraph,mansourlakouraj2022multi,he2022msgnn}. 
For example, graph Laplacian is closely related to many properties, such as the connectivity of a graph. By studying the properties of Laplacian, it is possible to provide proof of graph neural network models' properties and to propose better models with desired advantages, such as robustness~\citep{fu2022p,runwal2022robustifying}.

Spectral graph theory provides a lot of useful insights into graph representation learning from a new perspective. There is a lot to be done in this direction.

\subsubsection{From Graph to Manifolds}

Many researchers are devoted to the direction of learning graph representation in non-Euclidean spaces~\citep{asif2021graph,saxena2020survey}. That is to say, to embed and compute on some other spaces that are not Euclidean, such as hyperbolic and spherical spaces.

Theoretical reasoning and experimental results have shown certain advantages of working on manifolds instead of standard Euclidean space. It is believed that these advantages are brought by their abilities to capture complex correlations on the surface manifold~\citep{zhou2022dynamic}. 
Besides, researchers have shown that, by combining standard graph representation learning strategies and manifold assumptions, models work better on preserving and acquiring the locality and similarity relationships ~\citep{fu2021recent}. Intuitively, sometimes two nodes' embeddings are regarded as way too similar in Euclidean space, but in non-Euclidean space, they are easily distinguishable. 

\section{Conclusion}

In this survey, we present a comprehensive and up-to-date overview of deep graph representation learning. We present a novel taxonomy of existing algorithms categorized into GNN architectures, learning paradigms, and applications. Technically, we first summarize the ways of GNN architectures namely graph convolutions, graph kernel neural networks, graph pooling, and graph transformer. Based on the different training objectives, we present three types of the most recent advanced learning paradigms namely: supervised/semi-supervised learning on graphs, graph self-supervised learning, and graph structure learning. Then, we provide several promising applications to demonstrate the effectiveness of deep graph representation learning. Last but not least, we discuss the future directions in deep graph representation learning that have potential opportunities.

\begin{acks}
This paper is partially supported by National Key Research and Development Program of China with Grant No. 2023YFC3341203, the National Natural Science Foundation of China (NSFC Grant Numbers 62306014 and 62276002) as well as the China Postdoctoral Science Foundation with Grant No. 2023M730057. All authors contributed equally to this research.
\end{acks}

\bibliographystyle{ACM-Reference-Format}
\bibliography{19_ref}


\end{document}